\documentclass[journal, 9pt, a4paper]{IEEEtran}
\usepackage{scalerel}
\usepackage{tikz}
\usetikzlibrary{svg.path}

\definecolor{orcidlogocol}{HTML}{A6CE39}
\tikzset{
  orcidlogo/.pic={
    \fill[orcidlogocol] svg{M256,128c0,70.7-57.3,128-128,128C57.3,256,0,198.7,0,128C0,57.3,57.3,0,128,0C198.7,0,256,57.3,256,128z};
    \fill[white] svg{M86.3,186.2H70.9V79.1h15.4v48.4V186.2z}
                 svg{M108.9,79.1h41.6c39.6,0,57,28.3,57,53.6c0,27.5-21.5,53.6-56.8,53.6h-41.8V79.1z M124.3,172.4h24.5c34.9,0,42.9-26.5,42.9-39.7c0-21.5-13.7-39.7-43.7-39.7h-23.7V172.4z}
                 svg{M88.7,56.8c0,5.5-4.5,10.1-10.1,10.1c-5.6,0-10.1-4.6-10.1-10.1c0-5.6,4.5-10.1,10.1-10.1C84.2,46.7,88.7,51.3,88.7,56.8z};
  }
}

\newcommand\orcidicon[1]{\href{https://orcid.org/#1}{\mbox{
      \hspace*{-3ex}
      \raisebox{0.5ex}{
      \scalerel*{     
\begin{tikzpicture}[yscale=-1,transform shape]
\pic{orcidlogo};
\end{tikzpicture}
}{k}}}}}

\usepackage{color}
\usepackage{relsize}
\usepackage{amsmath,amsfonts,amssymb}

\usepackage[caption=false]{subfig}
\usepackage[inline]{enumitem}
\usepackage{varwidth}
\usepackage{braket}
\usepackage{placeins}

\usepackage{nicefrac}

\usepackage{algorithm}
\usepackage{cite}

\usepackage{comment}

\usepackage{mathtools}
\usepackage{colonequals}
\usepackage[noend]{algpseudocode}

\usepackage{microtype}

\usepackage{graphicx}
\usepackage{epstopdf}
\ifpdf
  \DeclareGraphicsExtensions{.eps,.pdf,.png,.jpg}
\else
  \DeclareGraphicsExtensions{.eps}
\fi

\usepackage{amsopn}

\usepackage{hyperref}
\hypersetup{draft}

\usepackage{cleveref}
\crefname{algorithm}{Alg.}{Algs.}
\Crefname{algorithm}{Algorithm}{Algorithms}
\crefname{figure}{Fig.}{Figs.}
\Crefname{figure}{Figure}{Figures}
\crefname{section}{Section}{Sections.}
\crefname{subsection}{Subsection}{Subsections.}
\crefname{equation}{Eq.}{Eqs.}

\makeatletter
\newcommand{\@giventhatstar}[2]{\left(#1\,\middle|\,#2\right)}
\newcommand{\@giventhatnostar}[3][]{#1(#2\,#1|\,#3#1)}
\newcommand{\giventhat}{\@ifstar\@giventhatstar\@giventhatnostar}
\makeatother

\makeatletter
\newcommand*{\rom}[1]{\expandafter\@slowromancap\romannumeral #1@}
\makeatother

\newcommand{\0}{\M{0}}
\newcommand{\Expec}{\mathsf{E}}
\newcommand{\M}[1]{\mathbf{#1}}
\newcommand{\Nul}{\mathcal{N}}

\newcommand{\R}{\mathbb{R}}
\newcommand{\T}{\top}
\newcommand{\Vm}{V^-}
\newcommand{\Vpm}{V^{\pm}}
\newcommand{\Vp}{V^+}
\newcommand{\Zplus}{\mathbb{Z}_{\geqslant{0}}}
\newcommand{\Z}{\mathbb{Z}}

\newcommand{\detrue}{\de_*}
\newcommand{\de}{\boldsymbol{\theta}}

\newcommand{\etab}{\boldsymbol{\eta}}
\newcommand{\e}{\mathrm{e}}

\newcommand{\kappab}{\boldsymbol{\kappa}}
\newcommand{\lambpm}{\lambda_{\pm}}
\newcommand{\locus}[1]{\mathcal{E}_{#1}}
\newcommand{\obs}[1]{\hat{#1}}

\newcommand{\pib}{\boldsymbol{\pi}}
\newcommand{\pro}[2][P]{\M{#1}_{#2}^\perp}

\newcommand{\truede}{\detrue}
\newcommand{\ud}{\,\mathrm d}
\newcommand{\ve}[1]{\mathbf{#1}}
\newcommand{\xib}{\boldsymbol{\xi}}

\newcommand{\estAML}{\widehat{\de}_{\mathrm{AML}}}
\renewcommand{\estAML}{\widehat{\de}}

\newcommand{\Lambdab}{\boldsymbol{\Lambda}}
\newcommand{\cov}[2][]{\Lambdab_{#2}^{#1}}

\DeclareMathOperator*{\psf}{psf}
\DeclareMathOperator*{\prf}{prf}
\DeclareMathOperator*{\argmin}{arg\,min}
\DeclareMathOperator{\arccot}{arccot}

\definecolor{ml-col}{RGB}{5,137,252}
\definecolor{dir-points-col}{RGB}{179,179,179}
\definecolor{dir-region-col}{RGB}{255, 222 , 89}

\makeatletter
\def\multiunderbracex#1#2{\mathop{\vtop{\m@th\ialign{##\crcr
   $\hfil\displaystyle{#2}\hfil$\crcr
   \noalign{\kern3\p@\nointerlineskip}%
   #1\crcr\noalign{\kern3\p@}}}}\limits}

\def\multiupbracefilla{$\m@th \setbox\z@\hbox{$\braceld$}%
  \bracelu\leaders\vrule \@height\ht\z@ \@depth\z@\hfill 
\kern\p@\vrule \@width\p@\kern\p@\vrule \@width\p@\kern\p@\vrule \@width\p@
$}

\def\multiupbracefillb{$\m@th \setbox\z@\hbox{$\braceld$}%
\vrule \@width\p@\kern\p@\vrule \@width\p@\kern\p@\vrule \@width\p@\kern\p@
 \leaders\vrule \@height\ht\z@ \@depth\z@\hfill\bracerd
  \braceld\leaders\vrule \@height\ht\z@ \@depth\z@\hfill
\kern\p@\vrule \@width\p@\kern\p@\vrule \@width\p@\kern\p@\vrule \@width\p@
$}

\def\multiupbracefillc{$\m@th \setbox\z@\hbox{$\braceld$}%
\vrule \@width\p@\kern\p@\vrule \@width\p@\kern\p@\vrule \@width\p@\kern\p@
\leaders\vrule \@height\ht\z@ \@depth\z@\hfill
\kern\p@\vrule \@width\p@\kern\p@\vrule \@width\p@\kern\p@\vrule \@width\p@
$}

\def\multiunderbraced{\multiunderbracex\multiupbracefill}
\def\multiupbracefill{$\m@th \setbox\z@\hbox{$\braceld$}%
\vrule \@width\p@\kern\p@\vrule \@width\p@\kern\p@\vrule \@width\p@\kern\p@
 \leaders\vrule \@height\ht\z@ \@depth\z@\hfill\braceru$}

 \def\multiunderbracee{\multiunderbracex\multiupbracefille}
 \def\multiupbracefille{$\m@th \setbox\z@\hbox{$\braceld$}%
  \bracelu\leaders\vrule \@height\ht\z@ \@depth\z@\hfill 
\m@th \setbox\z@\hbox{$\braceld$}%
\vrule \@width\p@ %\kern\p@\vrule \@width\p@\kern\p@\vrule \@width\p@\kern\p@  \leaders\vrule 
\leaders\vrule 
\@height\ht\z@ \@depth\z@\hfill \bracerd
  \braceld\leaders\vrule \@height\ht\z@ \@depth\z@\hfill
\kern\p@\vrule \@width\p@\kern\p@\vrule \@width\p@\kern\p@\vrule \@width\p@
$}
\makeatother
\newlength{\mytextsize}

\begin{document}

\title{Determining Ellipses from Low-Resolution Images with a Comprehensive Image Formation Model}

\author{\IEEEauthorblockN{Wojciech Chojnacki \protect\orcidicon{0000-0001-7782-1956} and Zygmunt L. Szpak \protect\orcidicon{0000-0002-0694-4622}} \\
\IEEEauthorblockA{Australian Institute for Machine Learning,
The University of Adelaide, SA 5005, Australia\\
Email: \{wojciech.chojnacki, zygmunt.szpak\}@adelaide.edu.au}
}

% \author[1]{Wojciech Chojnacki}
% \author[1,*]{Zygmunt L. Szpak}

% \affil[1]{Australian Institute for Machine Learning, The University of Adelaide, SA 5005, Australia}

% \affil[*]{Corresponding author: zygmunt.szpak@adelaide.edu.au}

%% To be edited by editor
% \dates{Compiled \today}

%\ociscodes{(100.2000)   Digital image processing; (100.2960)   Image analysis; (150.1135)   Algorithms; (350.5730)   Resolution.}

%% To be edited by editor
 %\doi{}
% \doi{\url{http://dx.doi.org/10.1364/XX.XX.XXXXXX}}

%\setboolean{displaycopyright}{true}

%\begin{document}

\maketitle

\begin{abstract}
  When determining the parameters of a parametric planar shape based on a single low-resolution image, common estimation paradigms lead to inaccurate parameter estimates. The reason behind poor estimation results is that standard estimation frameworks fail to model the image formation process at a sufficiently detailed level of analysis.  We propose a new method for estimating the parameters of a planar elliptic shape based on a single photon-limited, low-resolution image.  Our technique incorporates the effects of several elements---point-spread function, discretisation step, quantisation step, and photon noise---into a single cohesive and manageable statistical model.  While we concentrate on the particular task of estimating the parameters of elliptic shapes, our ideas and methods have a much broader scope and can be used to address the problem of estimating the parameters of an arbitrary parametrically representable planar shape.  Comprehensive experimental results on simulated and real imagery demonstrate that our approach yields parameter estimates with unprecedented accuracy.  Furthermore, our method supplies a parameter covariance matrix as a measure of uncertainty for the estimated parameters, as well as a planar confidence region as a means for visualising the parameter uncertainty.  The mathematical model developed in this paper may prove useful in a variety of disciplines which operate with imagery at the limits of resolution.
\end{abstract}


\graphicspath{{./}{../figs/}}
%\graphicspath{./}{../}
%\graphicspath{{./TMP/}}

\section{Introduction}
\label{sec:introduction}

\begin{figure*}[!tb]
  \centering
  \begin{subfloat}
    \centering
    \includegraphics[scale = 0.32]{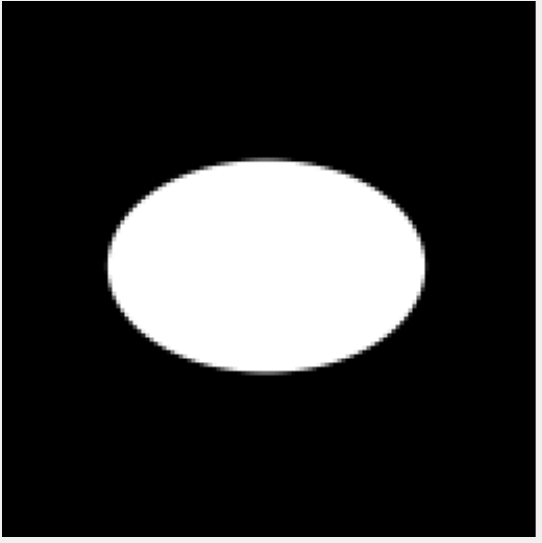}
  \end{subfloat}
  \begin{subfloat}
    \centering
    \includegraphics[scale = 0.32]{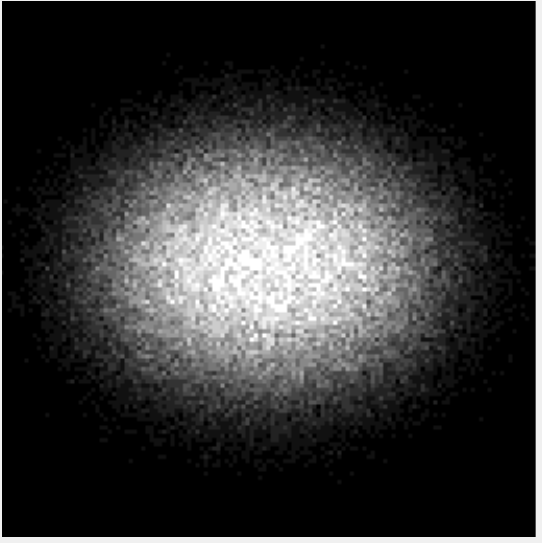}
  \end{subfloat}
  \begin{subfloat}
    \centering
    \includegraphics[scale = 0.32]{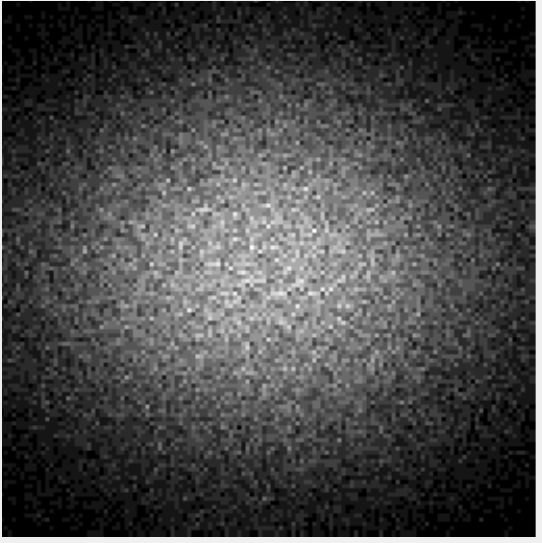}
  \end{subfloat}
  \begin{subfloat}
    \centering
    \includegraphics[scale = 0.32]{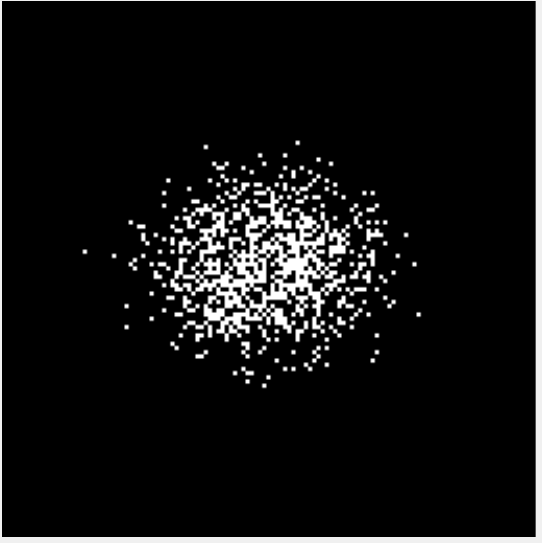}
  \end{subfloat}
  \begin{subfloat}
    \centering
    \includegraphics[scale = 0.32]{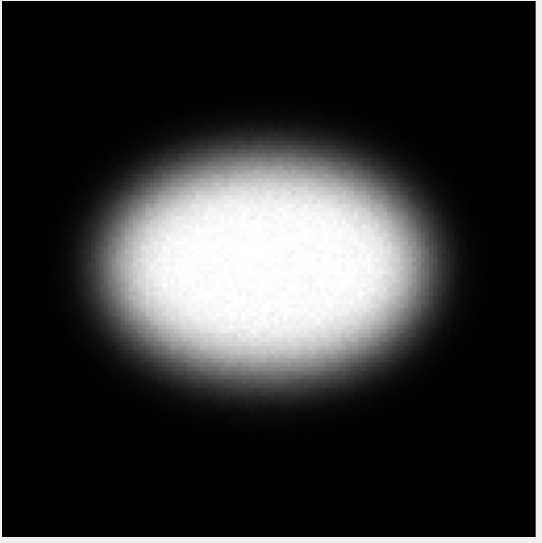}
  \end{subfloat} \\
  \vspace{0.1in}
    \begin{subfloat}
    \centering
    \includegraphics[scale = 0.32]{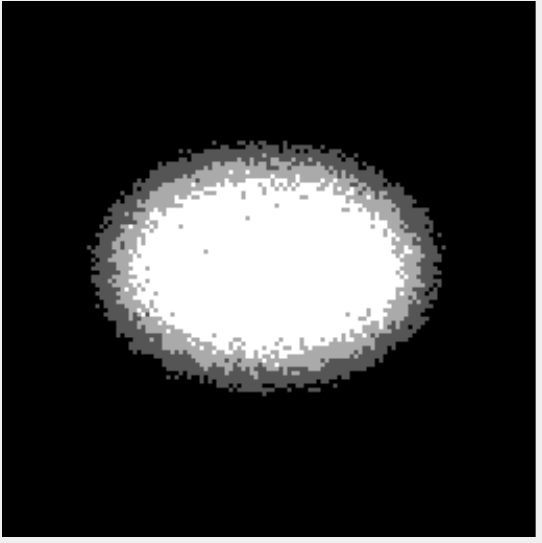}
  \end{subfloat}
  \begin{subfloat}
    \centering
    \includegraphics[scale = 0.32]{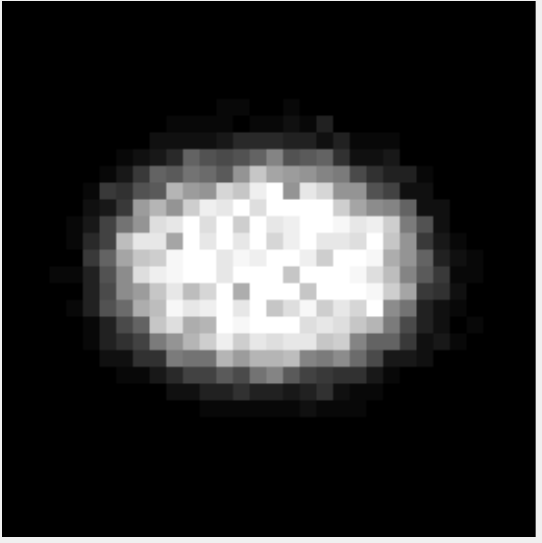}
  \end{subfloat}
  \begin{subfloat}
    \centering
    \includegraphics[scale = 0.32]{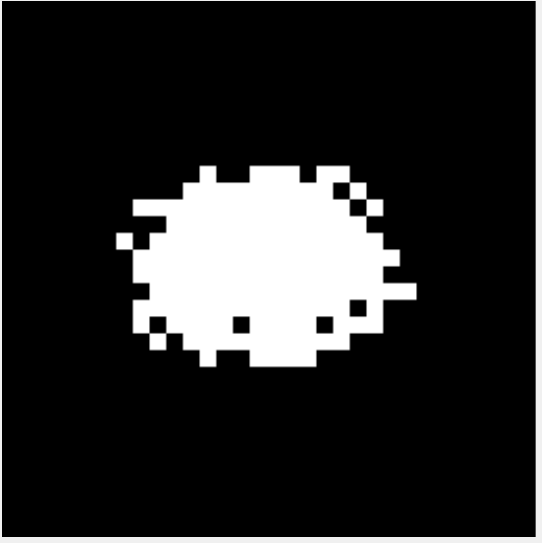}
  \end{subfloat}
  \begin{subfloat}
    \centering
    \includegraphics[scale = 0.32]{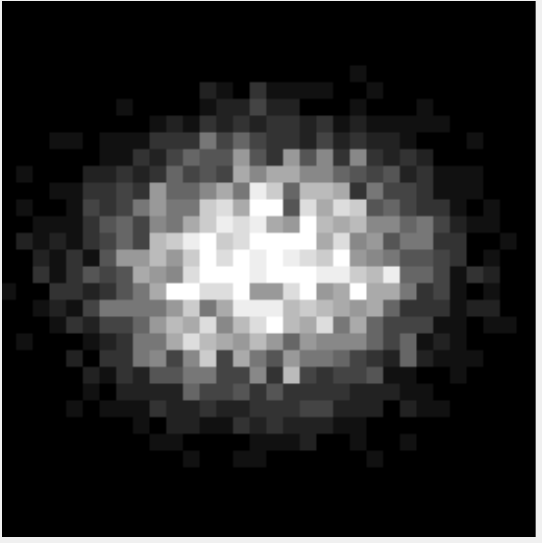}
  \end{subfloat}
  \begin{subfloat}
    \centering
    \includegraphics[scale = 0.32]{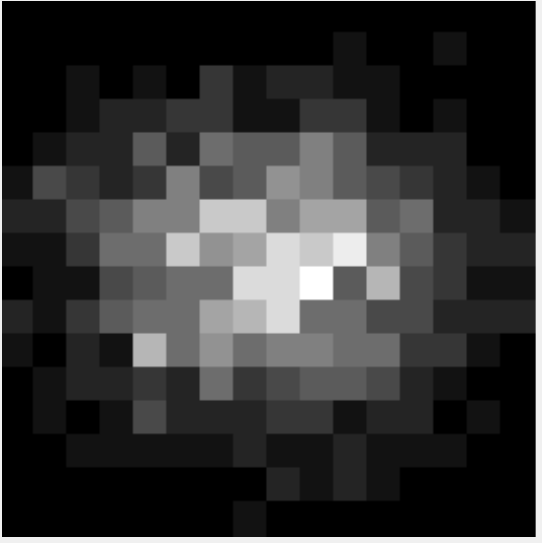}
  \end{subfloat} \\
   \vspace{0.1in}
   \hspace{0.05in}
    \begin{subfloat}
    \centering
    \includegraphics[scale = 0.32]{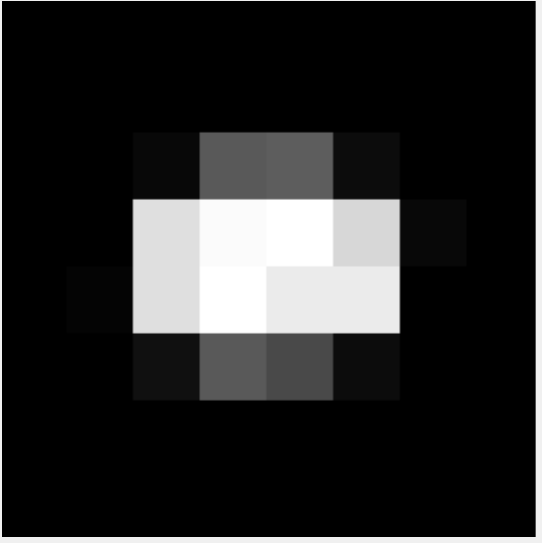}
  \end{subfloat}
  \begin{subfloat}
    \centering
    \includegraphics[scale = 0.32]{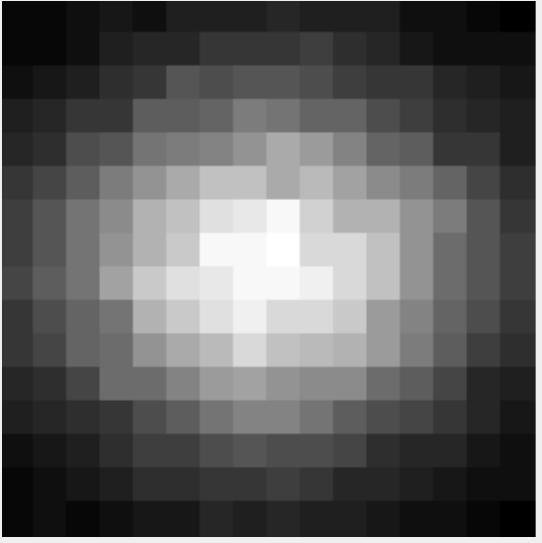}
  \end{subfloat}
  \begin{subfloat}
    \centering
    \includegraphics[scale = 0.32]{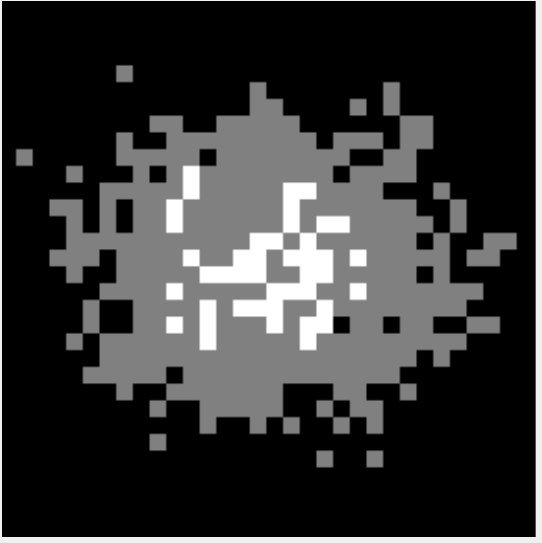}
  \end{subfloat}
  \begin{subfloat}
    \centering
    \includegraphics[scale = 0.32]{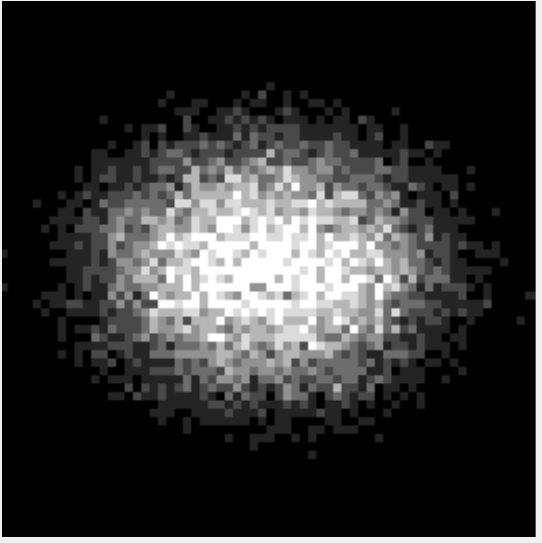}
  \end{subfloat}
  \begin{subfloat}
    \centering
    \includegraphics[scale = 0.32]{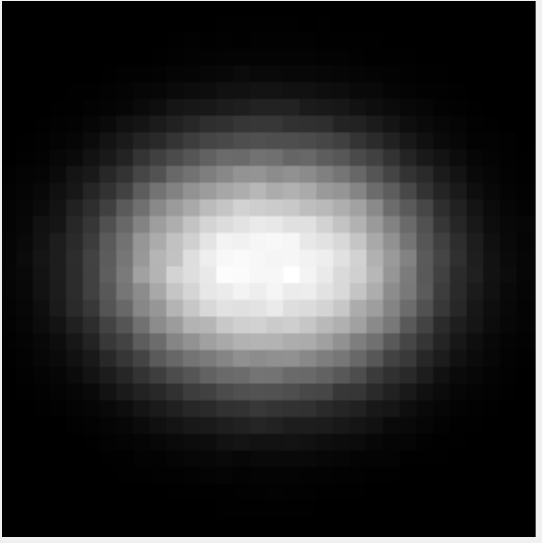}
  \end{subfloat}
  \caption{Examples of different digital images of the same elliptic region. The top-left image denotes an ideal digital image of an elliptic region. All the other images are observations of this image
  with different point-spread functions at varying levels of resolution, noise level, and quantisation. Our aim is to recover the ellipse parameters associated with the ideal digital image (top-left) given its corrupted observation.}
  \label{fig:montage}
\end{figure*}

\IEEEPARstart{W}{e} present a method for recovering the parameters of a planar elliptic shape from a low-resolution, photon-limited digital image. Our procedure provides unparalleled parameter estimation accuracy. We develop a systematic but manageable statistical model of the image formation process that distinguishes our approach from contemporary methods. Our model accounts for the point-spread function (PSF), the inherent continuous-to-discrete mapping of the image formation process, as well as the uncertainty due to quantisation and photon noise.  \Cref{fig:montage} illustrates an example of the diverse images that our model accommodates. While our paper focuses on the particular task of estimating the parameters of elliptic shapes, the ideas and methods formulated in this article have a much broader scope and can be used to address the problem of estimating the parameters of an arbitrary parametrically representable planar shape. Determining the parameters of an ellipse from a low-resolution planar image has essential applications in camera calibration \cite{mariyanayagam18:_pose}, but the core of our contribution lies in the details of the mathematical framework and conceptual methodology. We believe that our general approach is worthy of imitation and may lead to substantial progress in confocal microscopy, long-range surveillance, high-accuracy camera calibration, and astronomy.

\section{Related work}
\label{sec:related-work}

The majority of ellipse estimation methods fit a curve to a planar set of points.  One distinguishes between point-based ellipse fitting methods by considering the nature of the cost function that the algorithms minimise.  Methods which explicitly decrease the distance between the points and the ellipse curve are considered geometric methods.  The quintessential geometric method is orthogonal distance regression, which minimises the orthogonal distance from a point to the curve \cite{anh:_least_squares_orthogonal_fitting, zhengyou_parameter_1997, kim02:_orthellipse, gander_least-squares_1994, sturm_conicfitting, chernov11_least_squares_quadratic_curve, chernov14:_fittin}. Algebraic methods, on the other hand, try to ensure that the data points satisfy an ellipse implicit equation as accurately as possible. One differentiates between algebraic methods by considering how they penalise the degree to which a data point fails to satisfy an implicit equation \cite{rosin93, al-sharadqah12, kukush04:_consis, hunyadi13:_const}.  Algebraic methods, in particular, have been the focus of considerable study, and recent works have concentrated on improving their statistical efficacy and accuracy \cite{kanatani_ellipse_2006:hyper,szpak2012comparison,kanatani12:_renor_retur, collett2014ellipse}.

The fact that the most advanced ellipse fitting methods operate on data points is problematic when one wishes to fit an ellipse to a photon-limited low-resolution image of an elliptic region. The difficulty lies in extracting a set of data points that precisely lie on the contour of the ellipse. The standard approach involves gradient estimation, non-maxima suppression, and thresholding. These steps produce, at best, a set of data points that approximate the contour only at the level of resolution of the pixel grid.  Moreover, each of these steps introduces substantial errors and biases which the noise models of prevailing ellipse fitting methods disregard.

It is possible to obtain data points with sub-pixel coordinates by using sub-pixel edge detection methods \cite{fabijanska2015subpixel}.  However, sub-pixel techniques usually do not characterise the uncertainty or bias of their estimates, and so one cannot attribute meaningful covariance matrices to the sub-pixel data points. The inability to characterise the uncertainty and bias of the sub-pixel points is a severe limitation and, effectively, violates the modelling assumptions associated with point-based ellipse fitting methods.  Moreover, the sub-pixel estimation methods are themselves sensitive to noise and do not take quantisation into account.

As an alternative to point-based ellipse fitting, Ouellet and H{\'e}bert \cite{Ouellet2008} proposed a region-based method.  The region-based method exploits a duality relationship between a point and a line in projective geometry.  The duality relationship states that the homogeneous representation of a point may simultaneously be interpreted as a description of a line.  Hence, one can equivalently state the problem of fitting an ellipse to a set of points, as the task of fitting an ellipse to a set of lines.  The particular set of lines that satisfy an ellipse equation are called the envelope of tangent lines.  The region-based method takes advantage of the observation that lines perpendicular to the gradient of an ideal ellipse image are tangent to the ellipse.  The method proceeds by computing the gradient of an image and discarding pixels where the magnitude of the gradient is below a specified threshold.  For each remaining pixel, the method constructs a line that is perpendicular to the orientation of the gradient. The algorithm then minimises an algebraic error, which is similar to the well-known point-based direct ellipse fit (DEF) \cite{fitzgibbon99:_direc, halir:_numer}.  An important difference is that each line contributes to the algebraic cost by a weight equal to the magnitude of the gradient from which the line was derived.

The limitations mentioned for the point-based ellipse fitting methods also hold for the aforementioned region-based method. This particular region-based method does not model the image formation process, and so cannot accommodate Poisson noise or quantisation in a principled manner. Furthermore, the method still operates at the level of pixels and so does not address the resolution problem.

We address the shortcomings of these established ellipse fitting methods by developing a new estimation framework which incorporates an intricate but tractable model of the image formation process. 
  
\section{Image formation process}
\label{sec:image-form-proc}

\begin{figure}[!tb]
\centering
  \includegraphics[scale = 1]{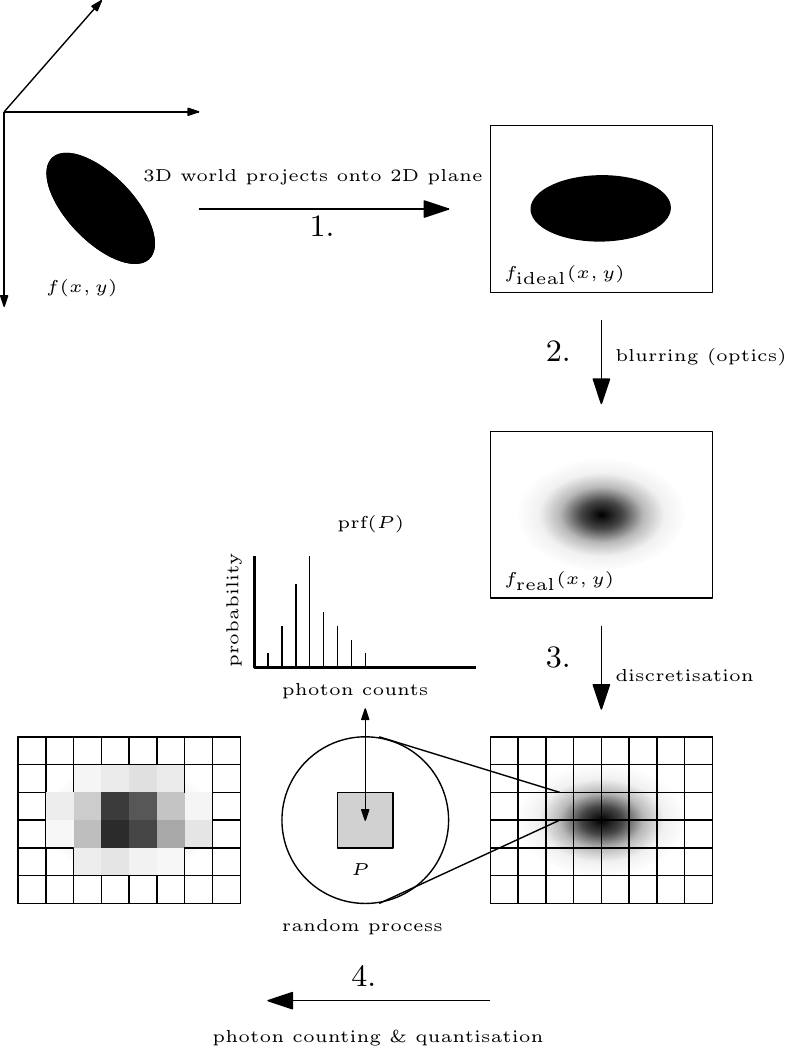}
  \caption{An overview of an image formation model. In the first step, the 3D world is projected onto a 2D image plane resulting in an ideal geometric image. The second stage accounts for imprecisions such as geometric distortions and blurring. The third step imposes a grid of pixels, introduces a statistical model for the number of photons that hit each pixel, and accounts for noise. The final step models the loss of information due to quantisation of the photon counts.}
 \label{fig:image-formation}
\end{figure}

% When describing the  image formation process, it is  helpful to regard
% the  imaged planar  elliptic shape  as a  scalar-valued function  on a
% two-dimensional Euclidean space, where the scalar values represent the
% reflected optical flux per unit area \cite{barrett2004foundations}. In
% this paper we shall consider a  uniform white elliptic shape against a
% black or grey  background. Let $D$ be a planar  region in $\R^2$ whose
% boundary is an ellipse given by
% \begin{displaymath}
%   p(x,y) = 0,
% \end{displaymath}
% where
% \begin{displaymath}
%   p(x,y) = ax^2 + bxy + cy^2 + dx + ey + f
% \end{displaymath}
% with $a$, $b$,  $c$, $d$, $e$, $f$ real numbers. 
% Our planar elliptic shape can then be represented by
% \begin{displaymath}
%   f(x,y) = 1_D(x,y). 
% \end{displaymath}
% Here, for a given set $E$, $1_E$ denotes the characteristic function of $E$.

One can conceptualise an imaging system as a continuous-to-discrete operator which maps a function of continuous variables (the elliptic shape in our 3D world) to a finite set of numbers (the discrete image) \cite{barrett2004foundations}. The information loss in the passage from the continuous domain to the digital one occurs in four stages (see \cref{fig:image-formation}).  First, the 3D world is projected onto a 2D plane using one of several projection methods available (perspective, fish-eye, catadioptric, etc.).  The projection process produces an analogue image with infinite resolution, or an ideal geometric image.  The geometric image is an idealisation---any real optical device (e.g., a camera lens) imposes certain imprecisions such as geometric distortions and blurring.  The second stage models the effect of the errors and leads to the real analogue image.  This image still resides in the analogue domain and is not directly observable.  The third and fourth stage, discretisation plus addition of noise (including photon counting noise, electronic noise and quantisation round-off noise), finally produces an observable digital image.

To keep the transition from the ideal to the real analogue image tractable, it is standard to assume that the amount of blurring does not vary within the field of view, and as such can be modelled by a convolution of the ideal image with a single function \cite{pratt13:_introd_digit_image_proces}. This function is known as the PSF of the image acquisition device.  With the PSF at hand, the relation between the ideal and real analogue images can be written as
\begin{equation}
  \label{eq:1}
  f_{\mathrm{real}}(x,y)
  = (f_{\mathrm{ideal}} * \psf)(x,y),
\end{equation}
where $*$ denotes convolution in the plane \cite{koethe08:_what_can_we_learn_discr}.  As it turns out, the PSF can in practice be well approximated by a Gaussian kernel,
\begin{equation}
  \label{eq:2}
  \mathrm{psf}(x,y)
  =
  \frac{1}{2 \pi \sigma_{\mathrm{PSF}}^2}
  \exp \left( - \frac{x^2 + y^2}{2 \sigma_{\mathrm{PSF}}^2} \right)
\end{equation}
with some positive $\sigma_{\mathrm{PSF}}$. We shall adopt this approximation in our discussion.

The discretisation stage is reflective of the fact that the image values are constant over each pixel from a grid of pixels, and that they are multiples of a single numerical value.  The latter effect is explained by the image intensity at a particular pixel being proportional to the number of photo-electrons recorded by the pixel's sensor.  Instrumental in the passage from the real analogue to the digital image is a pixel response function:
\begin{equation}
  \label{eq:3}
  \prf(P) = \frac{1}{|P|} \int_P f_{\mathrm{real}}(x,y) \ud x \ud y,
\end{equation}
where, for a pixel $P$, $|P|$ denotes the area of $P$. We shall regard
the image  intensity at the  pixel $P$  as a random  value fluctuating
around $\prf(P)$.

\section{Probabilistic model}
\label{sec:probmodel}

Let $C$ be the conversion factor linking the image intensity with the photo-electron count. If the image intensity is a number between, say, $0$ and $1$, then the corresponding value of the photo-electron count is an integer between $0$ and $C$.  Neglecting---temporarily---the digitisation error, it is natural to model the photo-electron count $N_P$ at $P$ stochastically by applying a Poisson noise to $\prf(P)$, i.e., by letting
\begin{equation}
  N_P = X_P,
\end{equation}
where  $X_P$  is  a  Poisson-distributed  random  variable  with  mean
$C \prf(P)$,
\begin{equation}
  \label{eq:4}
  \Pr(X_P=k) = \frac{(C \prf(P))^k}{k!} \e^{-k
    C \prf(P)}
\end{equation}
for $k = 0, 1, 2, \dots$.

To include the quantisation error, we modify the above definition and add to $X_P$ an integer-valued random variable $U_P$ uniformly distributed in the range
\begin{equation}
  [-b, b] \cap \Z = \{-b, -b+1, \dots, -1, 0, 1, \dots, b-1, b\},
\end{equation}
where $b$ is a non-negative integer.  In other words, we let
\begin{equation}
  \label{eq:5}
  N_P = X_P + U_P.
\end{equation}
Our proposed image recovery method will be based on the statistical model of a pixel value embodied in the above formula. To proceed, it will be critical to identify the form of the pixel response function and the shape of the probability distribution of $N_P$.

\section{Pixel response function}
\label{sec:pixel-resp-funct}

We now provide a computationally convenient expression for the pixel response function for an image of a uniform white planar shape.  We consider two scenarios whereby the shape appears against (1) a completely black backdrop and (2) a grey backdrop.

\subsection{Black background}
\label{sec:black-background}

Let $D$ be a uniform white region on a black planar background.  Then the image associated with $D$ can be described as
\begin{equation}
  f_{\mathrm{ideal}}(x,y) = 1_{D}(x,y),
\end{equation}
where $1_D$ stands for the characteristic function of $D$.  In view of \eqref{eq:1},
\begin{equation}
  \label{eq:6}
  f_{\mathrm{real}}(x,y)
  =
  \int_{\R^2}1_{D}(x-s,y-t) \psf(s,t) \ud s \ud t.
\end{equation}
 For $(s,t) \in \R^2$, let $T_{(s,t)}D$ denote the \emph{translate} of $D$ by $(s,t)$,
\begin{equation}
  T_{(s,t)}D = \{(x,y) \in \R^2
  \mid
  (x - s, y -t) \in D \}.
\end{equation}
It is readily checked that
\begin{equation}
  1_D(x-s,y-t) = 1_{T_{(s,t)}D}(x,y).
\end{equation}
With this in mind, for a given pixel $P$, we have 
\begin{multline}
  \label{eq:7}
  \int_P \left[ \int_{\R^2}1_{D}(x-s,y-t) \psf(s,t) \ud s \ud
    t \right] \ud x \ud y
  \\
  \begin{aligned}[b]
  & =
  \int_P \left[ \int_{\R^2}1_{T_{(s,t)} D}(x,y) \psf(s,t) \ud s \ud
    t \right] \ud x \ud y
  \\
  & =
   \int_{\R^2} \left[ \int_P 1_{T_{(s,t)} D}(x,y) \ud x \ud
    y \right]  \psf(s,t) \ud s \ud t
  \\
  & =
    \int_{\R^2} \left| P \cap T_{(s,t)} D \right|  \psf(s,t) \ud s
    \ud t
  \\
  & =
    \int_{\R^2} \left|  T_{(-s,-t)} P \cap D  \right|  \psf(s,t) \ud s
    \ud t,
  \end{aligned}
\end{multline}
where the identity
\begin{equation}
   \left| P \cap T_{(s,t)} D \right|  = \left|  T_{(-s,-t)} P \cap D  \right|  
 \end{equation}
 results from 
 \begin{equation}
   T_{(-s,-t)} (P \cap T_{(s,t)} D) =  T_{(-s,-t)} P \cap D 
 \end{equation}
 and the fact that the area of $P \cap T_{(s,t)} D$ is the same as the
 area  of  the  translate  $T_{(-s,-t)}   (P  \cap  T_{(s,t)}  D)$  of
 $P     \cap     T_{(s,t)}     D$     by     $(-s,-t)$.      Combining
 \cref{eq:3,eq:6,eq:7}, we finally obtain
 \begin{equation}
   \label{eq:8}
   \prf(P) = \frac{1}{|P|}
   \int_{\R^2} \left|  T_{(-s,-t)} P \cap D  \right|  \psf(s,t) \ud s
   \ud t.
 \end{equation}

 \subsection{Grey background}
\label{sec:grey-background}

Let $D$ be a uniform white region on a grey planar background. Suppose that the background has intensity $c$, where $0 \leq c < 1$. Then the image associated with $D$ can be described as
\begin{equation}
  f_{\mathrm{ideal}}(x,y)
  = 1_{D}(x,y) + c 1_{\R^2 \setminus D}(x,y)
\end{equation}
or, equivalently, as
\begin{equation}
  f_{\mathrm{ideal}}(x,y) = (1 - c) 1_{D}(x,y) + c.
\end{equation}
It then immediately follows from  \eqref{eq:8} that the pixel response
function in this case is given by
\begin{equation}
  \label{eq:9}
  \prf(P) = c + \frac{1 -c}{|P|}
   \int_{\R^2} \left|  T_{(-s,-t)} P \cap D  \right|  \psf(s,t) \ud s
   \ud t.
 \end{equation}
 The above formula will play a key role in the subsequent development.  Note that \eqref{eq:8} is a particular case of \eqref{eq:9} with $c = 0$.

 \section{Probability distribution}

 We now calculate the probability distribution of the random variable $N_P$ defined in \eqref{eq:5}.

 Suppose that $X$ and $Y$ are two independent random variables, with $X$ being Poisson-distributed with parameter $\lambda$, and $Y$ being discrete uniformly distributed over the set
\begin{math}
  [-b, b] \cap \Z,
\end{math}
\begin{equation}
  X \sim \mathit{Poisson}(\lambda),
  \quad
  Y \sim \mathit{DU}(-b,b).
\end{equation}
Then, for each $n \in [-b, \infty) \cap \Z = \{-b, b+1, \dots \}$, the event $X + Y = n$ is the union of the pairwise disjoint events $\{X + Y = n\}$ and $\{Y = m\}$, where $m$ runs over $[-b, b] \cap \Z$.  It follows that
\begin{equation}
  \Pr(X + Y = n)
  =
  \sum_{m=-b}^b \Pr(X + Y = n, \, Y = m).
\end{equation}
Since
\begin{equation}
  \begin{aligned}[b]
    \Pr(X + Y = n, \, Y = m)
    & =
    \Pr(X = n - Y, \, Y = m)
    \\
    & = \Pr(X = n - m, \, Y = m)
    \\
    & = \Pr(X = n - m) \Pr(Y = m),
  \end{aligned}
\end{equation}
where the last equality holds by the independence of $X$ and $Y$, and
\begin{align}
  \Pr(X = n - m) & = \e^{-\lambda}\frac{\lambda^{n-m}}{(n-m)!}
    1_{[0,\infty)}(n-m),
    \\
    \Pr(Y = m) & = \frac{1}{2b + 1},
\end{align}
we have
\begin{equation}
  \begin{aligned}[b]
    \Pr(X + Y = n)
    & =
    \frac{1}{2b+1}\sum_{m=-b}^b \e^{-\lambda}
    \frac{\lambda^{n-m}}{(n-m)!}
    1_{[0,\infty)}(n-m)
    \\
    & =
    \frac{1}{2b+1}\sum_{m=-b}^b \e^{-\lambda}
    \frac{\lambda^{n+m}}{(n+m)!}
    1_{[0,\infty)}(n+m).
  \end{aligned}
\end{equation}
Since $1_{[0,\infty)}(n+m) = 0$ whenever $m < -n$, the dummy integer-valued variable $m$ in the last sum has to satisfy two conditions: $-b \leq m \leq b$ and $n + m \geq 0$ or, equivalently, $-n \leq m$.  These conditions may conveniently be combined into a single condition, namely, $\max(-b,-n) \leq m \leq b$ or, what is the same, $-\min(b,n) \leq m \leq b$, with the proviso that $-\min(b,n) \leq b$; if $-\min(b,n) > b$, we necessarily have $\Pr(X + Y = n) = 0$. Thus
% \begin{equation}
%   \label{eq:10}
%   \Pr(X + Y = n) =
%   \begin{cases}
%     %\displaystyle
%     \dfrac{1}{2b+1}\sum_{m=-\min(b,n)}^b \e^{-\lambda}
%     \dfrac{\lambda^{n+m}}{(n+m)!}
%     &
%     \text{if $-\min(b,n) \leq b$,}
%     \\
%     0 & \text{otherwise}.
%   \end{cases}
% \end{equation}
%
% \begin{equation}
%   \label{eq:11}
%   \Pr(X + Y = n) =
%   \begin{cases}
%     \begin{alignedat}{2}
%       &\dfrac{1}{2b+1} &\smash[b]{\sum_{m=-\min(b,n)}^b} \e^{-\lambda}
%       &\dfrac{\lambda^{n+m}}{(n+m)!}
%       \\[1ex]
%       &&&\text{if $-\min(b,n) \leq b$,}
%       \\
%       & 0 
%       && \text{otherwise.}
%    \end{alignedat}
%  \end{cases}
% \end{equation}
%
\begin{multline}
  \label{eq:12}
  \Pr(X + Y = n) 
  \\
  =
  \begin{cases}
    %\displaystyle
    \dfrac{1}{2b+1}\sum_{m=-\min(b,n)}^b \e^{-\lambda}
    \dfrac{\lambda^{n+m}}{(n+m)!}
    &
    \text{if $-\min(b,n) \leq b$,}
    \\
    0 & \text{otherwise}.
  \end{cases}
\end{multline}
We immediately infer from this formula that if $b = 0$, then
\begin{equation}
  \Pr(X + Y = n) = \e^{-\lambda}  \frac{\lambda^n}{n!} = \Pr(X)
\end{equation}
for each $n \in \Zplus \colonequals \{0, 1, 2, \dots\}$.

\begin{figure}[!tb]
\centering
  \includegraphics[scale = 0.6]{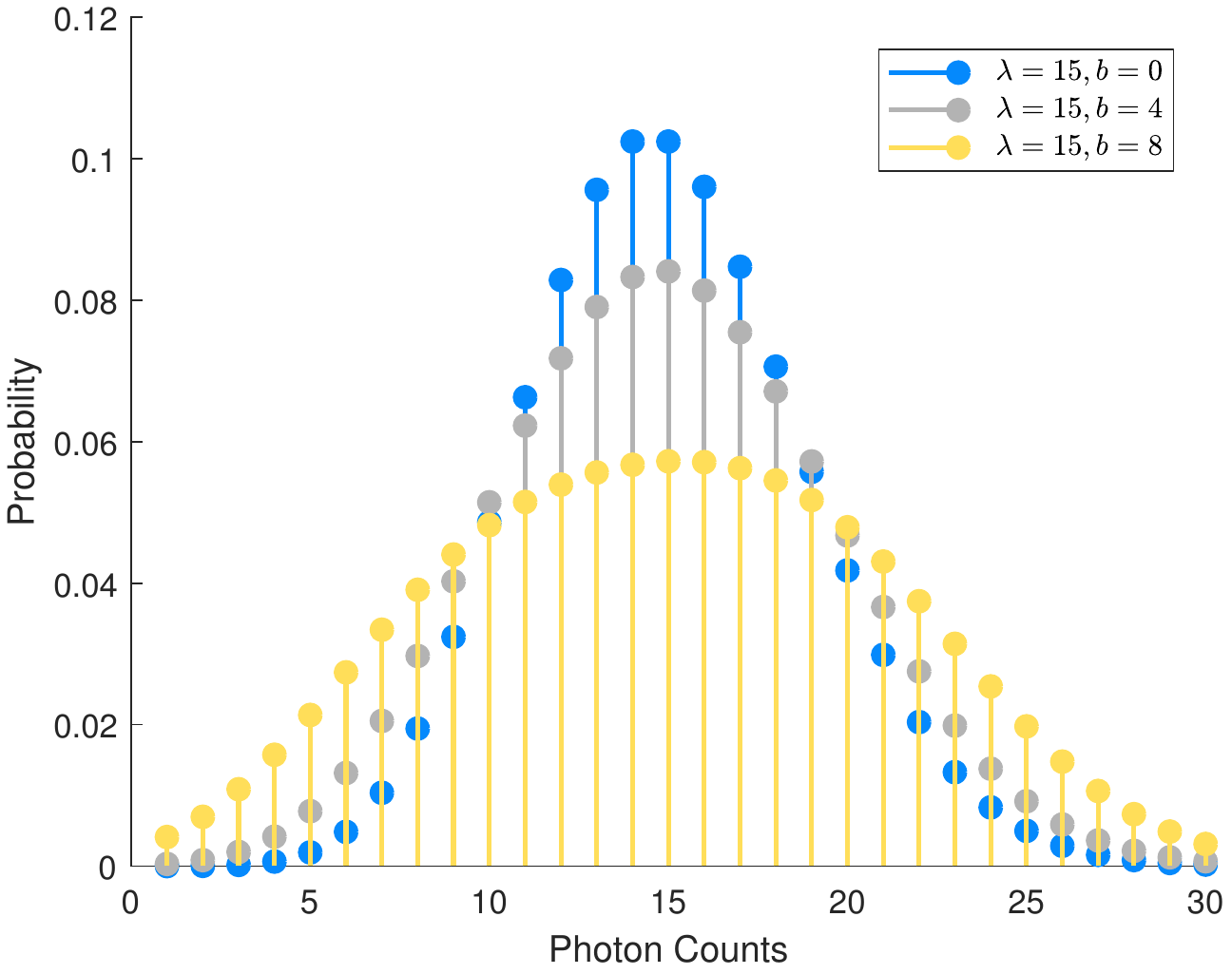}
  \caption{A visualisation of the probability mass function of quantised photon counts given by equations \cref{eq:12,eq:17}.}
 \label{fig:poissonUniformProbability}
\end{figure}

Suppose that $b > 0$. Then if $n < -b$, then $\min(b,n) = n$ and so $-\min(b,n) > b$, implying, according to \eqref{eq:12}, that
\begin{equation}
  \label{eq:13}
  \Pr(X + Y = n)  = 0.
\end{equation}
If $-b \leq n \leq b$, then $\min(b,n) = n$, and we have
\begin{equation}
  \label{eq:14}
  \begin{aligned}[b]
  \Pr(X + Y = n) & =
  \frac{1}{2b+1}\sum_{m=-n}^b \e^{-\lambda}
  \frac{\lambda^{n+m}}{(n+m)!}
  \\
  & =
  \frac{1}{2b+1}\sum_{k=0}^{n+b} \e^{-\lambda}
  \frac{\lambda^k}{k!}.
\end{aligned}
\end{equation}
If $n > b$, then $\min(b,n) = b$, and we have
\begin{equation}
  \label{eq:15}
  \begin{aligned}[b]
    \Pr(X + Y = n)
    & =
    \frac{1}{2b+1}\sum_{m=-b}^b \e^{-\lambda}
    \frac{\lambda^{n+m}}{(n+m)!}
    \\
    & =
    \frac{1}{2b+1}\sum_{k=n-b}^{n+b} \e^{-\lambda}
    \frac{\lambda^k}{k!}
    \\
    & =
    \frac{1}{2b+1}\left(\sum_{k=0}^{n+b} \e^{-\lambda}
      \frac{\lambda^k}{k!}
      -
      \sum_{k=0}^{n-b} \e^{-\lambda}
      \frac{\lambda^k}{k!}
    \right).   
  \end{aligned}
\end{equation}
Using the upper incomplete gamma function
\begin{equation}
  \Gamma(\alpha, x) = \int_x^{\infty} \e^{-t} t^{\alpha-1} \ud t
\end{equation}
and the formula
\begin{equation}
  \Gamma(n+1,x) = n! \sum_{m=0}^n \e^{-x} \frac{x^m}{m!}
\end{equation}
for $n \in \Zplus$, we can rewrite \Crefrange{eq:13}{eq:15} as
% \begin{equation}
%   \label{eq:16}
%   \Pr(X + Y = n) =
%   \begin{cases}
%     0 & \text{if $n < -b$,}
%     \\[3ex]
%     %\displaystyle
%     \dfrac{\Gamma(n + b +1, \lambda)}{(2b+1)(n+b)!} &
%     \text{if $-b \leq n \leq b$,}
%     \\[3ex]
%     %\displaystyle
%     \dfrac{1}{2b+1}
%     \left(
%     \dfrac{\Gamma(n - b +1, \lambda)}{(n+b)!}
%     -
%     \dfrac{\Gamma(n - b +1, \lambda)}{(n+b)!}
%     \right)
%     &
%     \text{if $n > b$.}
%   \end{cases}
% \end{equation}

\begin{multline}
  \label{eq:17}
  \Pr(X + Y = n) 
  \\
  =
  \begin{cases}
    0 & \text{if $n < -b$,}
    \\[3ex]
    % \displaystyle
    \dfrac{\Gamma(n + b +1, \lambda)}{(2b+1)(n+b)!} &
    \text{if $-b \leq n \leq b$,}
    \\[3ex]
    % \displaystyle
    \begin{multlined}
      \shoveleft{
        \dfrac{1}{2b+1}
        \left(
          \dfrac{\Gamma(n + b +1, \lambda)}{(n+b)!}
        \right.
        \\
        \left.
          -
          \dfrac{\Gamma(n - b +1, \lambda)}{(n-b)!}
        \right)}
    \end{multlined}
    &
    \text{if $n > b$.}
  \end{cases}
\end{multline}
An illustration of this probability mass function is presented in \cref{fig:poissonUniformProbability}.

\section{Estimation method}
\label{sec:estimation-method}

Suppose that our region of interest, say the interior of an ellipse, is a member of a family of candidate regions $\{D_{\xib} \mid \xib \in \boldsymbol{\Xi} \}$ indexed by a vector of parameters $\xib$ running over a parameter space $\boldsymbol{\Xi}$.  Let $\xib_*$ be the vector of parameters determining the region of interest.  For each $\xib \in \boldsymbol{\Xi}$, suppose that $D_{\xib}$ is transformed via a physical process into a digital image.  Let $\mathcal{P}$ denote the set of corresponding pixels.  For each pixel $P$ in $\mathcal{P}$, we assume that the image intensity $f_P$ of $P$ is modelled as a random variable $N_P$ given by \eqref{eq:5}.  Moreover, we assume that the $N_P$, $P \in \mathcal{P}$, are stochastically independent. This means that with
\begin{equation}
  N = (N_P)_{P \in \mathcal{P}}
  \quad
  \text{and}
  \quad
  f_{\mathrm{digital}} = (f_P)_{P \in \mathcal{P}},
\end{equation}
we have
\begin{equation}
  \Pr(N = f_{\mathrm{digital}})
  =
  \prod_{P \in \mathcal{P}} \Pr(N_P = f_P).
\end{equation}
The expression on the right hand side depends on the parameters $\xib$, $b$, $c$, $C$, and $\sigma_{\mathrm{PSF}}$.  More explicitly, we have, in accordance with \eqref{eq:17},
% \begin{displaymath}
%   \Pr(N_P = f_P)
%   =
%   \begin{cases}
%     0 & \text{if $n < -b$,}
%     \\[3ex]
%     %\displaystyle
%     \dfrac{\Gamma(f_P + b +1, \lambda_P)}{(2b+1)(f_P +b)!} &
%     \text{if $-b \leq f_P  \leq b$,}
%     \\[3ex]
%     %\displaystyle
%     \dfrac{1}{2b+1}
%     \left(
%     \dfrac{\Gamma(f_P - b +1, \lambda_P)}{(f_P+b)!}
%     -
%     \dfrac{\Gamma(f_P - b +1, \lambda_P)}{(f_P +b)!}
%     \right)
%     &
%     \text{if $f_P > b$,}
%   \end{cases}
% \end{displaymath}
% %
\begin{multline}
  \Pr(N_P = f_P)
  \\
  =
  \begin{cases}
    0 & \text{if $n < -b$,}
    \\[3ex]
    %\displaystyle
    \dfrac{\Gamma(f_P + b +1, \lambda_P)}{(2b+1)(f_P +b)!} &
    \text{if $-b \leq f_P  \leq b$,}
    \\[3ex]
    % \displaystyle
    \begin{multlined}
      \shoveleft{
        \dfrac{1}{2b+1}
        \left(
          \dfrac{\Gamma(f_P + b +1, \lambda_P)}{(f_P+b)!}
        \right.
        \\
        \left.
          -
          \dfrac{\Gamma(f_P - b +1, \lambda_P)}{(f_P -b)!}
        \right)
      }
    \end{multlined}
    &
    \text{if $f_P > b$,}
  \end{cases}
\end{multline}
where, in line with \cref{eq:4,eq:9},
\begin{equation}
  \label{eq:18}
  \lambda_P = C \left(c + \frac{1 -c}{|P|}
   \int_{\R^2} \left|  T_{(-s,-t)} P \cap D_{\xib} \right|  \psf(s,t) \ud s
            \ud t
            \right)
\end{equation}
depends on $\xib$, $c$, $C$, and, in view of \eqref{eq:2}, also on $\sigma_{\mathrm{PSF}}$. We treat $b$, $c$ and $C$ as values known a priori and fixed.  Writing, more emphatically, $\Pr(N = f_{\mathrm{digital}})$ as $\Pr(f_{\mathrm{digital}} \mid \xib, \sigma_{\mathrm{PSF}})$, we may treat $(\xib, \sigma_{\mathrm{PSF}}) \mapsto \Pr(f_{\mathrm{digital}} \mid \xib, \sigma_{\mathrm{PSF}})$ as the likelihood function for $\xib$ and $\sigma_{\mathrm{PSF}}$.  Using the maximum likelihood (ML) principle, we may next estimate $\xib$ and $\sigma_{\mathrm{PSF}}$, given an observed digital image $f_{\mathrm{observed}} = (f^{\mathrm{ob}}_{P})_{P \in \mathcal{P}}$, by minimising the corresponding negative log-likelihood function
\begin{equation}
  \label{eq:19}
  \ell(\xib, \sigma_{\mathrm{PSF}})
  \colonequals
  - \sum_{P \in \mathcal{P}} \ln \Pr(N_P = f^{\mathrm{ob}}_{P}).
\end{equation}
In other words,
\begin{equation}
  \{\widehat{\xib}, 
  \widehat{\sigma}_{\mathrm{PSF}}\}
  =
  \argmin_{\xib,  \sigma_{\mathrm{PSF}}}
  \ell(\xib, \sigma_{\mathrm{PSF}}).
\end{equation}
We are solely interested in obtaining an estimate of the region of concern, so once the minimisation is performed, we may discard $\widehat{\sigma}_{\mathrm{PSF}}$ and the remaining $\widehat{\xib}$ is then an estimate of $\xib_*$.  We refer to $\widehat{\xib}$ as the \emph{ML estimate} of $\xib_*$, and dub the method of generating $\widehat{\xib}$ the \emph{ML estimator} for region estimation.

The above-described estimation method requires, critically, a means for calculating the term $\left| T_{(-s,-t)} P \cap D_{\xib} \right|$ in \eqref{eq:18}.  As it turns out, the evaluation of $\left| T_{(-s,-t)} P \cap D_{\xib} \right|$ can be performed effectively in the case where $D_{\xib}$ is an ellipse (with $\xib$ the vector of the ellipse's parameters).  This is due to the fact that there exist explicit formulae for determining the area of intersection between an ellipse and a rectangle (a pixel) in the case that the sides of the rectangle are parallel to the semi-axes of the ellipse.  We present these rather involved formulae in Appendix A
%\cref{area-intersection-ellipse-rectangle}
and henceforth concentrate our discussion exclusively on ellipse estimation.

\subsection{Characterising an ellipse region}
\label{characterising-an-ellipse-region}

An ellipse in general position can be expressed parametrically as
\begin{equation}
\begin{aligned}[b]
  x & = H + A\cos{\alpha}\cos{\tau} - B\sin{\alpha}\sin{\tau},\\
  y & = K + A\cos{\alpha}\sin{\tau} + B\sin{\alpha}\cos{\tau}.
\end{aligned}
\end{equation}
Here, $A$ and $B$ represent the length of the semi-major and semi-minor axis of the ellipse, $H$ and $K$ denote the $x$ and $y$ coordinates of the centre of the ellipse, $\tau$ is the angle formed by the major axis with the positive $x$ axis, and $\alpha$ is the angular co-ordinate of the point $(x,y)$ on the ellipse.  The vector $\xib = [A,B,H,K,\tau]^\T$ (excluding $\alpha$) encompasses the geometric parameters of the ellipse and uniquely describes the ellipse as a set. Alongside the parametric form, the ellipse can be represented in Cartesian form as the locus of points $(x,y)$ in the plane satisfying
\begin{equation}
  \label{eq:20}
  ax^2+ bxy + cy^2 + dx + ey + f = 0
\end{equation}
where $a,b,c,d,e,f$ are real numbers such that $b^2 < 4ac$. The vector $\de = [a,b,c,d,e,f]^\T$ of the algebraic parameters of the ellipse determines the ellipse uniquely, however the reverse correspondence is not univocal---all non-zero multiples of $\de$ describe one and the same ellipse.  Using the Cartesian form, the interior of the ellipse can be conveniently characterised as the locus of points $(x,y)$ in the plane satisfying
\begin{equation}
  ax^2+ bxy + cy^2 + dx + ey + f < 0.
\end{equation}

The above two descriptions of an ellipse are fully equivalent, each being obtainable from the other by means of a conversion formula. The explicit formulas for conversion will be of relevance in what follows. The rule for the passage from the geometric parameters to the algebraic parameters is given by
\begin{equation}
  \label{eq:21}
  \begin{split}
    a &= \frac{\cos^2 \tau}{A^2} + \frac{\sin^2 \tau}{B^2}, \\
    b & = \left(\frac{1}{A^2} - \frac{1}{B^2} \right) \sin{2\tau}, \\
    c & = \frac{\cos^2 \tau}{B^2} + \frac{\sin^2 \tau}{A^2}, \\
    d & = \frac{2 \sin{\tau} \left (K \cos {\tau} - H
        \sin{\tau}\right)}{B^2}-\frac{2 \cos^2 \tau\left(H+K
        \tan{\tau}\right)}{A^2}, \\
    e & = \frac{2 \cos{\tau} \left(H \sin{\tau}-K
        \cos{\tau}\right)}{B^2}-\frac{2 \sin{\tau} \left(H \cos{\tau}+K
        \sin{\tau}\right)}{A^2}, \\
    f & = \frac{\left(H \cos{\tau}+K \sin{\tau}\right)^2}{A^2}
    +\frac{\left(K \cos{\tau}-H \sin{\tau}\right)^2}{B^2}-1.
  \end{split}
\end{equation}
To present the rule for the passage from the algebraic parameters into geometric parameters, we first let
\begin{equation}
\begin{aligned}[b]
  \Delta & = b^2  - 4ac,
  \\
  \lambpm
         & =
           \frac{1}{2}\left(a + c \mp \left(b^2+(a-c)^2\right)^{1/2}\right),
  \\
  \psi & =  bde - ae^2 - b^2f + c(4af - d^2),
  \\
  \Vpm & = \left(\frac{\psi}{\lambpm \Delta}\right)^{1/2},
\end{aligned}
\end{equation}
where $\pm$ and $\mp$ are shorthand for $+$ \emph{or} $-$, which allows presentation of two expressions in one formula, with the upper $-$ of $\mp$ associated with the $+$ of $\pm$.  We can now state the rule in question as
\begin{equation}
\begin{gathered}
  A = \max(\Vp,\Vm),
  \quad 
  B = \min(\Vp,\Vm), 
  %\displaybreak[3]
  \\
  H  = \frac{2cd- be}{\Delta}, \quad  K =  \frac{2ae - bd}{\Delta},
  %\displaybreak[3]
  \\ 
  \tau 
  = 
  \begin{cases}
    %% Vp >= Vm
    \frac{1}{2}\arccot\left(\frac{a-c}{b}\right)
    & \text{if $b < 0$, $a < c$  and $\Vp \ge \Vm$}, 
    %\displaybreak[3]
    \\
    \frac{\pi}{4}
    & \text{if $b < 0$, $a = c$  and $\Vp \ge \Vm$},
    %\displaybreak[3]
    \\
    \frac{1}{2}\arccot\left(\frac{a-c}{b}\right) + \frac{\pi}{2}
    & \text{if $b < 0$, $a > c$ and $\Vp \ge \Vm$},
    %\displaybreak[3]
    \\
    0
    & \text{if $b = 0$, $a < c$  and $\Vp \ge \Vm$},
    %\displaybreak[3]
    \\
    \frac{\pi}{2}
    & \text{if $b = 0$, $a \ge c$ and $\Vp \ge \Vm$},
    %\displaybreak[3]
    \\
    \frac{1}{2}\arccot\left(\frac{a-c}{b}\right) + \pi
    & \text{if $b > 0$, $a < c$  and $\Vp \ge \Vm$},
    %\displaybreak[3]
    \\
    \frac{3\pi}{4}
    & \text{if $b > 0$, $a = c$  and $\Vp \ge \Vm$},
    %\displaybreak[3]
    \\
    \frac{1}{2}\arccot\left(\frac{a-c}{b}\right) + \frac{\pi}{2}
    & \text{if $b > 0$, $a > c$  and $\Vp \ge \Vm$},
    %\displaybreak[3]
    \\
    %% Vp < Vm
    \frac{1}{2}\arccot\left(\frac{a-c}{b}\right) + \frac{\pi}{2}
    & \text{if $b < 0$, $a < c$  and $\Vp < \Vm$},
    %\displaybreak[3]
    \\
    \frac{3\pi}{4}
    & \text{if $b < 0$, $a = c$  and $\Vp < \Vm$},
    %\displaybreak[3]
    \\
    \frac{1}{2}\arccot\left(\frac{a-c}{b}\right) + \pi
    & \text{if $b < 0$, $a > c$ and $\Vp < \Vm$},
    %\displaybreak[3]
    \\
    \frac{\pi}{2} & \text{if $b = 0$, $a < c$  and $\Vp < \Vm$},
    %\displaybreak[3]
    \\
    0
    & \text{if $b = 0$, $a \ge c$ and $\Vp < \Vm$},
    %\displaybreak[3]
    \\
    \frac{1}{2}\arccot\left(\frac{a-c}{b}\right) + \frac{\pi}{2}
    & \text{if $b > 0$, $a < c$  and $\Vp < \Vm$},
    %\displaybreak[3]
    \\
    \frac{\pi}{4}& \text{if $b > 0$, $a = c$  and $\Vp < \Vm$},
    %\displaybreak[3]
    \\
    \frac{1}{2}\arccot\left(\frac{a-c}{b}\right)
    & \text{if $b > 0$, $a > c$  and $\Vp < \Vm$}
  \end{cases}
\end{gathered}
\end{equation}
(see \cite[Section 4.10.2]{zwillinger12:_crc} for the starting point of the derivation of the formulas). We remark that the formula for $\tau$ is valid only under the assumption that the ellipse is not a circle, i.e., provided the inequality $(a-c)^2 + b^2 > 0$ holds.

\subsection{Forming a digital image of an ellipse region}
\label{sec:digitalImageFormation}

\begin{figure}
  \centering
  \includegraphics[scale=1]{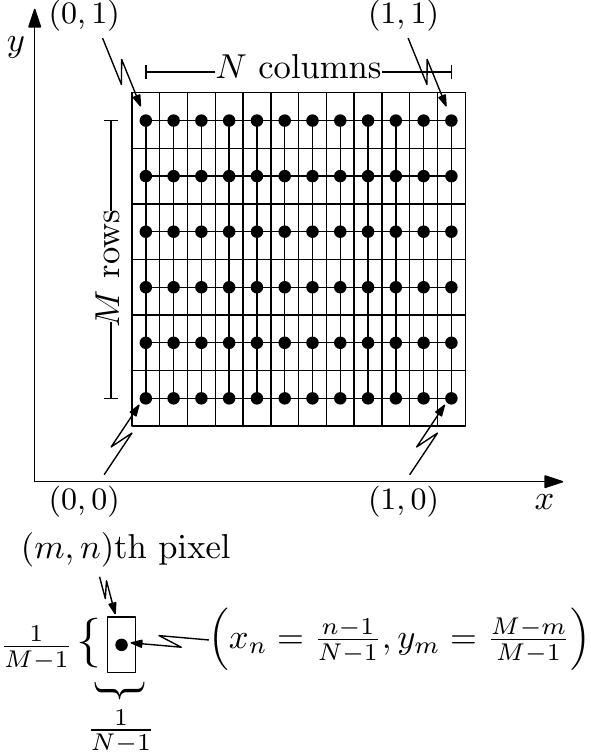}
  \caption{$M \times N$ array of pixels with the corresponding grid
    of pixel centres  spanning the unit box $[0,1]  \times [0,1]$. The
    diagram includes a formula for  converting the matrix indices of a
    pixel to the pixel's Cartesian coordinates.}
  \label{image-vs-pixel-coordinates-explanation}
\end{figure}

To be able to make use of the negative log-likelihood given in \eqref{eq:19}, one needs to have a way of constructing digital images of candidate ellipse regions that incorporate the effects of the PSF, discretisation step, quantisation step, and photon noise. In this section, we outline the procedure for constructing such images.

With a view to generating a single specific image, given a pair of integers $M$ and $N$ such that $M \geq 2$ and $N \geq 2$, we create a grid of pixels in the form of an $M \times N$ array of rectangles aligned with the $x$ and $y$ axes, each of size $(M - 1)^{-1} \times (N - 1)^{-1}$, with the centre $(x_n,y_m)$ of the $(m,n)$th rectangle specified by
\begin{equation}
  x_n  =  \frac{n-1}{N-1}
  \quad 
  \text{and}
  \quad
  y_m = \frac{M-m}{M-1}
\end{equation}
for every $1 \leq m \leq M$ and every $1 \leq n \leq N$ (see \cref{image-vs-pixel-coordinates-explanation}).  Given a particular ellipse specified by a parameter vector $\de$, we construct a digital image via the following steps.  We first determine the geometric parameters of the ellipse.  The relevant formulae are given in \cref{characterising-an-ellipse-region}.  Next, we take advantage of the fact that an application of the coordinate transformation
\begin{equation}
\begin{aligned}[b]
  x' & = (x - H) \cos \tau + (y - K) \sin \tau, \\
  y' & = - (x - H) \sin \tau + (y - K) \cos \tau
\end{aligned}
\end{equation}
brings the ellipse to a standard form.  More specifically, we apply the above transformation to each pixel centre, obtaining points $(x'_n,y'_m)$ ($m = 1, \dots, M$, $n = 1, \dots, N$).  For each pair $(m,n)$, we form a rectangle of size $(M - 1)^{-1} \times (N - 1)^{-1}$ centred at $(x'_n,y'_m)$ and aligned with the $x'$ and $y'$ axes, and calculate the area of intersection between this rectangle and the ellipse transformed to the $x'$-$y'$ coordinate system (this ellipse is uniquely determined by $A$ and $B$).  In our calculations we use formulae from Appendix A.
%\cref{area-intersection-ellipse-rectangle}.
Subsequently, we divide the intersection area by the area of the rectangle.  The outcome yields the value of a pixel-averaged ideal digital image at the $(n,m)$th pixel, $f_{\mathrm{averaged}}(x_n,y_m)$.  To incorporate the effect of the PSF, we implement, for a value $c$ of the background intensity, a discretised version of \eqref{eq:9} in the form
\begin{equation}
  \prf(x_n,y_m)
  = c + 
  \begin{multlined}[t]
  \frac{1-c}{Z}
  %\frac{1}{2 \pi \sigma_{\mathrm{PSF}}^2 Z}
  \sum_{s=1}^{M}\sum_{t=1}^{N}
  f_{\mathrm{averaged}}(x_s,y_t)
  \\
  \times
  \exp
    \left( - \frac{(x_s - x_n)^2 + (y_t - y_m)^2}{2
        \sigma_{\mathrm{PSF}}^2} 
    \right),
  \end{multlined}
\end{equation}
where $Z$ is  a normalisation constant given by
\begin{equation}
  Z
  =
  % \frac{1}{2 \pi \sigma_{\mathrm{PSF}}^2}
  \sum_{s=1}^{M} \sum_{t=1}^{N}
  \exp \left( - \frac{(x_s - x_n)^2 + (y_t - y_m)^2}{2
      \sigma_{\mathrm{PSF}}^2} 
  \right).
\end{equation}
The array $[ \prf(x_n,y_m)]_{1 \leq m \leq M, \, 1 \leq n \leq N}$
%
% , representing the pixel response function, %
%
has entries between $0$ and $1$.  Scaling each entry of this array by a conversion factor $C$ and simulating, for each pair $(m,n)$ independently, effects of Poisson noise with parameter $C \prf(x_n,y_m)$ with the aid of \Cref{alg:randomPoissonVariate}, we next obtain an array $[f_{\mathrm{Poisson}}(x_n,y_m)]_{1 \leq m \leq M, \, 1 \leq n \leq N}$ of plausible photon counts, or a Poisson-corrupted image.
%
% The resulting real image consists of unscaled image intensities
% ranging from zero to one. To convert the image into plausible photon
% counts we scale the unit normalised image intensities by a conversion
% factor $C$. In so doing we are able to introduce signal dependent
% \emph{Poisson noise} as per \eqref{eq:4} and
% \cref{alg:randomPoissonVariate}. 
%
Recall that the standard deviation of Poisson noise is equal to the square-root of the average number of events.  Hence, when applying Poisson noise to an image, the signal-to-noise ratio is equal to
\begin{equation}
\label{eq:snr}
  \textrm{SNR} =
  \frac{C \prf(x_n,y_m)}{\sqrt{C
      \prf(x_n,y_m)}} 
  = \sqrt{C \prf(x_n,y_m)}. 
\end{equation}
For a large choice of $C$, the signal-to-noise ratio will be significant, and the image will appear relatively noise free. Conversely, for small values of $C$, corresponding to low-light conditions, the noise will be much more prominent.  To model the quantisation step of the digital image formation process, we partition the Poisson-corrupted image into $G$ grey levels. The partitioning is achieved by grouping the intensities into discrete bins. Let $b$ denote the half-width of a bin. For modelling convenience, we shall assume that both $b$ and $C$ are powers of two, which ensures that the number of grey levels $G = C/(2b)$ is also a power of two. With the quantisation function
\begin{equation}
  q(x) = 
  \begin{cases}
    %% Vp >= Vm
    1
    & \text{if $ -\infty <x< 2b\times 1 $}, 
    \\
    2
    & \text{if $ 2b\times 1 \le x < 2b\times 2 $},  
    \\
    3
    & \text{if $ 2b\times 2 \le x < 2b\times 3 $},  \\
    \vdots \\
     G-1   & \text{if $ 2b\times (G-2) \le x <  2b\times (G-1) $},  \\
     G  & \text{if $2b\times (G-1) \le x < \infty$},
  \end{cases}
\end{equation}
the final digital image is given by the relation
\begin{equation}
 f_{\mathrm{digital}}{(x_n,y_m)} = 2b \times q\left( C f_{\mathrm{Poisson}}{(x_n,y_m)} \right) - b.
\end{equation} 
Our quantisation model can be interpreted as follows. The scale factor $C$ denotes the number of photons that would yield a maximum amount of charge in a pixel and produce the brightest intensity.  The interval from zero to $C$ is partitioned into sub-intervals (bins), and generally the observed photon count is replaced by the centre value of the interval into which the photon count falls. The last interval extends into positive infinity to capture the notion of saturation. A pixel is said to be saturated if at least $C$ photons have reached it. If the number of photons exceeds $C$, then any additional photons that reach the pixel will not be registered and hence effectively quantised to the same value as the maximum count $C$.

\begin{figure*}[tbh]
  \centering
  \subfloat[][]{
    \includegraphics[scale = 0.32]{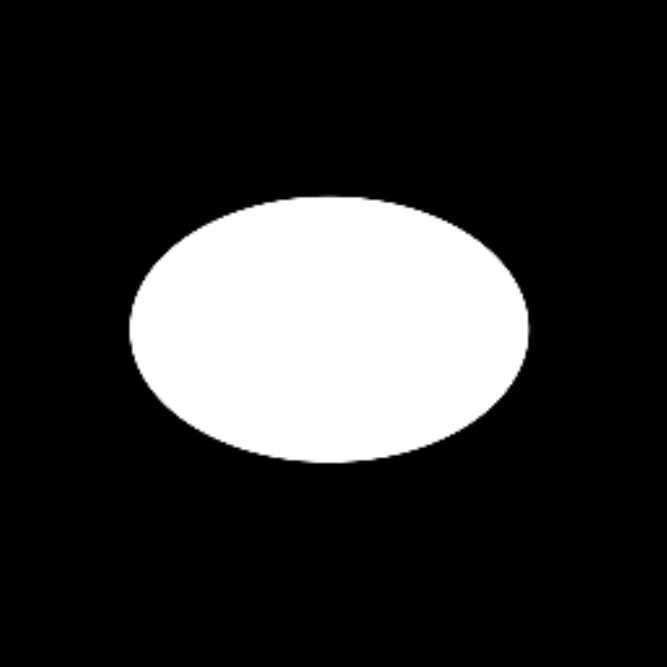}
    \label{example-analogue}
  }
  \hfil
  \subfloat[][]{
    \includegraphics[scale = 0.32]{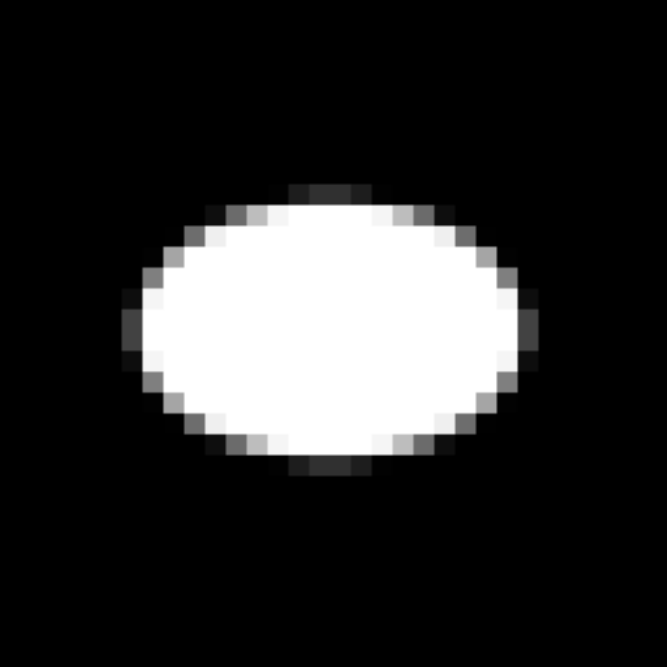}
    \label{example-discretisation}
  }
  \hfil
  \subfloat[][]{
    \includegraphics[scale = 0.32]{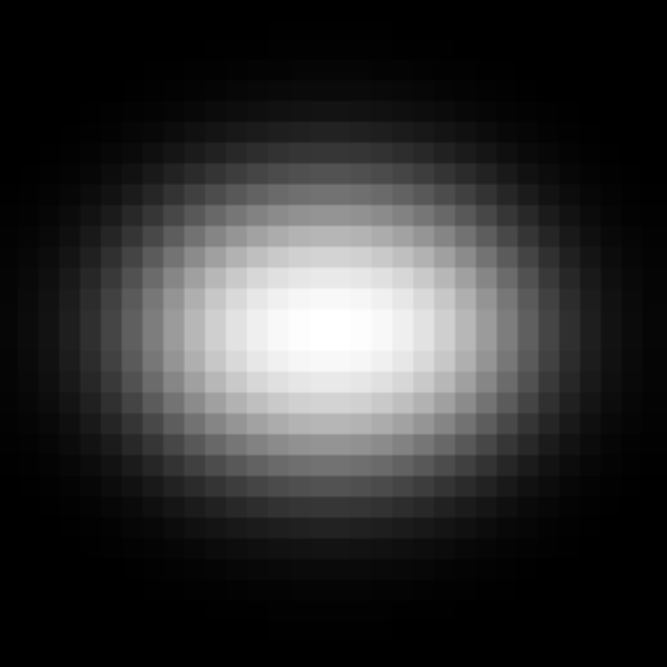}
    \label{example-psf}
  }
  \hfil
  \subfloat[][]{
    \includegraphics[scale = 0.32]{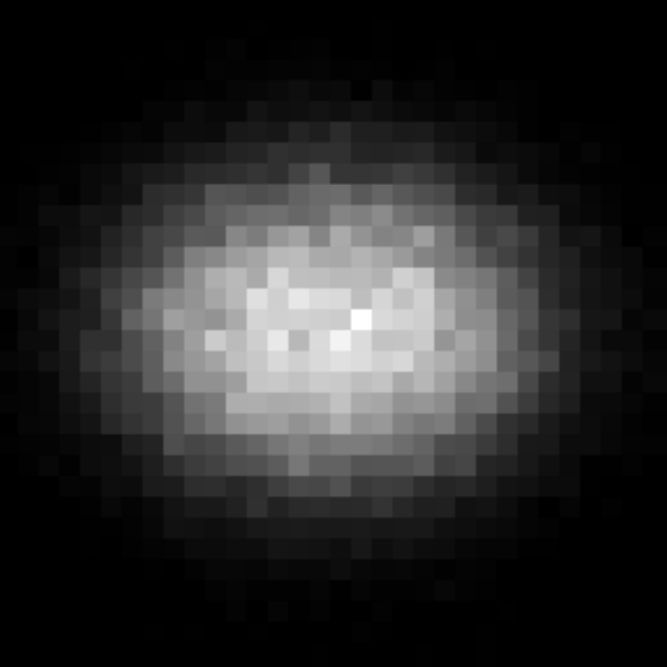}
    \label{example-poisson}
  }
  \hfil
  \subfloat[][]{
    \includegraphics[scale = 0.32]{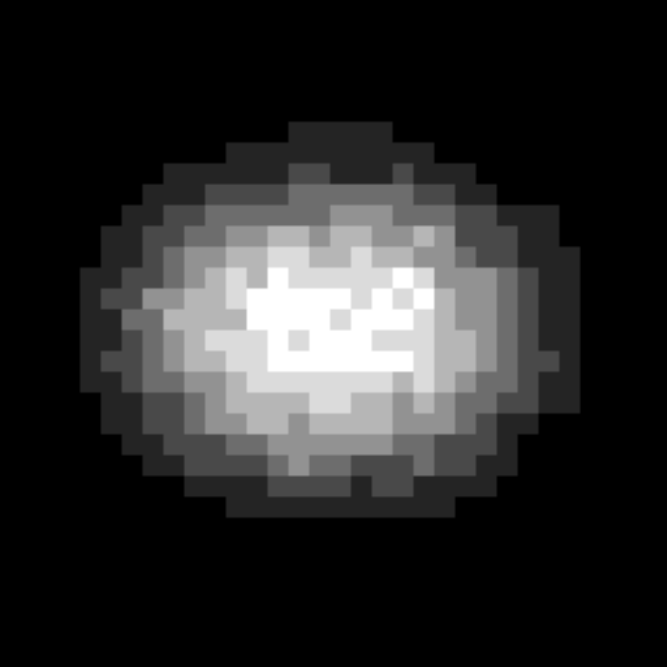}
    \label{example-quantised}
  }
  \caption{Example images detailing the stages of the image formation process. (a) The ideal (unobservable) analogue image.  (b) Consequence of discretisation.  (c) Consequence of discretisation and the point spread function. (d) Consequence of discretisation, the point spread function and Poisson noise. (e) Consequence of discretisation, the point spread function, Poisson noise and quantisation.}
  \label{fig:stages-of-image-formation}
\end{figure*}

An illustration of the different stages of the digital image formation process is presented in \cref{fig:stages-of-image-formation}.

% \begin{algorithm}
%   \caption{RandomPoissonVariate($\lambda$) \cite{knuth:1997}}
%   \label{alg:randomPoissonVariate}
%   \begin{algorithmic}[1]
%     \Require $\lambda >0$
%     \State $L \gets -\lambda$
%     \State $k \gets 0$
%     \State $p \gets  0$
%     \State condition $\gets$ True
%     \While{condition}
%      \State $k \gets k +1$
%      \State $p \gets p +  \log \left(\textsc{Rand} \right)$ \Comment{\textsc{Rand} generates a uniform number in the interval $[0,1]$.}
%      \State condition $\gets (p \ge L)$  \Comment{Inequality is either True or False.}
%     \EndWhile
%     \State \textbf{return} $k-1$ \Comment{Result is a random sample from a Poisson distribution with mean $\lambda$.}
%   \end{algorithmic}
% \end{algorithm}

\begin{algorithm}[t]
  \caption{RandomPoissonVariate($\lambda$) \cite{knuth:1997}}
  \label{alg:randomPoissonVariate}
  \begin{algorithmic}[1]
    \Require $\lambda >0$
    \State $L \gets -\lambda$
    \State $k \gets 0$
    \State $p \gets  0$
    \State condition $\gets$ True
    \While{condition}
    \State $k \gets k +1$
    \State $p \gets p +  \ln \left(\textsc{Rand} \right)$
    % \Comment{
    %   \begin{minipage}[t]{0.5\linewidth}
    %     \textsc{Rand} generates a uniform number in the interval $[0,1]$.
    %   \end{minipage}
    % }
    \Comment{\textsc{Rand} generates a uniform number in the interval $[0,1]$.}
    \State condition $\gets (p \ge L)$
    % \Comment{
    %   \begin{minipage}[t]{0.5\linewidth}
    %     Inequality is either True or False.
    %   \end{minipage}
    %   }
    \Comment{Inequality is either True or False.}
    \EndWhile
    \State \textbf{return} $k-1$
    % \Comment{
    %   \begin{minipage}[t]{0.7\linewidth}
    %     Result is a random sample from a Poisson distribution with mean $\lambda$.
    %   \end{minipage}
    % }
    \Comment{Result is a random sample from a Poisson distribution with mean $\lambda$.}
  \end{algorithmic}
\end{algorithm}

\subsection{Implementing the maximum likelihood estimator}
\label{sec:impl-maxim-likel}

A numerically stable implementation of formula (\ref{eq:19}) for the negative log-likelihood is presented in \Cref{alg:likelihoodDiscreteUniformPoisson} and \Cref{alg:logSumExp}. We minimise the negative log-likelihood using the BFGS Quasi-Newton method, and evaluate the required gradient and approximate Hessian numerically. Among the variables that we choose to parametrise the log-likelihood with in \Cref{alg:likelihoodDiscreteUniformPoisson} are real-valued variables labelled $\sqrt{A}$, $\sqrt{B}$, and $\sqrt{\sigma_{\mathrm{PSF}}}$.
%
% The squares of these variables are obviously meant to represent the lengths of the semi-major and semi-minor axes, and the value of the standard deviation of the point spread function, respectively.
%
During the optimisation process we square $\sqrt{A}$, $\sqrt{B}$, and $\sqrt{\sigma_{\mathrm{PSF}}}$, and ensure in that way that the values of $A$, $B$, and $\sigma_{\mathrm{PSF}}$ are non-negative.  To prevent $\sigma_{\mathrm{PSF}}$ from attaining the value of zero, we add a small positive constant $\epsilon$ to $(\sqrt{\sigma_{\mathrm{PSF}}})^2$. We assume that the scale factor $C$ and the quantisation factor $b$ are known or have been estimated from the data, and we do not optimise them further.  Thus our overall parameter vector is $\etab = [\sqrt{A}, \sqrt{B}, H,K, \tau,\sqrt{\sigma_{\mathrm{PSF}}}]^\T$, with the derived vector of geometric parameters $\xib = [(\sqrt{A})^2,(\sqrt{B})^2, H,K, \tau]^\T$.

\begin{algorithm}[t]
  \caption{LikelihoodDiscreteUniformPoisson($\etab$, $f_{\mathrm{digital}}$,$C$,$b$)}
  \label{alg:likelihoodDiscreteUniformPoisson}
  \begin{algorithmic}[1]
    \Require $\etab = [\sqrt{A},\sqrt{B}, H, K, \tau,\sqrt{\sigma_{\mathrm{PSF}}}]^\T$, $C \in \mathbb{N}$, $b \in \mathbb{N}$
    \State $\xib = [(\sqrt{A})^2,(\sqrt{B})^2, H,K, \tau]^\T$
    %\Comment{Square $A_s$ and $B_s$ from $\etab$ to ensure non-negativity} 
    \State $f_{\mathrm{digital}} \gets \textsc{DigitalImage}\left(\xib, C,b, \left(\sqrt{\sigma_{\mathrm{PSF}}}\right)^2 + \epsilon  \right)$
    % \newline\hspace*{0cm}\hfill
    % \begin{flushright}
    \Comment{
      % \begin{minipage}[t]{0.8\linewidth}
      Form a candidate digital image as per \cref{sec:digitalImageFormation}.
      % \end{minipage}
    }
    % \end{flushright}
    % \Comment{ \begin{varwidth}[t]{\linewidth}  Form a digital image \par as per \cref{sec:digitalImageFormation}.\end{varwidth}}
    \State $l \gets 0$ 
    \For{$m \gets 1\ldots M$}
    \For{$n \gets 1 \ldots N$}
    \State $v \gets f_{\mathrm{observed}}{(x_n,y_m)}$
    % \newline \hspace*{0cm} \hfill 
    \Comment{The variable $f_{\mathrm{observed}}{(x_n,y_m)}$ is the observed digital image.}
    \State $\lambda \gets \max\left(f_{\mathrm{digital}}{(x_n,y_m)}, \epsilon \right)$
    \Comment{Precaution to avoid taking the $\log$ of zero.}
    \For{$w \gets -\min(b,v) \ldots b$}
    \State
    % \begin{varwidth}[t]{\linewidth}     		 	
    \begin{math}
      \begin{multlined}[t]
        z\left[w+\min(b,v)\right] \gets
        \\
        -\lambda +
        (v+w) \ln \lambda
        % \Big(
        %   \textsc{GammaLn}(\lambda + 1)
        % \\
        %   {} - \textsc{GammaLn}(\lambda)
        % \Big)
        %\\
        -\textsc{GammaLn}(w+v+1)
      \end{multlined}
    \end{math}
    % \newline \hspace*{0cm} \hfill 
    \Comment{
      % \begin{varwidth}[t]{\linewidth}
      Use  the natural logarithm   of  the  gamma function.
      % \end{varwidth}
    } 
    % \end{varwidth}
    \EndFor
    \State $ l \gets l - \ln (2b+1) + \textsc{LogSumExp}(z)$
    \Comment{Apply the log-sum-exp trick  for numerically stable evaluation of the sum of exponentials.} 
    \EndFor
    \EndFor
    \State \textbf{return} $-l$
    \Comment{Result is the negative log-likelihood.}
  \end{algorithmic}
\end{algorithm}

\begin{algorithm}[t]
  \caption{LogSumExp($z$)}
  \label{alg:logSumExp}
  \begin{algorithmic}[1]
    \Require length-N array $z$ %with entries that are the logarithm of exponentials
    \State $a \gets \max \{z[n] \colon n = 1\ldots N\}$
    \Comment{Take the largest entry.}
    \State $s \gets 0$
    \For{$n \gets 1 \ldots N$}
    \State $s \gets s +  \exp \left( z[n] - a \right) $
    \Comment{$\ln(\sum_{n}^{N}\e^{z_n}) = a + \ln(\sum_{n}^{N}\e^{z_n-a})$.}
    \EndFor     	  
    \State \textbf{return} $a + \ln s$ 
  \end{algorithmic}
\end{algorithm}

\subsection{Characterising the uncertainty of the estimate}
\label{sec:char-uncert-estim}

To characterise the uncertainty or reliability of the ML estimate $\widehat{\xib}$ of the geometric parameters of the region's bounding ellipse, we use the covariance matrix $\M{\Lambda}_{\widehat{\xib}}$ of $\widehat{\xib}$.  We calculate the latter by exploiting the covariance matrix $\M{\Lambda}_{\widehat{\etab}}$ of the maximum likelihood estimate $\widehat{\etab}$.  Taking into account that the covariance matrix of a ML estimate is approximately equal to the inverse Hessian of the negative log-likelihood at the estimate, we let
\begin{equation}
  \M{\Lambda}_{\widehat{\etab}}
  =
  \left(
    \left[
      \frac{\partial^2 \ell(\etab, b, C)}
      {\partial \etab \partial \etab^\T}
    \right]_{\etab = \widehat{\etab}}
  \right)^{-1}
\end{equation}
(see \cite[Section~3.2]{sivia96:_data_analy}).  Next, applying the rule of covariance propagation, we propagate $\M{\Lambda}_{\widehat{\etab}}$ through the transformation $\etab \mapsto \xib$ to obtain
\begin{equation}
  \M{\Lambda}_{\widehat{\xib}}
  = \left[\partial_{\etab}\xib \right]_{\etab =  \widehat{\etab}} \M{\Lambda}_{\widehat{\etab}}
  \left(\left[\partial_{\etab}\xib\right]_{\etab =  \widehat{\etab}}\right)^\T.
\end{equation}
The Jacobian matrix of $\etab \mapsto \xib$ is explicitly given by
\begin{equation}
  \partial_{\etab}\xib =
  \begin{bmatrix}
    2\sqrt{A} & 0 & 0 & 0 & 0 & 0\\
    0 & 2\sqrt{B}& 0 & 0 & 0 & 0 \\
    0& 0 & 1 & 0 & 0 & 0\\
    0& 0 & 0 & 1 & 0 & 0\\
    0& 0 & 0 & 0 & 1 & 0\\
  \end{bmatrix}.
\end{equation}

\subsubsection{Visualising a planar confidence region}
\label{sec:visu-plan-conf}

The reliability of $\widehat{\xib}$ can alternatively be expressed in terms of a confidence region.  One typically constructs a confidence region of a parameter vector estimate as a portion of the parameter space that contains the correct parameter vector with a given high probability. But since the parameter space for the totality of all ellipses is five-dimensional,  a canonical confidence region is difficult to visualise and interpret. Hence, we formulate a more visually appealing form of an ellipse-specific confidence region, namely a confidence region in the plane.  Such an area is meant to cover the in-plane locus of the actual ellipse with a specified high probability.

The first to consider planar confidence regions for ellipse fits was Porrill \cite{porrill90:_fittin_kalman}.  Our approach is inspired by Scheff\'e's $S$-method for constructing simultaneous confidence bands for linear regression \cite{scheffe53}, \cite[Sect.  9.4--5]{lehmann05:_testin_statis_hypot}, and we have previously used it to establish planar confidence regions for a point-based ellipse fitting method \cite{szpak2015guaran_ellip_uncert}.  The construction that we set forth exploits the algebraic parameters of the ellipse and in particular involves the covariance matrix of an algebraically expressed ML estimate of the ellipse. To obtain a meaningful expression for such a matrix, it is mandatory to eliminate a redundant indeterminate scale of algebraic parameters \cite{kanatani01:_gauges,triggs99:_bundl}. We proceed with scale elimination by imposing the normalisation constraint $\| \de \| = 1$.  Let $\kappab$ denote the mapping $\xib \mapsto \de$ defined in \cref{eq:21} and let $\pib$ denote the normalisation transformation $\de \mapsto \| \de \|^{-1} \de$. We take $\widehat{\de} = \pib(\kappab(\widehat{\xib}))$ for the algebraically expressed ML estimate, with $\pib$ here guaranteeing that $\widehat{\de}$ is unit-normalised. Applying the rule of covariance propagation, we find that the concomitant algebraic covariance matrix $\M{\Lambda}_{\widehat{\de}}$ is given by
\begin{equation}
  \M{\Lambda}_{\widehat{\de}}
  =
  \left[\partial_{\de}\pib \right]_{\de = \kappab(\widehat{\xib})}
  \left[\partial_{\xib}\kappab \right]_{\xib = \widehat{\xib}}
  \M{\Lambda}_{\widehat{\xib}}
  \left(
    \left[\partial_{\xib}\kappab \right]_{\xib = \widehat{\xib}}
  \right)^\T
  \left(
    \left[\partial_{\de}\pib \right]_{\de =  \kappab(\widehat{\xib})}
    \right)^\T. 
\end{equation}
The Jacobian matrix of $\kappab$,
\begin{equation}
  \partial_{\xib}\kappab = [\partial_{\xib}a^\T, \partial_{\xib}b^\T, \partial_{\xib}c^\T, \partial_{\xib}d^\T, \partial_{\xib}e^\T, \partial_{\xib}f^\T]^\T,
\end{equation}
is given explicitly by
\begin{align}
  \partial_{\xib}a & = 
\begin{bmatrix}
  -\dfrac{2 \cos ^2\tau }{A^3}, & -\dfrac{2 \sin ^2\tau }{B^3}, & 0, & 0, & \left(\dfrac{1}{B^2}-\dfrac{1}{A^2}\right) \sin 2 \tau 
\end{bmatrix},
    \displaybreak[3] \nonumber \\                                                                          
\partial_{\xib}b & = 
\begin{bmatrix}
 -\dfrac{2 \sin 2 \tau }{A^3}, & \dfrac{2 \sin 2 \tau }{B^3}, & 0, & 0, & 2 \left(\dfrac{1}{A^2}-\dfrac{1}{B^2}\right) \cos 2 \tau 
\end{bmatrix},
     \displaybreak[3] \nonumber \\                                                                          
\partial_{\xib}c & = 
\begin{bmatrix}
-\dfrac{2 \sin ^2\tau }{A^3}, & -\dfrac{2 \cos ^2 \tau }{B^3}, & 0, & 0, & \left(\dfrac{1}{A^2}-\dfrac{1}{B^2}\right) \sin 2 \tau 
\end{bmatrix},                                                                          \displaybreak[3] \nonumber
  \\ 
\partial_{\xib}d & = 
\left[\begin{matrix}
    \dfrac{4 \cos ^2 \tau  (H+K \tan \tau )}{A^3}, & \dfrac{4 \sin \tau  (H \sin \tau -K \cos \tau )}{B^3},
  \end{matrix}\right.
                                                     \displaybreak[3] \nonumber
  \\
& \quad
\begin{matrix}
    & -\dfrac{2 \cos ^2 \tau }{A^2}-\dfrac{2 \sin ^2 \tau }{B^2}, & \left(\dfrac{1}{B^2}-\dfrac{1}{A^2}\right) \sin 2 \tau,
  \end{matrix}
                                                                    \displaybreak[3] \nonumber
  \\
& \quad
\left.\begin{matrix}
& \dfrac{2 (A-B) (A+B) (K \cos 2 \tau -H \sin 2 \tau )}{A^2 B^2} 
\end{matrix}\right],
                       \displaybreak[3] \nonumber
  \\ 
\partial_{\xib}e & = 
\left[\begin{matrix}
    \dfrac{4 \sin \tau  (H \cos \tau +K \sin \tau )}{A^3}, & \dfrac{4 \cos \tau  (K \cos \tau -H \sin \tau)}{B^3},
  \end{matrix}\right. \displaybreak[3] \nonumber
  \\
& \quad
\begin{matrix}
& \left(\dfrac{1}{B^2}-\dfrac{1}{A^2}\right) \sin 2 \tau, & -\dfrac{2 \cos ^2\tau }{B^2}-\dfrac{2 \sin ^2\tau}{A^2}, 
\end{matrix} \displaybreak[3] \nonumber
  \\
& \quad
\left.\begin{matrix}
 & \dfrac{2 (A-B) (A+B) (H \cos 2 \tau +K \sin 2 \tau)}{A^2 B^2} 
\end{matrix}\right],
   \displaybreak[3] \nonumber
  \\ 
\partial_{\xib}f & = \left[\begin{matrix}
-\dfrac{2 (H \cos \tau +K \sin \tau )^2}{A^3}, & -\dfrac{2 (K \cos \tau -H \sin \tau )^2}{B^3},  
\end{matrix}\right. \displaybreak[3] \nonumber
  \\
& \quad
\left.\begin{matrix}
 & \dfrac{2 (H+K \tan \tau ) \cos ^2 \tau }{A^2}+\dfrac{2 (H \sin \tau -K \cos \tau ) \sin \tau }{B^2},
\end{matrix}\right.  \displaybreak[3] \nonumber
  \\
& \quad
\left.\begin{matrix}
& \dfrac{2 \sin \tau  (H \cos \tau +K \sin \tau )}{A^2}+\dfrac{2 \cos \tau  (K \cos \tau -H \sin \tau )}{B^2},
\end{matrix}\right.  \displaybreak[3] \nonumber
  \\
& \quad
\left.\begin{matrix}
 &  \dfrac{2 \left(B^2-A^2\right) (K \cos \tau -H \sin \tau ) (H \cos \tau +K \sin \tau )}{A^2 B^2}
\end{matrix}\right].
\end{align}
The Jacobian of $\pib$ is given  by the more concise formula
\begin{equation}
  \partial_{\de}\pib 
  = \|\de\|^{-1}\left(\M{I}_{6} - \| \de \|^{-2}\de \de^\T \right),
\end{equation}
where $\M{I}_{6}$ is the $6 \times 6$ identity matrix.

In what follows, we use the notation $\ve{x} = [x, y]^\T$ and $\ve{u}(\ve{x}) = [x^2, xy, y^2, x, y, 1]^\T$, with which the ellipse equation (\ref{eq:20}) can be succinctly written as
\begin{math}
  \de^\T \ve{u}(\ve{x}) = 0.
\end{math}
The starting point for the main construction is the observation that when $\estAML$ is viewed as a multivariate normally distributed random vector,
\begin{equation}
  \estAML \sim N(\detrue, \cov{\detrue}),
\end{equation}
where $\detrue$ is the unit-normalised parameter vector of the true ellipse and $\cov{\detrue}$ is a covariance matrix, the scalar random variable $\estAML^T \ve{u}(\ve{x})$ is normally distributed with variance $\ve{u}(\ve{x})^\T \cov{\detrue} \ve{u}(\ve{x})$ for every point $\ve{x}$ on the locus
\begin{math}
  \locus{\detrue}
  = \{\ve{x} \in \R^2 \mid \detrue^T \ve{u}(\ve{x}) = 0\}
\end{math}
of the true ellipse.  The observation is based on the fact that
\begin{math}
  \estAML^T  \ve{u}(\ve{x})
  = (\estAML  -   \detrue)^T  \ve{u}(\ve{x})
\end{math}
whenever $\ve{x} \in \locus{\detrue}$ and the fact that by the rule of covariance propagation $(\estAML - \detrue)^T \ve{u}(\ve{x})$ has variance $\ve{u}(\ve{x})^\T \cov{\detrue} \ve{u}(\ve{x})$.  Consequently, under the assumption that $\estAML$ is an unbiased estimate of $\detrue$,
\begin{equation}
  z_{\ve{x}}  =
  \frac{(\estAML^T  \ve{u}(\ve{x}))^2} {\ve{u}(\ve{x})^\T  \cov{\detrue}
    \ve{u}(\ve{x})}
\end{equation}
is a squared normal random variable for every $\ve{x} \in \locus{\detrue}$.  Each $z_{\ve{x}}$, insofar as $\ve{x}$ belongs to $\locus{\detrue}$, attains large values with less probability than small values, with the probability of any particular set of values regarded as large or small being independent of $\ve{x}$.  This suggests using the $z_{\ve{x}}$ as building blocks in the construction of a confidence region in the plane.  Since the covariance $\cov{\detrue}$ is unknown, the $z_{\ve{x}}$ do not have observable realisations and, for the sake of construction, have to be replaced with these variables' \emph{observable} variants
\begin{equation}
  \obs{z}_{\ve{x}} = 
  \frac{(\estAML^T  \ve{u}(\ve{x}))^2} {\ve{u}(\ve{x})^\T  \cov{\estAML}
    \ve{u}(\ve{x})},
\end{equation}
where  the covariance  estimate  
\begin{math}
  \cov{\estAML} 
\end{math}
serves as a natural replacement for $\cov{\detrue}$.  Again, large observed values of $\obs{z}_{\ve{x}}$ are less plausible than small observed values as long as $\ve{x} \in \locus{\detrue}$.  It is thus natural to consider confidence regions for $\estAML$ in the form
\begin{math}
  \Set{ \ve{x} \in \R^2 | \obs{z}_{\ve{x}} \leq c}, 
\end{math}
where $c$ is a positive constant.  Ideally, for a confidence region at (confidence) level $1 - \alpha$, we should choose $c$ such that
\begin{equation}
  \Pr
  \left(z_{\ve{x}} \leq c
    \
    \text{for all $\ve{x} \in \locus{\detrue}$}
  \right)
  =
  \Pr
  \left( \sup_{\ve{x} \in \locus{\detrue}} z_{\ve{x}}
     \leq c
  \right)
  = 1 - \alpha.
\end{equation}
%where $\Pr(A)$ denotes the probability of the event $A$.
But the distribution of
\begin{math}
  \sup_{\ve{x} \in \locus{\detrue}} 
    z_{\ve{x}}
\end{math}
is not easy to determine, so as a second best choice we shall replace
\begin{math}
  \sup_{\ve{x} \in \locus{\detrue}} 
  z_{\ve{x}}
\end{math}
by a random upper bound whose distribution can be readily calculated.  Proceeding to the specifics, we first note that, since $\| \estAML \| = \| \detrue \| = 1$, we have
\begin{math}
  (\estAML - \detrue)^ \T \detrue 
  = -  \| \estAML - \detrue \|^2/2.
\end{math}
Consequently, resorting to the first order approximation around $\detrue$, we may next assume that
\begin{equation}
  \label{eq:22}
  (\estAML - \detrue)^\T \detrue = 0.
\end{equation}
Given a length-$n$ vector $\ve{a}$, let $\pro{\ve{a}}$ denote the $n \times n$ symmetric projection matrix given by
\begin{equation}
  \pro{\ve{a}}
  = \M{I}_{n} - \| \ve{a} \|^{-2} \ve{a} \ve{a}^{\T}.
\end{equation}
It is readily seen that, for each length-$n$ vector $\ve{x}$, $\pro{\ve{a}} \ve{x}$ represents the orthogonal projection of $\ve{x}$ onto the orthogonal complement of the space spanned by $\ve{a}$ in $\R^n$.  Now, \eqref{eq:22} can be restated as
\begin{equation}
  \label{eq:23}
  \estAML - \detrue = \pro{\detrue}(\estAML - \detrue).
\end{equation}
We also note that, in view of \eqref{eq:22},
\begin{equation}
  \cov{\detrue} \detrue
  = \Expec((\estAML - \detrue)(\estAML - \detrue)^\T) \detrue
  = \0,
\end{equation}
where $\Expec(\M{X})$ denotes the expectation of the random matrix $\M{X}$. Thus the null space of $\cov{\detrue}$, $\Nul(\cov{\detrue})$, contains $\detrue$. Typically, $\Nul(\cov{\detrue})$ will be one-dimensional and will be spanned by $\detrue$. Given a matrix $\M{A}$, let $\M{A}^+$ denote the Moore--Penrose pseudo-inverse of $\M{A}$, and, when $\M{A}$ non-negative definite, let $\M{A}^{1/2}$ denote the unique non-negative definite square root of $\M{A}$. By a general rule,
\begin{math}
  (\cov[+]{\detrue})^{1/2} \cov[1/2]{\detrue}
  =
  (\cov[1/2]{\detrue})^+\cov[1/2]{\detrue}
\end{math}
is a symmetric projection matrix representing the orthogonal projection onto the orthogonal complement of $\Nul(\cov{\detrue})$. Assuming that $\Nul(\cov{\detrue})$ is spanned by $\detrue$,
\begin{math}
  (\cov[+]{\detrue})^{1/2} \cov[1/2]{\detrue}
\end{math}
is a symmetric projection matrix representing the orthogonal projection onto the orthogonal complement of the space spanned by $\detrue$.  This is the same as saying that
\begin{equation}
  \label{eq:24}
  (\cov[+]{\detrue})^{1/2} \cov[1/2]{\detrue} = \pro{\detrue}.
\end{equation}
Now note that if $\ve{x} \in \locus{\detrue}$, then, by \eqref{eq:23},
\begin{equation}
  \estAML^\T \ve{u}(\ve{x}) = (\estAML - \detrue)^\T \ve{u}(\ve{x})
  = (\estAML - \detrue)^\T \pro{\detrue} \ve{u}(\ve{x})
\end{equation}
and further, by \eqref{eq:24}, 
\begin{equation}
  \begin{aligned}[b]
    \estAML^\T \ve{u}(\ve{x})
    & =  (\estAML - \detrue)^\T (\cov[+]{\detrue})^{1/2}
    \cov[1/2]{\detrue} \ve{u}(\ve{x})
    \\
    & =
    ((\cov[+]{\detrue})^{1/2} (\estAML - \detrue))^\T
    \cov[1/2]{\detrue} \ve{u}(\ve{x}).
  \end{aligned}
\end{equation}
By the Cauchy--Bunyakovsky--Schwarz inequality,
\begin{equation}
  (\estAML^\T \ve{u}(\ve{x}))^2 \leq
  \| (\cov[+]{\detrue})^{1/2} (\estAML - \detrue) \|^2
  \| \cov[1/2]{\detrue} \ve{u}(\ve{x}) \|^2.
\end{equation}
Also,
\begin{equation}
  \| (\cov[+]{\detrue})^{1/2} (\estAML - \detrue) \|^2
  = 
  (\estAML - \detrue)^\T \cov[+]{\detrue} (\estAML - \detrue)
\end{equation}
and
\begin{equation}
  \| \cov[1/2]{\detrue} \ve{u}(\ve{x}) \|^2 = \ve{u}(\ve{x})^\T
  \cov{\detrue} \ve{u}(\ve{x}).
\end{equation}
Hence,
\begin{equation} 
  z_{\ve{x}}
  \leq
  (\estAML - \detrue)^\T \cov[+]{\detrue} (\estAML - \detrue).
\end{equation}
Since $\ve{x}$ is an arbitrary member of $\locus{\detrue}$, we have
\begin{equation}
  \label{eq:25}
  \sup_{\ve{x} \in \locus{\detrue}}
  z_{\ve{x}}
  \leq
  (\estAML - \detrue)^\T \cov[+]{\detrue} (\estAML - \detrue).
\end{equation}
Now the random variable
\begin{math}
   (\estAML - \detrue)^\T \cov[+]{\detrue} (\estAML - \detrue)
\end{math}
has approximately a chi-squared distribution with $5$ degree of freedom.  Let $\chi^2_{5, \alpha}$ denote the $100(1 - \alpha)\%$ percentile of the $\chi^2$ distribution with $5$ degrees of freedom, characterised by the relation
\begin{math}
  \Pr\left( \chi^2 \leq \chi^2_{5,\alpha}\right) = 1 - \alpha.
\end{math}
Inequality~(\ref{eq:25}) guarantees that
\begin{equation}
  \Pr\left(
    \sup_{\ve{x} \in \locus{\detrue}}
    z_{\ve{x}}
    \leq \chi^2_{5,\alpha}
  \right) \geq 1 - \alpha.
\end{equation}
Substituting    $\cov{\estAML}$   for    $\cov{\detrue}$,   we    also
approximately have
\begin{equation}
  \Pr\left(
    \sup_{\ve{x} \in \locus{\detrue}}
    \obs{z}_{\ve{x}}
    \leq \chi^2_{5,\alpha}
  \right) \geq 1 - \alpha.
\end{equation}
This allows an approximate confidence region at level $1 - \alpha$ for
$\estAML$ to be taken as
\begin{math}
  %\Omega_{\alpha} =
  \Set{ \ve{x} \in \R^2 |
    \obs{z}_{\ve{x}}
    \leq \chi^2_{5,\alpha}}.
\end{math}
We  finally  point  out  that  if $\alpha$  is  set  to  the  standard
conventional value of $0.05$, then $\chi^2_{5, \alpha} = 11.07$.

\section{Experimental results}
\label{sec:experimental-results}

We compared the ML estimator for ellipse fitting introduced in \cref{sec:estimation-method} against the point-based direct ellipse fit (DEF points) \cite{fitzgibbon99:_direc} and its region-based gradient variant (DEF gradient) \cite{Ouellet2008}.  We obtained 2D points for DEF by applying Canny edge detection to the input image.  The gradient for the gradient-based DEF was computed using the Sobel operator.  We seeded the ML method with the point-based DEF. For some experimental conditions the point-based DEF was of such poor quality that the ML method converged to a sub-optimal solution. The sub-optimal solutions manifest as outliers in the boxplots of the ML results---had the ML method been seeded with a better initial value it would have converged to a superior solution.  In all of our plots, we denote ML with blue (\textcolor{ml-col}{\rule[1.5pt]{0.5cm}{1.5pt}}), point-based DEF with grey (\textcolor{dir-points-col}{\rule[1.5pt]{0.5cm}{1.5pt}}), and gradient-based DEF with yellow (\textcolor{dir-region-col}{\rule[1.5pt]{0.5cm}{1.5pt}}) colours.

\subsection{Synthetic images}
\label{sec:synthetic-images}

In this section, we present comprehensive results on simulated data.  The simulated data methodically imitates the image formation process and incorporates the effects of the point spread function, Poisson noise, discretisation error, and quantisation error.  Without loss of generality, the actual ellipses which gave rise to digital images always lay within the unit box $[0,1] \times [0,1]$.  We followed the steps outlined in \cref{sec:digitalImageFormation}
%of \cref{sec:estimation-method}
to form the digital images.

The experimental design facilitates a cogent interpretation of the standard deviation of the PSF as percentage area. For example, a value of $\sigma_{\mathrm{PSF}} = 0.1$ corresponds to ten percent of the digital image area.

\subsubsection{Varying signal-to-noise ratio}
\label{sec:varying-signal-noise}

\begin{figure*}[!th]
	\centering
	\subfloat[The boxplots show the distribution of the estimated geometric ellipse parameters for varying SNR.    The solid red line in the boxplots represents the true ellipse parameters. To aid interpretation, an example image for each SNR is presented in the bottom panel.]{\includegraphics[scale = 0.42]{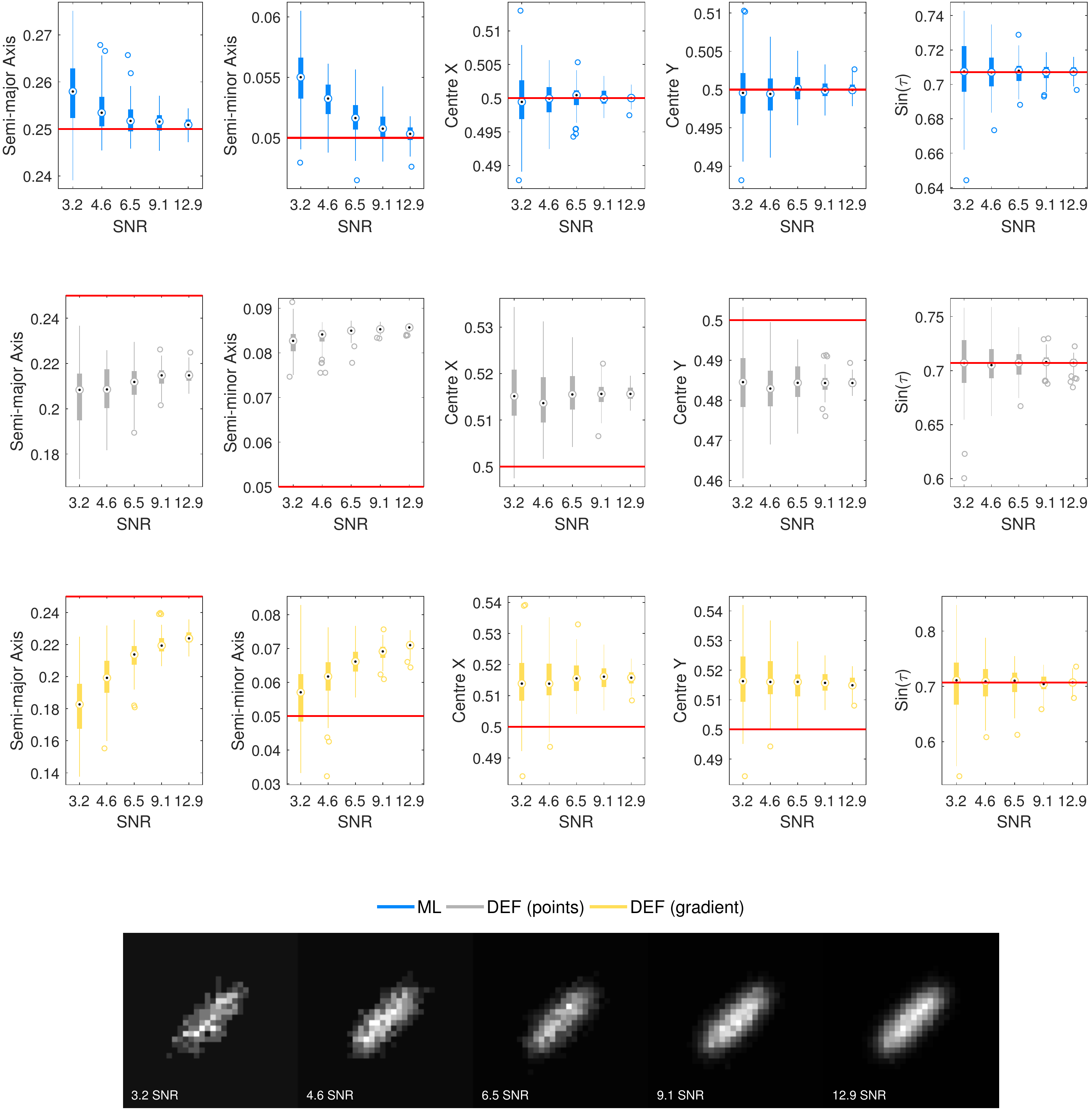}}
	\\
	\subfloat[Randomly chosen example result including a planar confidence region for each SNR level. The dotted black ellipse denotes the truth.]{
	\captionsetup[subfigure]{labelformat=empty}
	\subfloat[SNR=$3.2$]{\includegraphics[scale = 0.43]{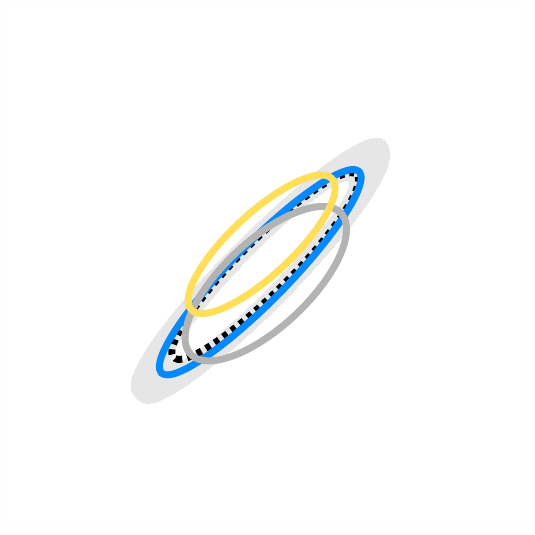}}
	\subfloat[SNR=$4.6$]{\includegraphics[scale = 0.43]{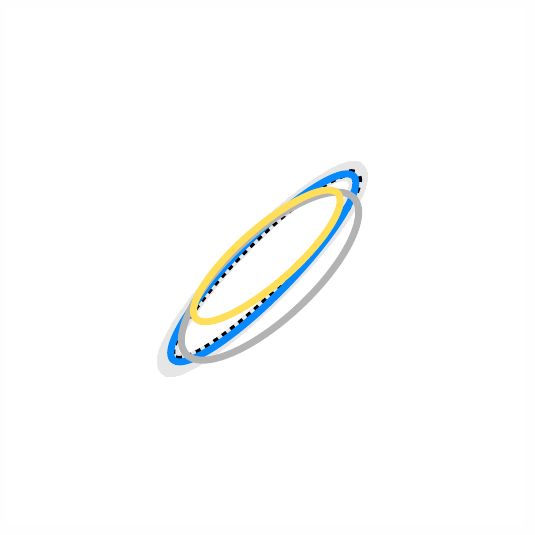}}
     \subfloat[SNR=$6.5$]{\includegraphics[scale = 0.43]{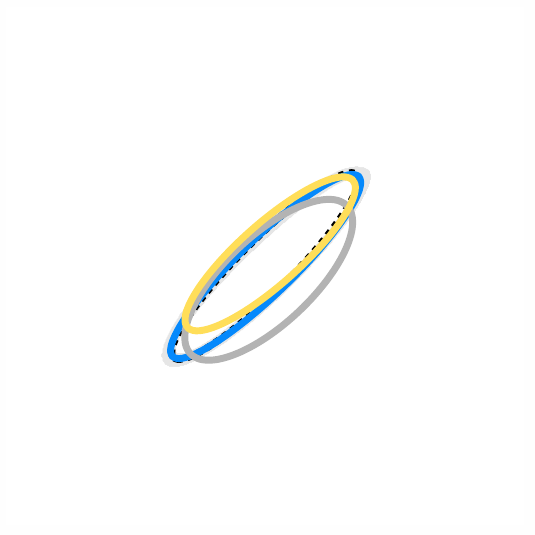}}
	\subfloat[SNR=$9.1$]{\includegraphics[scale = 0.43]{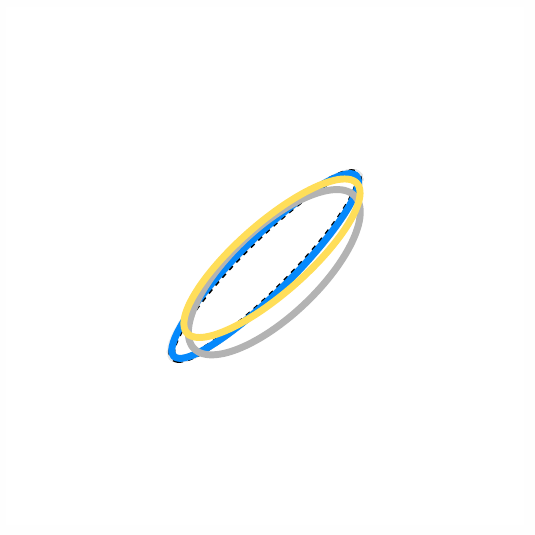}}
	\subfloat[SNR=$12.9$]{\includegraphics[scale = 0.43]{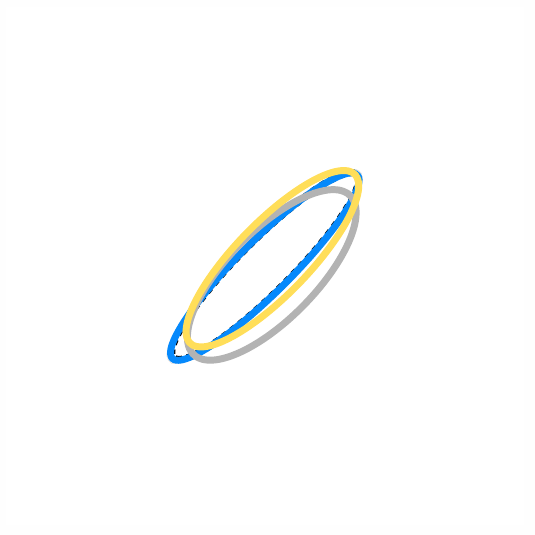}}
	\setcounter{subfigure}{2}% Reset subfigure counter
	}
  \caption{ Estimation results on ellipse with varying signal-to-noise ratios (SNRs) sampled on a $32 \times 32$ pixel grid with $\sigma_{\mathrm{PSF}} = 0.05$ and $b = 1$.  }
  \label{fig:5-Percent-Blur-32by32-1-b-boxplot}
\end{figure*}

Based on \eqref{eq:snr}, we characterised the signal-to-noise ratio as the square root of the number of photons in the brightest part of the \emph{real} digital image $f_{\mathrm{real}}$. This definition allowed us to distinguish between Poisson noise and the uncertainty introduced by quantisation.  In our first experiment the true parameter vector was given by $\xib =[0.25, 0.05, 0.5, 0.5, 0.785]^\T$. We sampled this ellipse with a square grid of 32 pixels and a Gaussian point spread function with a standard deviation of $5\%$. For the quantisation step, we set $b = 1$.  We explored how the photon count affects the accuracy of the estimator by varying the conversion factor $C$ in powers of two (16, 32, 64, 128, and 256).  Each conversion factor induced a different signal-to-noise ratio. For each choice of $C$ we conducted a hundred random trials.  The results of these experiments are displayed in \cref{fig:5-Percent-Blur-32by32-1-b-boxplot}.

\begin{figure*}[!th]
	\centering
	\subfloat[The boxplots show the distribution of the estimated geometric ellipse parameters for varying SNRs.    The solid red line in the boxplots represents the true ellipse parameters. To aid interpretation, an example image for each  		                     SNR is presented      in the bottom panel.]{\includegraphics[scale = 0.42]{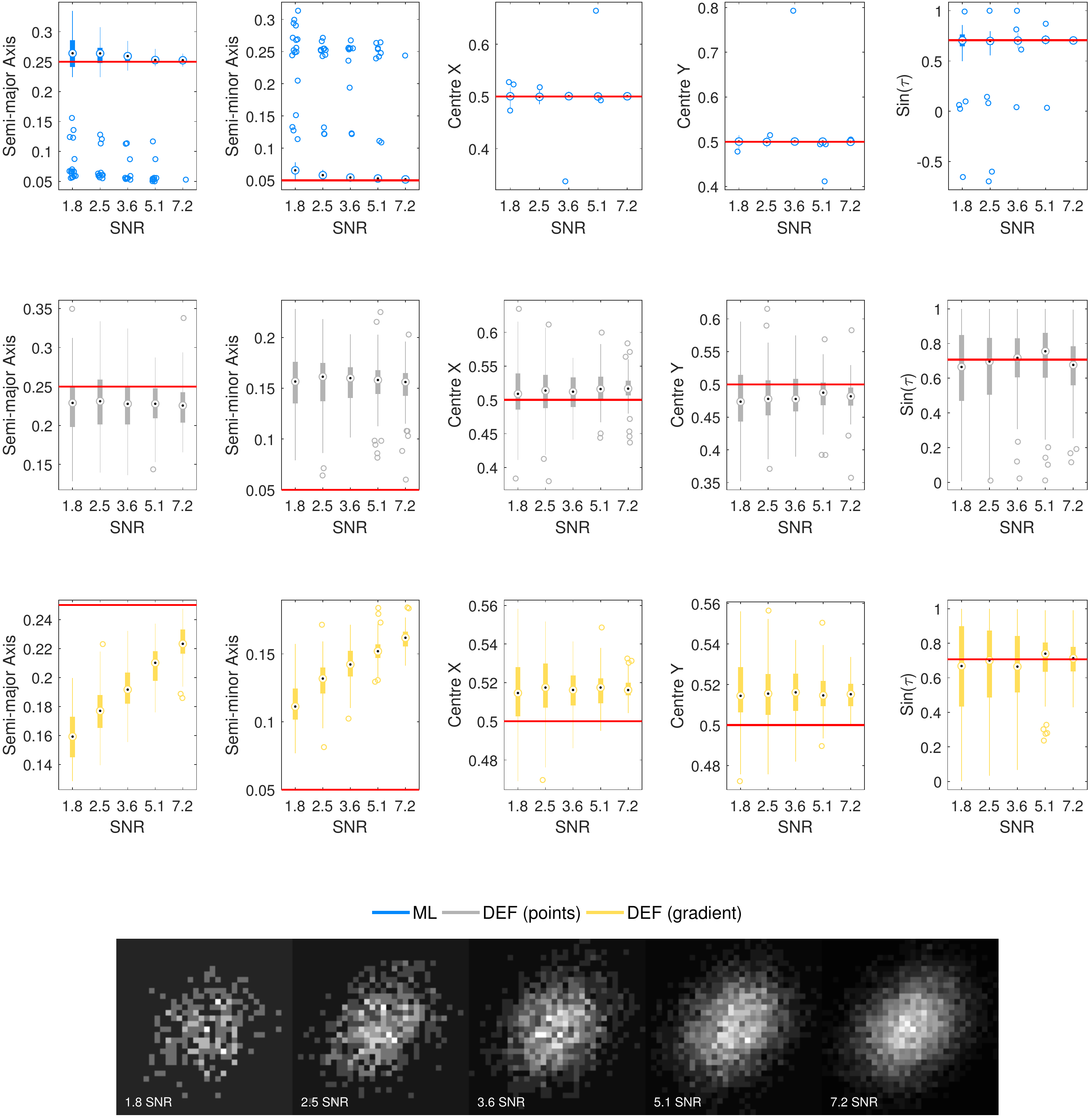}}
	\\
	\subfloat[Randomly chosen example result including a planar confidence region for each SNR level. The dotted black ellipse denotes the truth.]{
	\captionsetup[subfigure]{labelformat=empty}
	\subfloat[SNR=$3.2$]{\includegraphics[scale = 0.43]{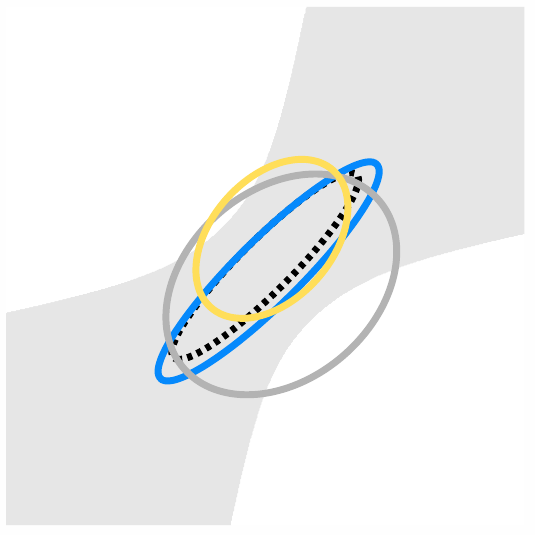}}
	\subfloat[SNR=$4.6$]{\includegraphics[scale = 0.43]{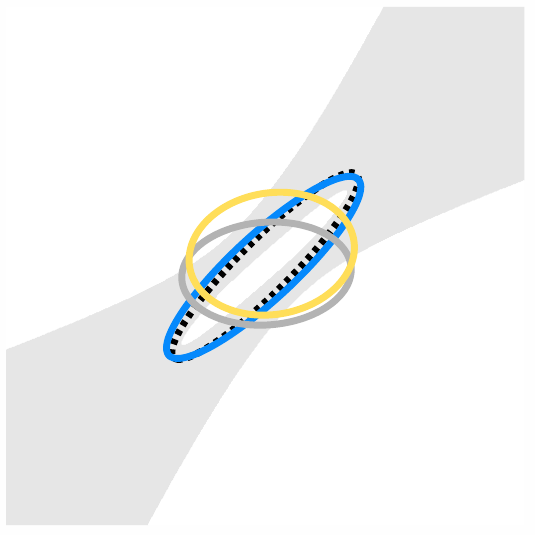}}
     \subfloat[SNR=$6.5$]{\includegraphics[scale = 0.43]{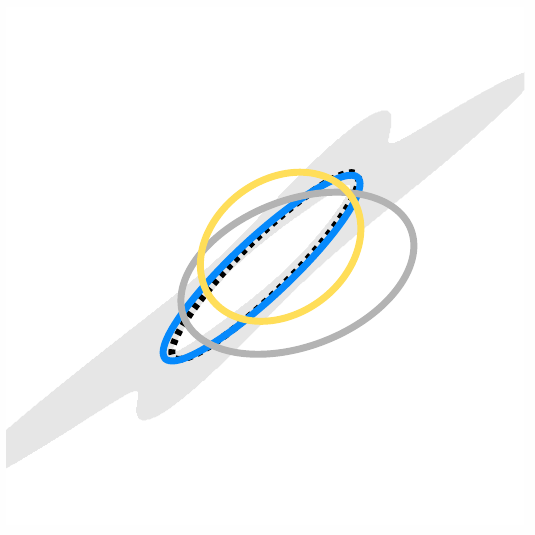}}
	\subfloat[SNR=$9.1$]{\includegraphics[scale = 0.43]{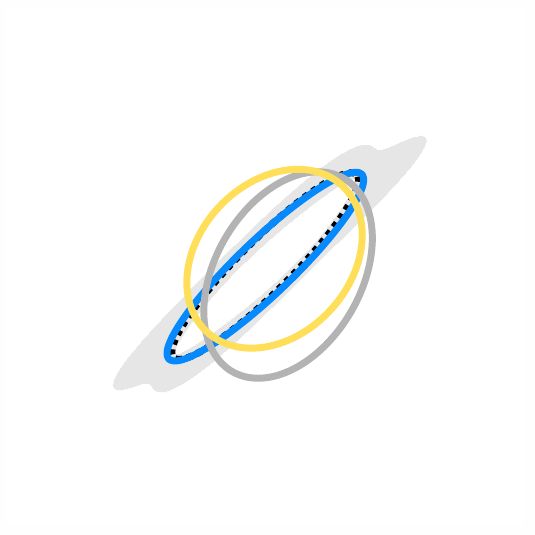}}
	\subfloat[SNR=$12.9$]{\includegraphics[scale = 0.43]{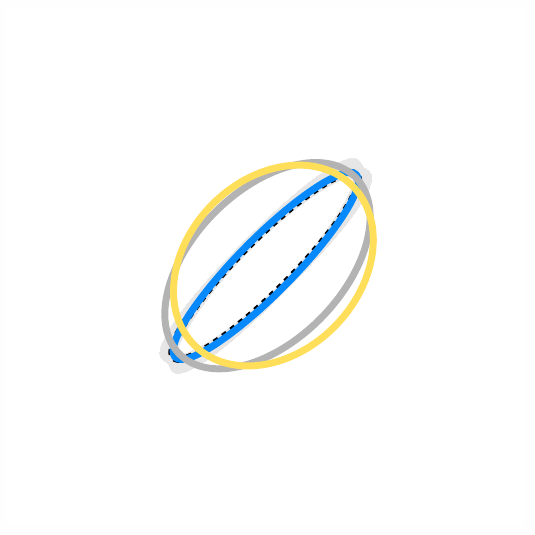}}
	\setcounter{subfigure}{2}% Reset subfigure counter
	}
  \caption{ Estimation results on ellipse with varying signal-to-noise ratios (SNR) sampled on a $32 \times 32$ pixel grid with $\sigma_{\mathrm{PSF}} = 0.15$ and $b = 1$. }
  \label{fig:15-Percent-Blur-32by32-1-b-boxplot}
\end{figure*}

Our second experiment was identical to the first, except that we used a Gaussian PSF with a  standard deviation of $15\%$. The results of the  second experiment are displayed in \cref{fig:15-Percent-Blur-32by32-1-b-boxplot}.

\paragraph*{Discussion}
\label{sec:discussion-1}

Both experiments demonstrate that prevailing ellipse estimation methods are remarkably inadequate when working at the limits of resolution.  The boxplots for the DEF estimates evidence a substantial bias and demonstrate that these methods cannot recover the true ellipse parameters.  The example ellipse fits communicate the deficiencies of DEF in a visually noticeable manner. In contrast the proposed ML method yields accurate estimates even with very low signal-to-noise ratios.  The planar confidence regions rightly communicate the fact that there is more uncertainty in the estimate of the major axis, and that the uncertainty is greater for photon-limited images.

\subsubsection{Varying quantisation}
\label{sec:varying-quantisation}

\begin{figure*}[!th]
	\centering
	\subfloat[The boxplots show the distribution of the estimated
  			  geometric ellipse parameters for varying quantisation levels.    The solid red line in the boxplots represents the true ellipse parameters. To aid interpretation an example image for each quantisation level $b$ is presented in the bottom panel together with a description of the final number of grey levels.]{\includegraphics[scale = 0.42]{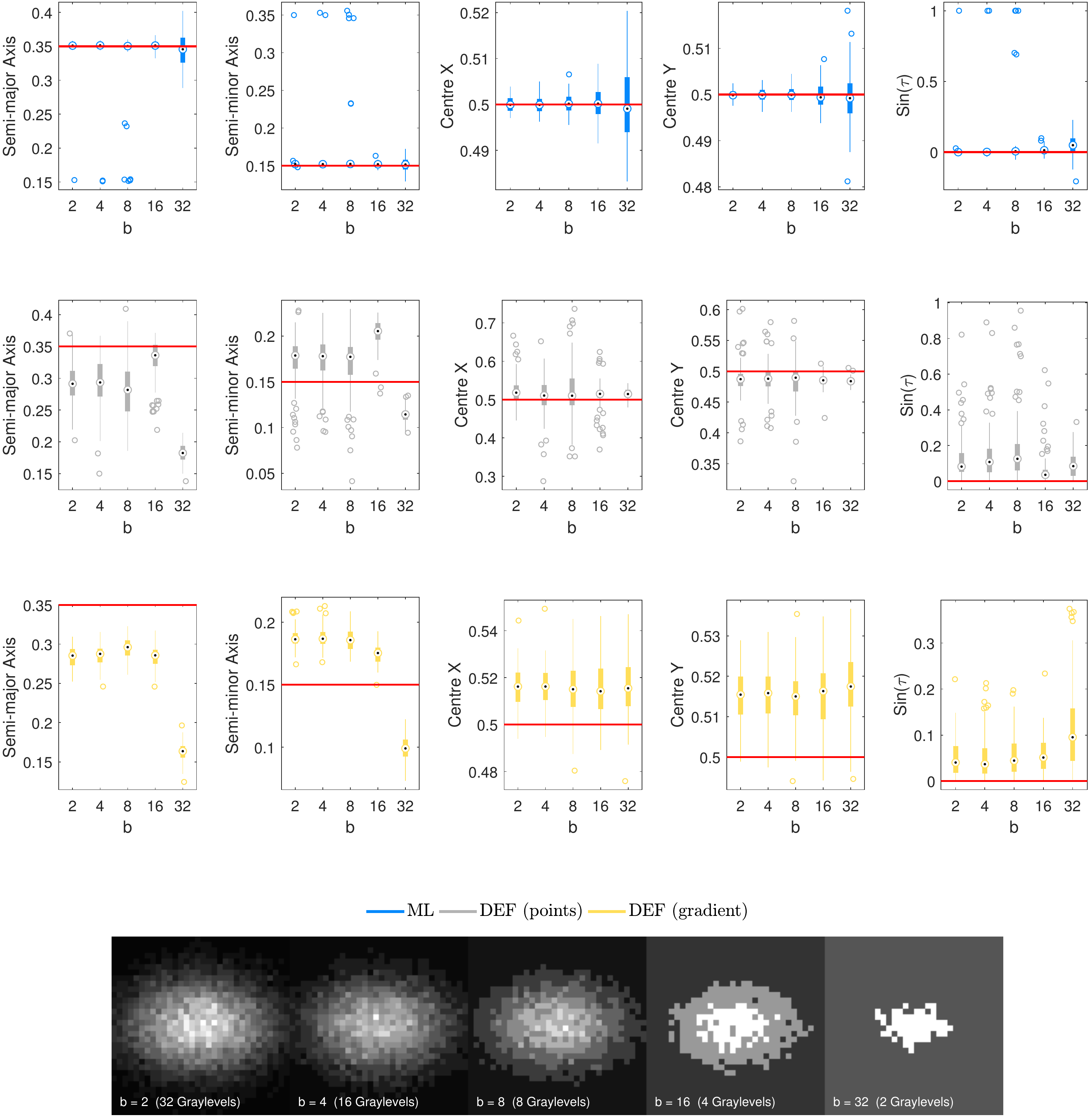}}
	\\
	\subfloat[Randomly chosen  example result including a planar confidence region for each quantisation level. The dotted black ellipse denotes the truth.]{
	\captionsetup[subfigure]{labelformat=empty}
	\subfloat[b=$2$]{\includegraphics[scale = 0.43]{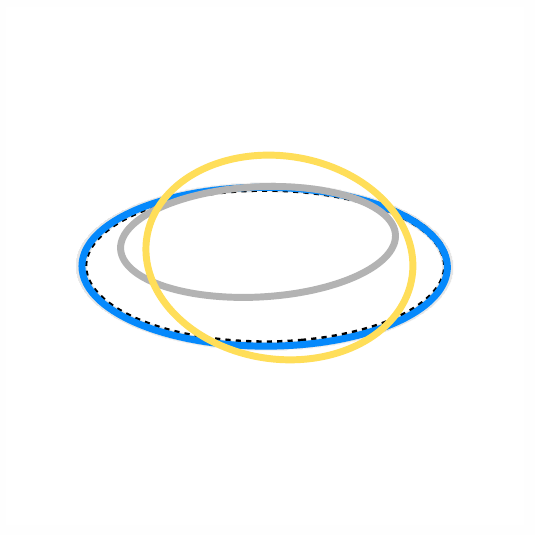}}
	\subfloat[b=$4$]{\includegraphics[scale = 0.43]{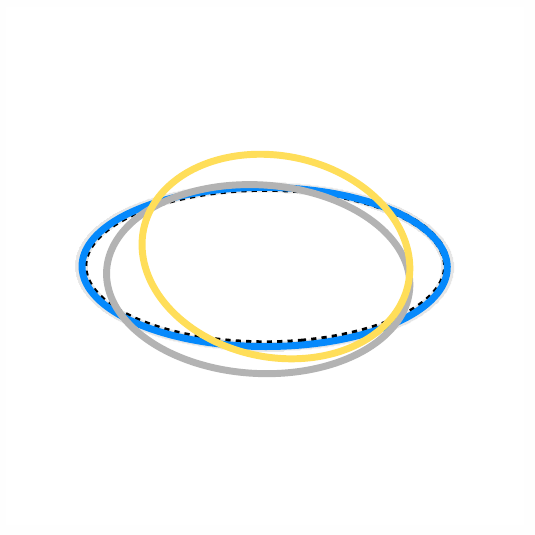}}
     \subfloat[b=$8$]{\includegraphics[scale = 0.43]{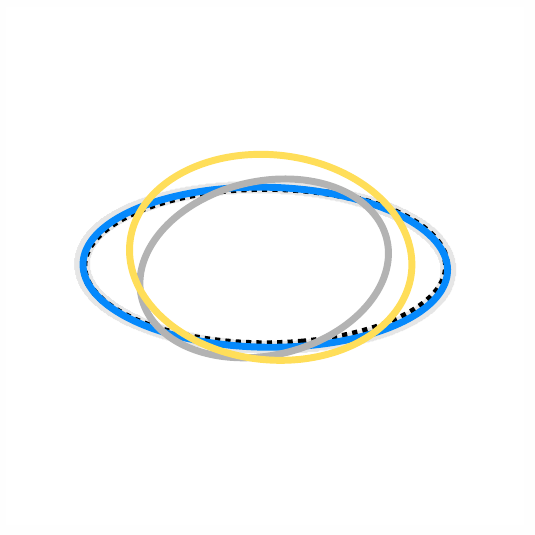}}
	\subfloat[b=$16$]{\includegraphics[scale = 0.43]{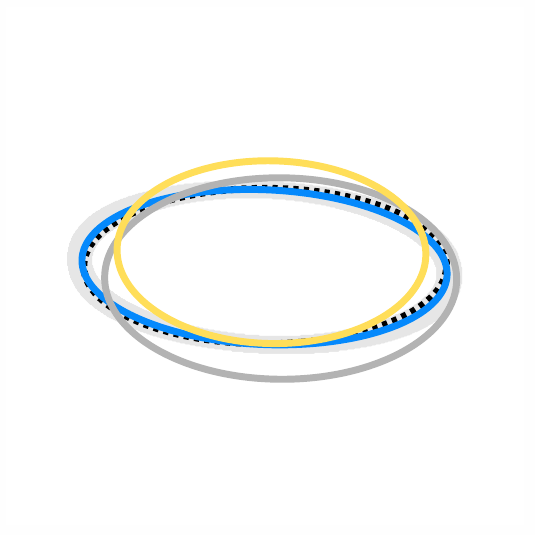}}
	\subfloat[b=$32$]{\includegraphics[scale = 0.43]{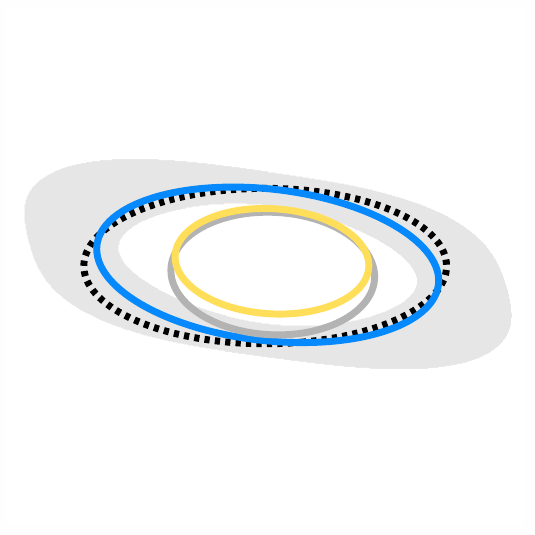}}
	\setcounter{subfigure}{2}% Reset subfigure counter
	}
   \caption{Estimation results on ellipses with varying quantisation levels sampled on a $32 \times 32$ pixel grid with $\sigma_{\mathrm{PSF}} = 0.15$ and SNR equal to $8.8683$.  }
   \label{fig:15-Percent-Blur-32by32-quantisation-boxplot}
\end{figure*}

For our third experiment we explored how different quantisation levels impact the precision of the estimates.  The true parameter vector was given by $\xib =[0.35, 0.15, 0.5,0.5,0]^\T$. We sampled this ellipse with a square grid of 32 pixels and a Gaussian point spread function with a standard deviation of $15\%$.  We modified the quantisation half-width $b$ in powers of two (2, 4, 8, 16, and 32) and conducted a hundred random trials for each value.  \Cref{fig:15-Percent-Blur-32by32-quantisation-boxplot} summarises the outcome of the third experiment.

\paragraph*{Discussion}
\label{sec:discussion-2}

The experiment validates our quantisation model. Even with extreme quantisation---a binary image---the ML method still yields estimates that are almost the correct parameters. The planar confidence regions also confirm that greater quantisation levels inflate the uncertainty of the estimates.  In comparison, both DEF estimates produce inadequate results for all quantisation levels.
 
\subsubsection{Varying eccentricity}
\label{sec:varying-eccentricity}

\begin{figure*}[!th]
	\centering
	\subfloat[The boxplots show the distribution of the estimated geometric ellipse parameters for varying eccentricities.    The solid red line in the boxplots represents the true ellipse parameters. To aid interpretation, an example image for each eccentricity is presented in the bottom panel.]{\includegraphics[scale = 0.42]{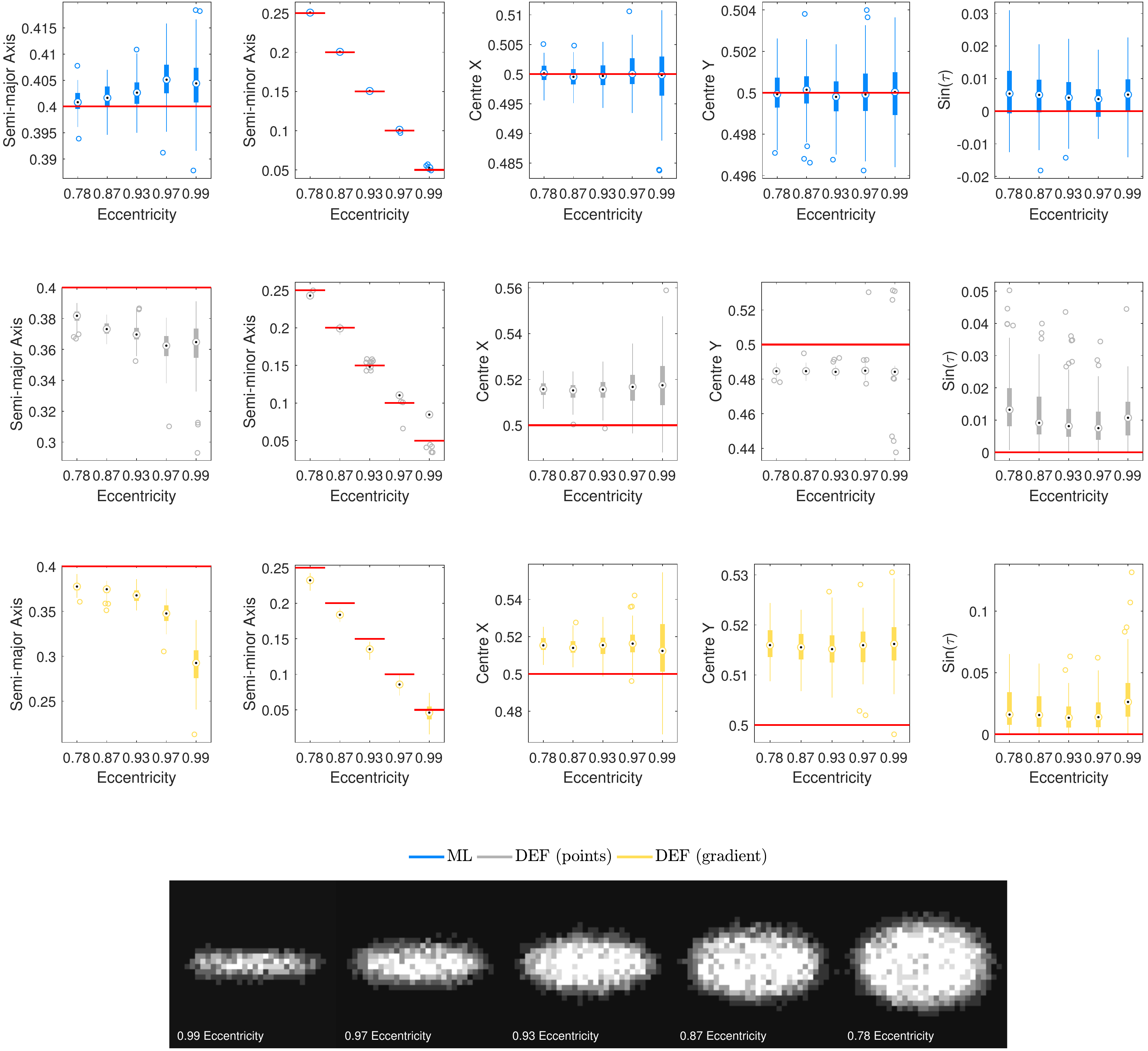}}
	\\
	\subfloat[Randomly chosen example result including a planar confidence region for each eccentricity. The dotted black ellipse denotes the truth. The planar confidence regions encompass the truth tightly reflecting that there is not much uncertainty. 					  They are not visible without zooming in. ]{
	\captionsetup[subfigure]{labelformat=empty}
	\subfloat[$0.99$]{\includegraphics[scale = 0.43]{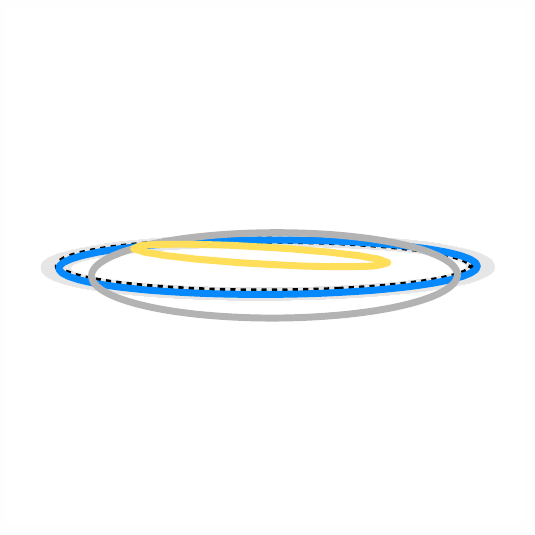}}
	\subfloat[$0.97$]{\includegraphics[scale = 0.43]{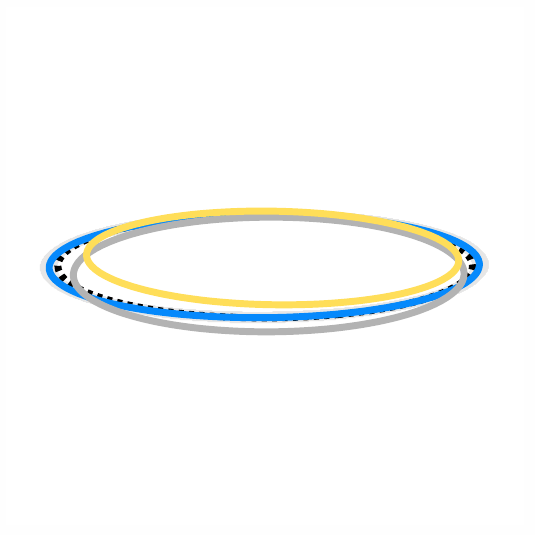}}
     \subfloat[$0.93$]{\includegraphics[scale = 0.43]{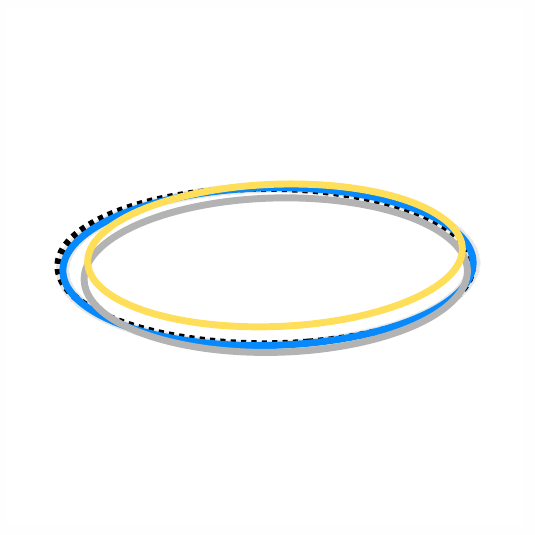}}
	\subfloat[$0.87$]{\includegraphics[scale = 0.43]{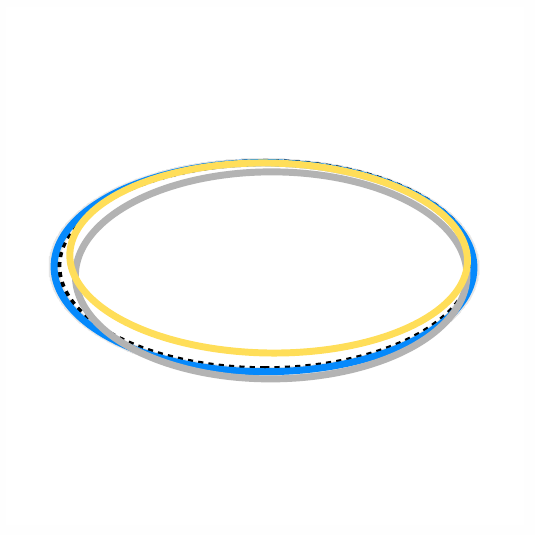}}
	\subfloat[$0.78$]{\includegraphics[scale = 0.43]{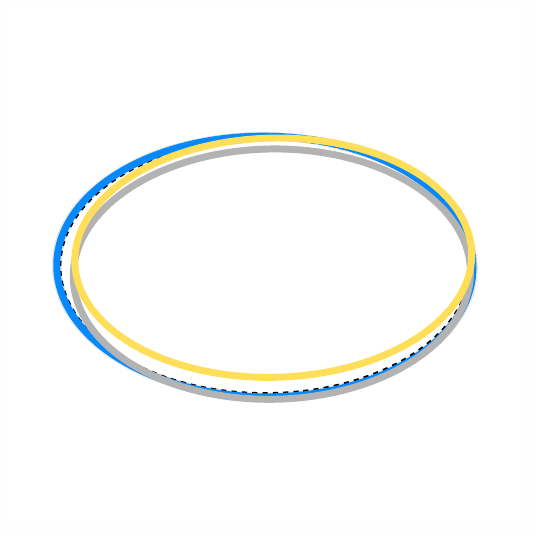}}
	\setcounter{subfigure}{2}% Reset subfigure counter
	}
   \caption{Estimation results on ellipses of varying eccentricity sampled on a $32 \times 32$ pixel grid with $\sigma_{\mathrm{PSF}} = 0.05$ and SNR ranging from $4.5200$ to $5.6568$.  }
   \label{fig:5-Percent-Blur-32by32-2-b-eccentricity-boxplot}
\end{figure*}

In general,  parameter estimation of an ellipse is more challenging when the eccentricity is substantial. In the fourth series of experiments, we investigated how eccentricity affects the quality of the estimates. In particular, we generated ellipses with eccentricities ranging from $0.78$ to $0.99$ and sampled these ellipse with a square grid of 32 pixels and a Gaussian PSF with a  standard deviation of $5\%$. For the quantisation step, we set $b = 1$.  
\Cref{fig:5-Percent-Blur-32by32-2-b-eccentricity-boxplot} summarises our findings. 

\paragraph*{Discussion}
\label{sec:discussion-3}

The results indicate that both DEF methods significantly and systematically underestimate the length of the semi-major axis. The bias is even more prominent for high-eccentricity ellipses. Contrastingly, the ML method produces accurate results for ellipses with high or low eccentricity.  The planar confidence regions indicate that the semi-major axis is less certain for high-eccentricity ellipses.

\subsubsection{Varying sampling grid}
\label{sec:vary-sampl-grid}

\begin{figure*}[!th]
  \centering
  \subfloat[The boxplots show the distribution of the estimated geometric ellipse parameters for varying sampling grids.    The solid red line in the boxplots represents the true ellipse parameters. A circle against a dashed black line in a boxplots indicates that a particular value exceeded the plot limit of the $y$ axis.  To aid interpretation, an example image for each grid size is presented in the bottom panel.]{\includegraphics[scale = 0.42]{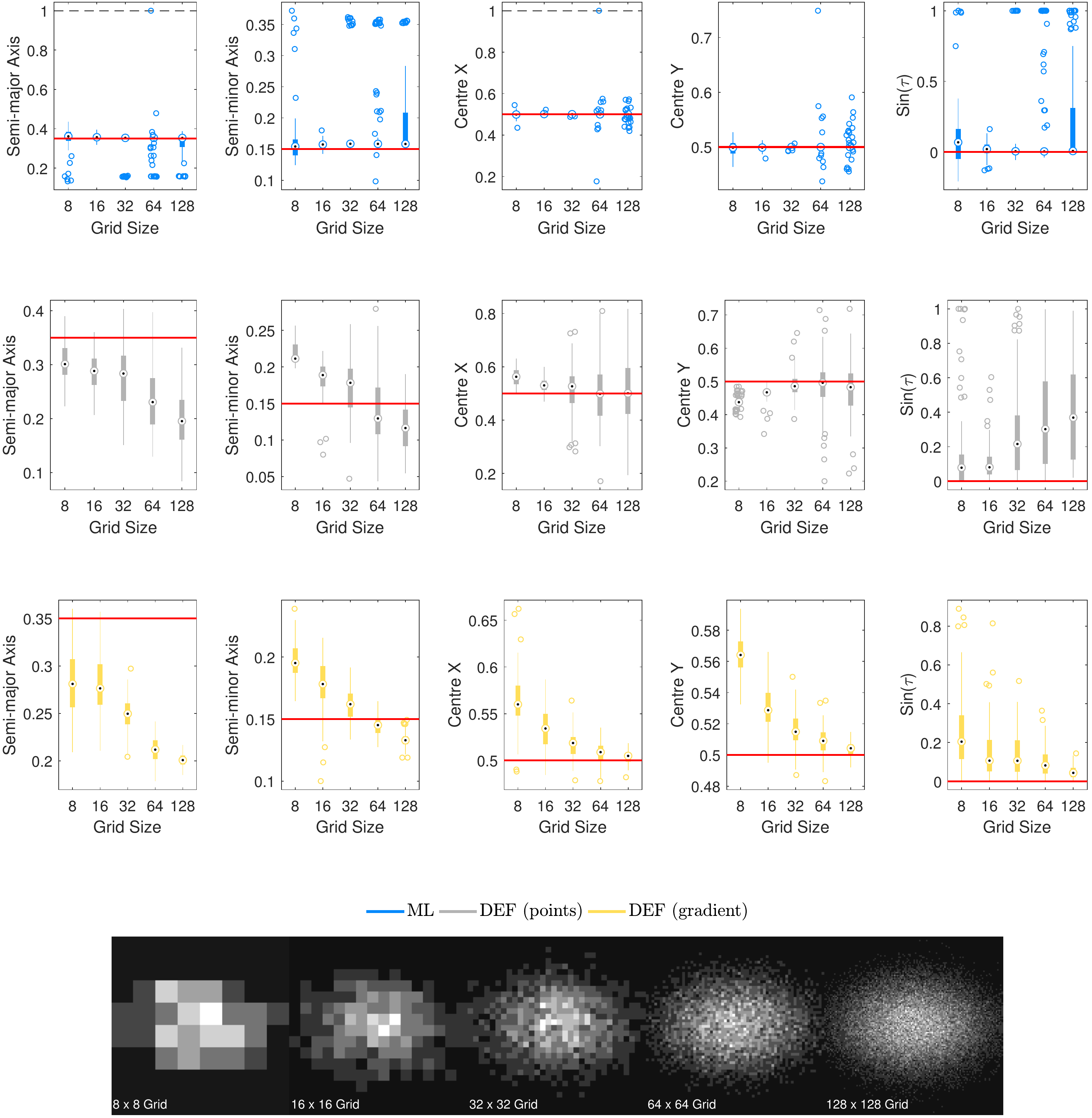}}
  \\
  \subfloat[Randomly chosen example result including a planar confidence region for each sampling grid size. The dotted black ellipse denotes the truth. ]{
    \captionsetup[subfigure]{labelformat=empty}
    \subfloat[$8 \times 8$]{\includegraphics[scale = 0.43]{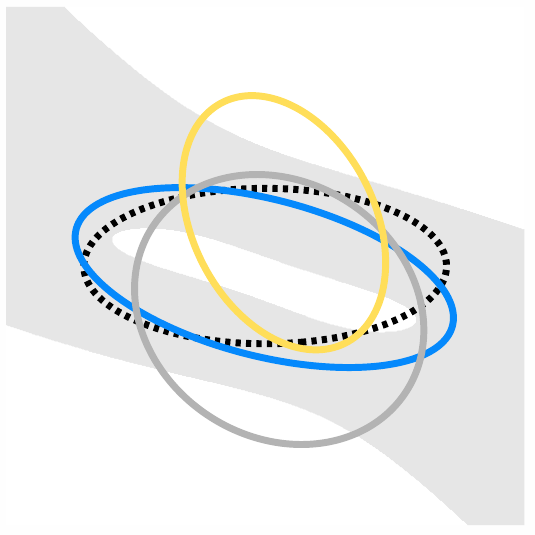}}
    \subfloat[$16 \times 16$]{\includegraphics[scale = 0.43]{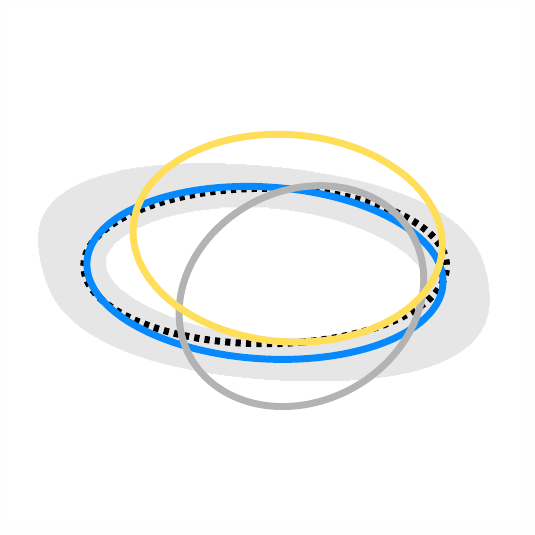}}
    \subfloat[$32 \times 32$]{\includegraphics[scale = 0.43]{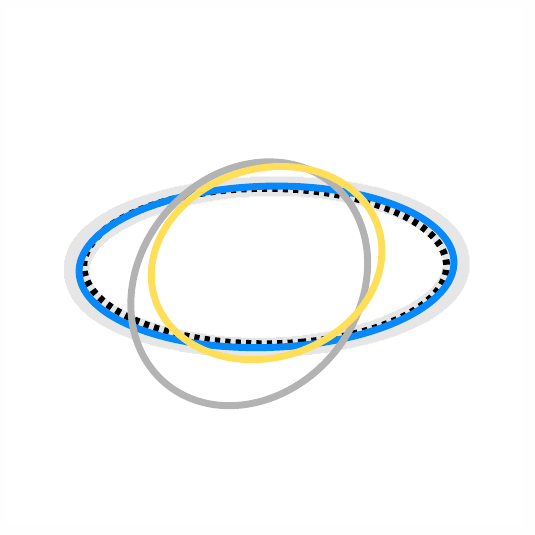}}
    \subfloat[$64 \times 64$]{\includegraphics[scale = 0.43]{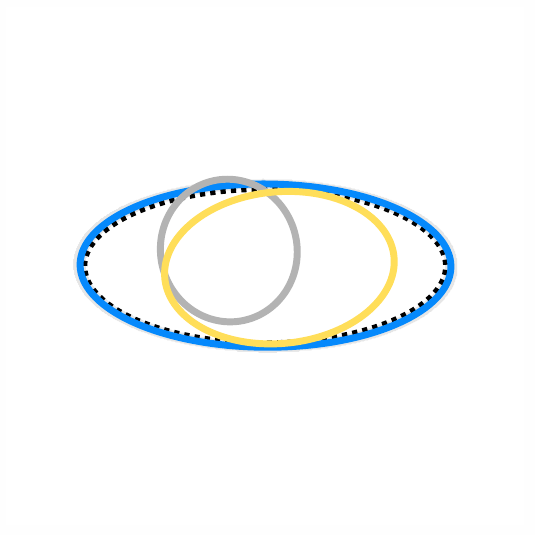}}
    \subfloat[$128 \times 128$]{\includegraphics[scale = 0.43]{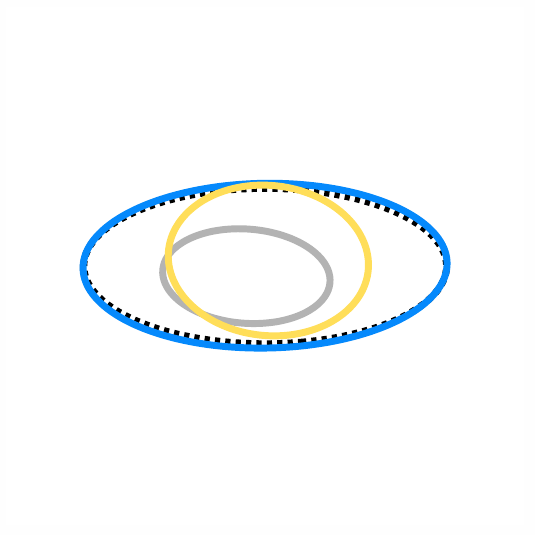}}
    \setcounter{subfigure}{2}% Reset subfigure counter
  }
  \caption{Estimation results on ellipses sampled on varying pixel grids with $\sigma_{\mathrm{PSF}} = 0.15$ and  SNR ranging from $4.1977 $ to $4.4527$. }
  \label{fig:32-Gain-15-Percent-Blur-2-b-sampling-boxplot}
\end{figure*}

In our final set of experiments, we investigated how image resolution influences the precision of the estimates.  Specifically, we varied the sampling grid in powers of two ($8 \times 8$, $16 \times 16$, $32 \times 32$, $64 \times 64$, and $128 \times 128$). The true parameter vector was given by $\xib =[0.35, 0.15, 0.5, 0.5, 0]^\T$. We used a Gaussian point spread function with a standard deviation of $15\%$ which produced a SNR ranging from $4.1977 $ to $4.4527$. For quantisation, we set $b = 1$.  \Cref{fig:32-Gain-15-Percent-Blur-2-b-sampling-boxplot} illustrates the results.

\paragraph*{Discussion}
\label{sec:discussion-4}

The final set of simulations affirm the necessity and efficacy of our maximum likelihood model. Even at the limits of resolution, using an $8 \times  8$ pixel grid, the ML method produces plausible parameter estimates. In accordance with expectation, the planar confidence regions demonstrate greater parameter uncertainty for low-resolution pixel grids than for higher resolution grids.  The performance of the DEF methods is poor. Evidently, the DEF methods are not applicable for these types of low-resolution images. 

\subsection{Real images}
\label{sec:real-images}

To corroborate the conclusions of the synthetic data experiments we conducted further laboratory experiments with real images. We used the UI-1220LE-M-GL camera (IDS Imaging Development Systems GmbH) and attached the MP0814-MP2 (IDS Imaging Development Systems GmbH) lens to the camera. The camera has a global shutter and an 8-bit monochrome CMOS sensor with a resolution of $752 \times 480$ pixels. 
We constructed a real ellipse region by glueing a white, elliptic sticker onto a piece of black cardboard.  

\subsubsection{Experimental design}
\label{sec:experimental-design}

The software for the camera permits the configuration of various low-level settings, such as exposure time, gain, black-level offset, and quantisation, to name but a few.  We set the gamma correction factor to unity to guarantee a linear luminance output.

To ensure that we obtained a reasonable approximation of the exact ellipse parameters from the real images, we carefully adjusted the lens and configured the camera to produce sharp and clear images.  We then recorded a sequence of 240 images and took the average geometric ellipse parameters estimated by our ML method as the best guess for the exact parameters. We substantiated our methodology by noting that the variance of the 240 estimated parameters was negligible and that the DEF methods produced similar estimates on the sharp and clear images.

The black cardboard was not absolutely black, and so, unlike our synthetic experiments, we did not estimate elliptic regions using a model of a black background.  Instead, we used the model for a grey background described in \cref{sec:grey-background}%
%of \cref{sec:pixel-resp-funct}
. Upon inspecting the histograms of a series of images, which revealed grey values in the range from 10 to 30 for the black cardboard, we set the background intensity value to $c = 0.15$.

Another important aspect when working with real images is finding an appropriate value for the conversion factor $C$ which links the image intensity with the photo-electron count (see \cref{sec:probmodel}). The value of $C$ determines the level of Poisson noise. In practice, we constrain $C$ to be a multiple of the maximum intensity in the image. The intuition underpinning this constraint is that we first need to adjust our model image which lies in the unit interval so that its brightest value (a value of $1$) matches the brightest observed intensity in the actual image (a value between $0$ and $255$). Subsequently, we need to convert the intensities into plausible photon counts. If the image is dark and noisy, then we multiply by a small positive integer to model a photon-limited scenario. If it looks relatively noise free, then we can multiply by a more significant positive integer. Since our model can generate a synthetic image, finding a suitable value for $C$ is not too complicated. A wrong choice of $C$ will result in a synthetic image that either looks too noisy or not noisy enough. The correct choice of $C$ will produce an image that resembles the observed image. Apart from choosing an appropriate value of $C$ by qualitatively comparing the synthetic images against the actual images, one could also quantify the root-mean-square error between the synthesised and actual image. Furthermore, one could develop a particular calibration step to identify the correct conversion factor. We opted to set $C$ based on empirical observations and settled on a value of $C = 25 \times G$, where $G$ is the maximum grey value in a given image.

By altering the configuration properties of the camera and adjusting the lens we were able to replicate many of the synthetic image experiments. For each experimental condition, we recorded a series of 240 images and used these to test the performance of the algorithms. 

\begin{figure}[t]
  \centering
  \captionsetup[subfigure]{labelformat=empty}
  \subfloat[Actual image]{\includegraphics[scale = 0.2]{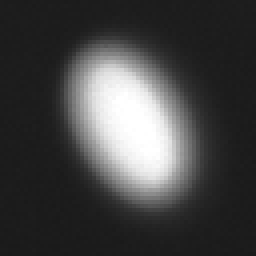}} 
  \hfil
  \subfloat[Synthesised image]{\includegraphics[scale = 0.2]{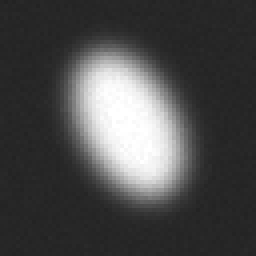}}
  
  \subfloat[]{\includegraphics[scale = 0.4]{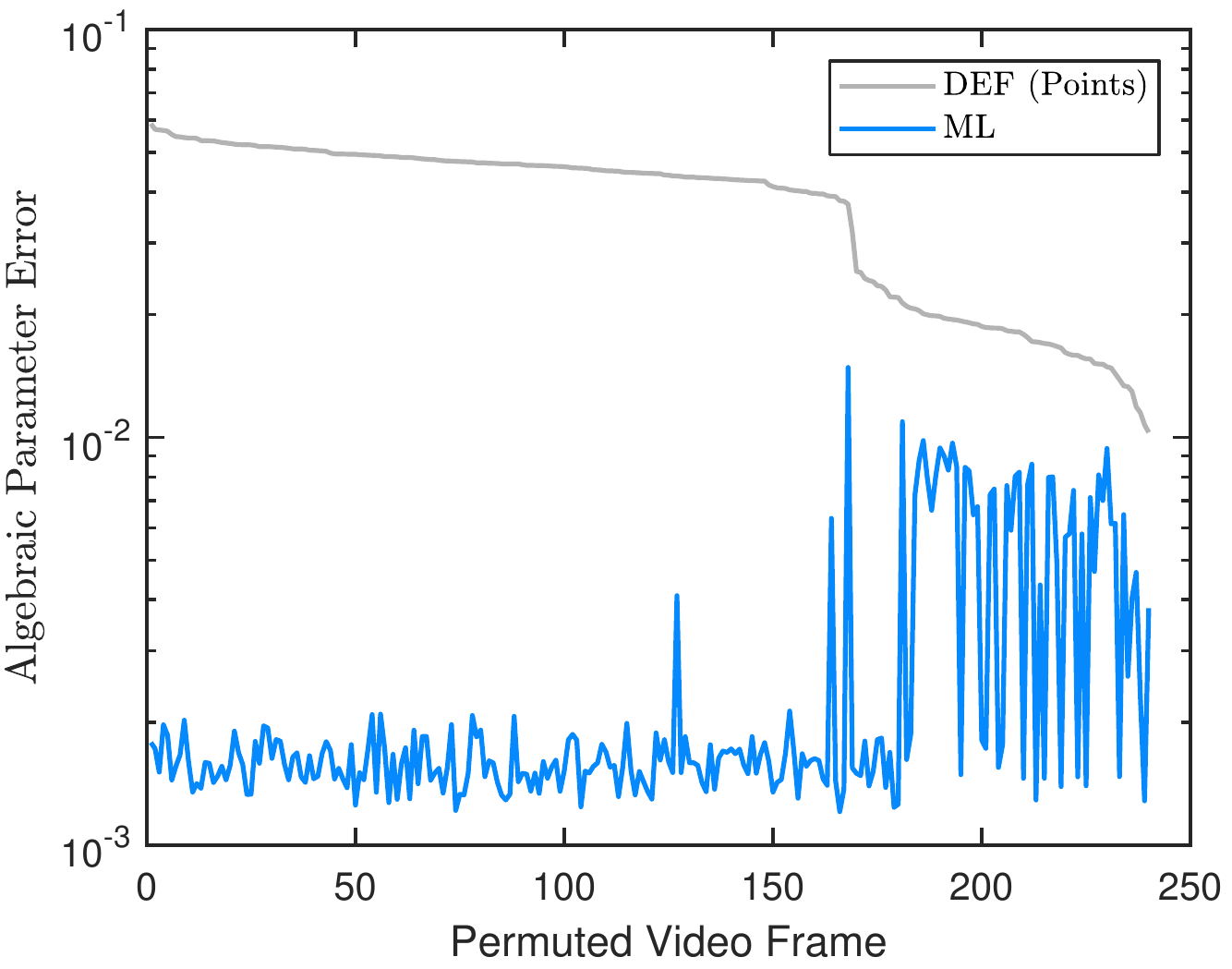}} 
  \hfil
  \subfloat[]{\includegraphics[scale = 0.4,trim=0 -2.7cm 0 0]{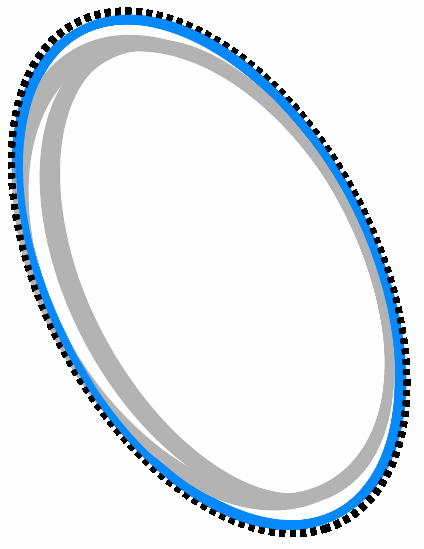}}
  % 
  % \begin{tabular}{cc}
  %   \subfloat[Actual image]{\includegraphics[scale = 0.4]{blur-light-camera}} 
  %   &
  %     \subfloat[Synthesised image]{\includegraphics[scale = 0.4]{blur-light-model}} \\
  %   \subfloat[]{\includegraphics[scale = 0.4]{blur-light-parameter-error}} 
  %   &
  %     \subfloat[]{\includegraphics[scale = 0.4]{blur-light-ellipses}}
  % \end{tabular}
  %
  \caption{Experiment 1. Estimation results on a sequence of real images ($64 \times 64$ pixels) where the lens was defocused to induce a blurred image. We permuted the sequence of images in the bottom left panel so that the error of the point-based direct ellipse fit is sorted in descending order. The bottom right panel shows the estimates for the entire sequence overlayed on a single figure. The dotted black line demarcates the true ellipse. }
  \label{fig:experiment1}
\end{figure}

\subsubsection{Experiments}
\label{sec:experiments}

We quantified the performance of the estimators by considering the algebraic ellipse parameters.  The fidelity of the algebraic parameters was evaluated by using an \emph{algebraic parameter error}, defined as
\begin{math}
  \left\| \pro{\truede} \widehat{\de} \right\|,
\end{math}  
where $\truede$ denotes the true value, and both $\truede$ and $\widehat{\de}$ are assumed to have unit norm. 

\paragraph*{Experiment 1}
\label{sec:experiment-1}

In our first set of experiments, we adjusted the camera lens so that the target image was out of focus and blurred.  We cropped a $64 \times 64$ region of interest that contained the ellipse region and used it as input to our estimators.  We initialised the Gaussian PSF with a standard deviation of $.5\%$ and set $b$ equal to $0$. 
The results are displayed  in \cref{fig:experiment1}.

\paragraph*{Experiment 2}
\label{sec:experiment-2}

\begin{figure}[!t]
  \centering
  \captionsetup[subfigure]{labelformat=empty}
  \subfloat[Actual image]{\includegraphics[scale = 0.2]{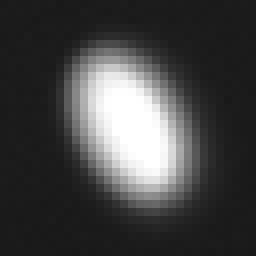}} 
  \hfil
  \subfloat[Synthesised image]{\includegraphics[scale = 0.2]{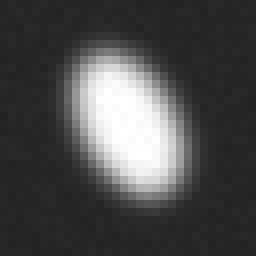}}
  
  \subfloat[]{\includegraphics[scale = 0.4]{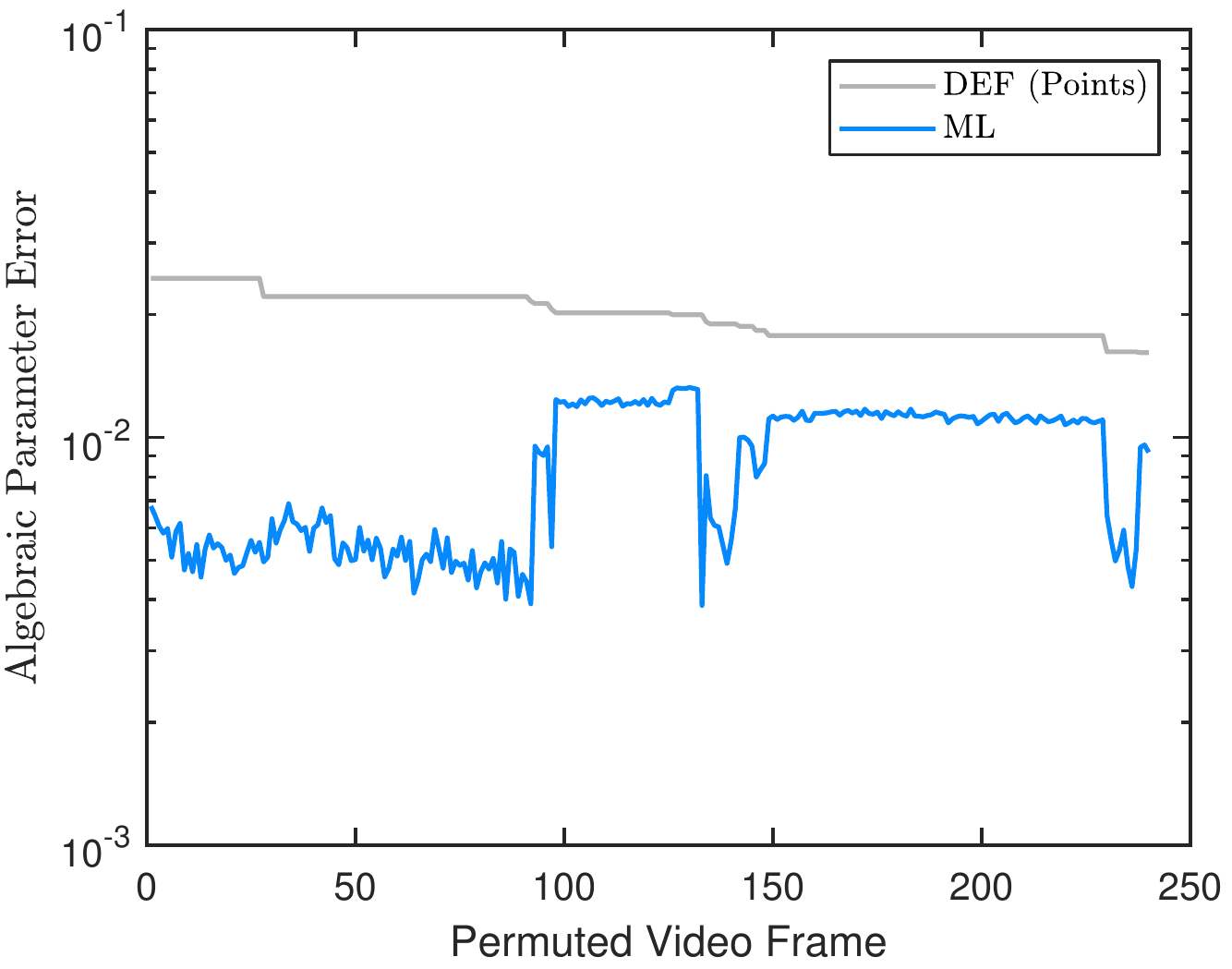}} 
  \hfil
  \subfloat[]{\includegraphics[scale = 0.4,trim=0 -2.7cm 0 0]{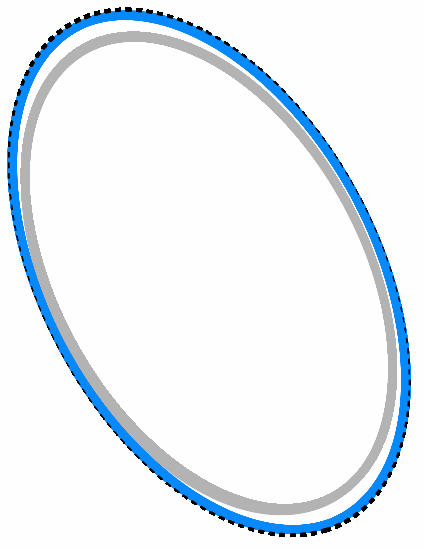}}
  %
  % \begin{tabular}{cc}
  %   \subfloat[Actual image]{\includegraphics[scale = 0.4]{blur-light-downsampled-2-camera}} 
  %   &
  %     \subfloat[Synthesised image]{\includegraphics[scale = 0.4]{blur-light-downsampled-2-model}} \\
  %   \subfloat[]{\includegraphics[scale = 0.4]{blur-light-downsampled-2-parameter-error}} 
  %   &
  %     \subfloat[]{\includegraphics[scale = 0.4,trim=0 -2.7cm 0 0]{blur-light-downsampled-2-ellipses}}
  % \end{tabular}
  %
  \caption{Experiment 2. Estimation results on a sequence of real  images ($32 \times 32$ pixels) where the lens was defocused to induce a blurred image. We permuted the sequence of images in the bottom left panel so that the error of the point-based direct ellipse fit is sorted in descending order. The bottom right panel shows the estimates for the entire sequence overlayed on a single figure. The dotted black line demarcates the true ellipse. }
  \label{fig:experiment2}
\end{figure}

The second experiment was identical to the first, except that we configured the camera to downsample the resolution by a half. After cropping the downsampled image, we obtained a $32 \times 32$ square grid of pixels that encapsulated the ellipse region.  We initialised the Gaussian PSF with a standard deviation of $1\%$ and set $b$ equal to $0$.  The results are displayed in \cref{fig:experiment2}.

\paragraph*{Experiment 3}
\label{sec:experiment-3}

\begin{figure}[t]
  \centering
  \captionsetup[subfigure]{labelformat=empty}
  \subfloat[Actual image]{\includegraphics[scale = 0.2]{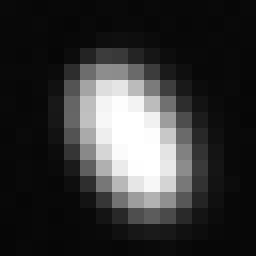}} 
  \hfil
  \subfloat[Synthesised image]{\includegraphics[scale = 0.2]{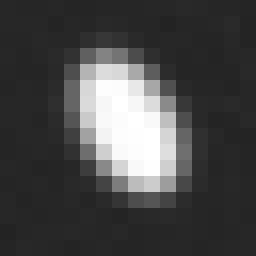}}
  
  \subfloat[]{\includegraphics[scale = 0.4]{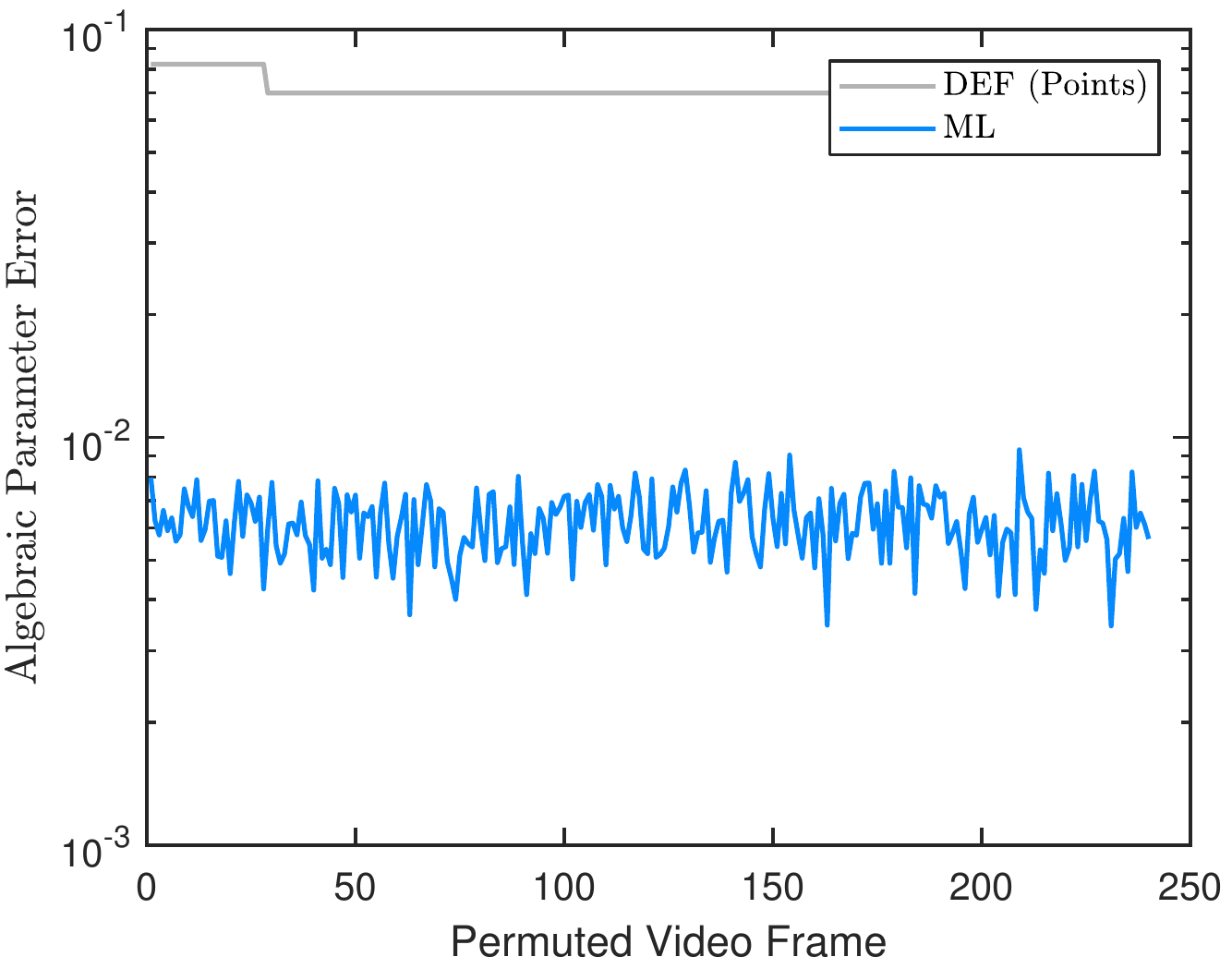}} 
  \hfil
  \subfloat[]{\includegraphics[scale = 0.4,trim=0 -2.7cm 0 0]{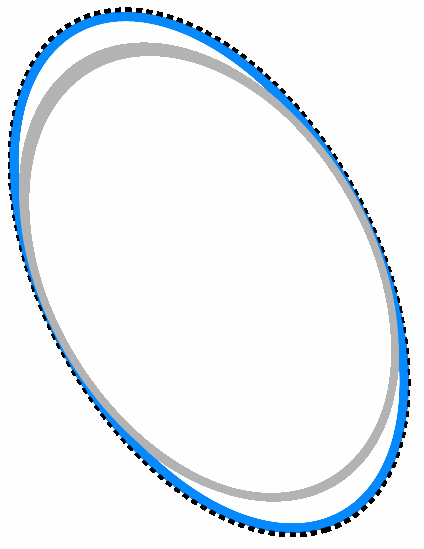}}
  % 
  %
  % \begin{tabular}{cc}
  %   \subfloat[Actual image]{\includegraphics[scale = 0.4]{blur-light-downsampled-4-camera}} 
  %   &
  %     \subfloat[Synthesised image]{\includegraphics[scale = 0.4]{blur-light-downsampled-4-model}} \\
  %   \subfloat[]{\includegraphics[scale = 0.4]{blur-light-downsampled-4-parameter-error}} 
  %   &
  %     \subfloat[]{\includegraphics[scale = 0.4]{blur-light-downsampled-4-ellipses}}
  % \end{tabular}
  %
  \caption{Experiment 3. Estimation results on a sequence of real images ($16 \times 16$ pixels) where the lens was defocused to induce a blurred image. We permuted the sequence of images in the bottom left panel so that the error of the point-based direct ellipse fit is sorted in descending order. The bottom right panel shows the estimates for the entire sequence overlayed on a single figure. The dotted black line demarcates the true ellipse. }
  \label{fig:experiment3}
\end{figure}

The third experiment was also identical to the first, except that we configured the camera to downsample the resolution by a quarter. Downsampling and cropping produced a $16 \times 16$ pixel grid of the ellipse region. We initialised the Gaussian PSF with a standard deviation of $2\%$ and set $b$ equal to $0$.  The results are displayed in \cref{fig:experiment3}.

\paragraph*{Experiment 4}
\label{sec:experiment-4}

\begin{figure}[!t]
  \centering
  \captionsetup[subfigure]{labelformat=empty}
  \subfloat[Actual image]{\includegraphics[scale = 0.2]{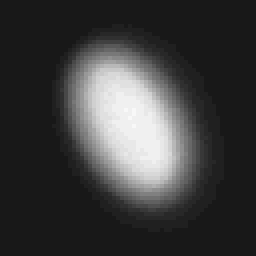}} 
  \hfil
  \subfloat[Synthesised image]{\includegraphics[scale = 0.2]{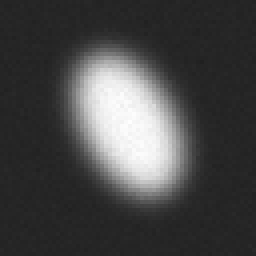}} \\

  \subfloat[]{\includegraphics[scale = 0.4]{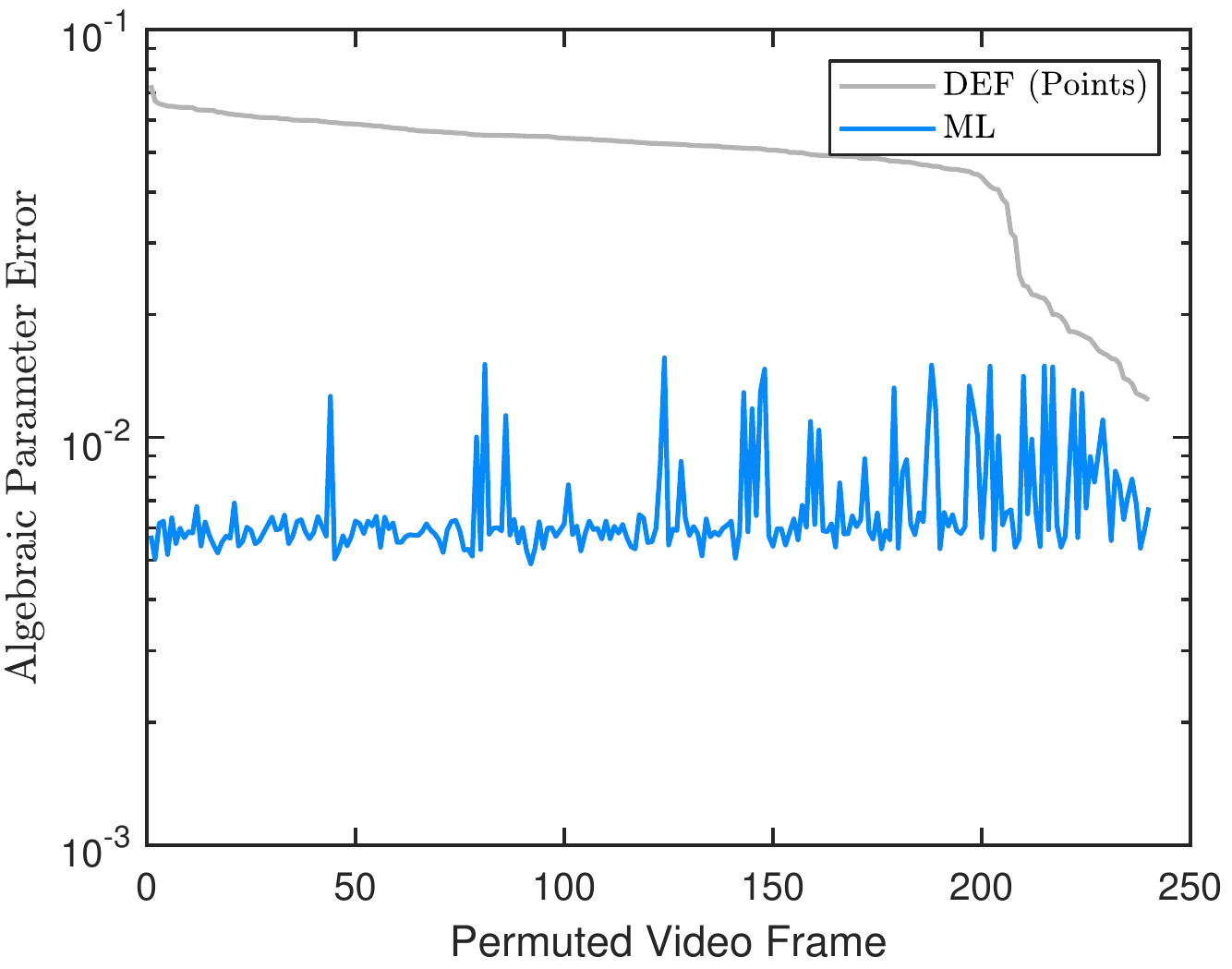}} 
  \hfil
  \subfloat[]{\includegraphics[scale = 0.4,trim=0 -2.7cm 0 0]{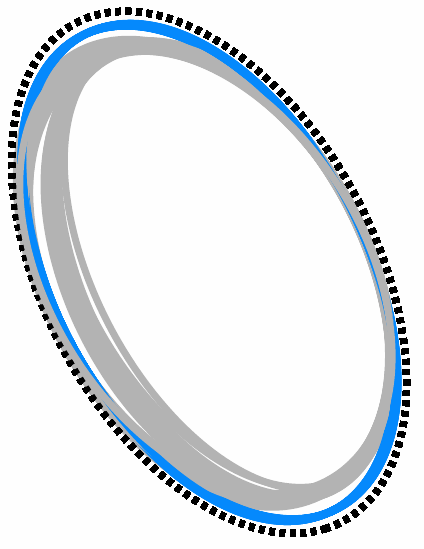}}
  %
  % \begin{tabular}{cc}
  %   \subfloat[Actual image]{\includegraphics[scale = 0.4]{blur-light-quantised-camera}} 
  %   &
  %     \subfloat[Synthesised image]{\includegraphics[scale = 0.4]{blur-light-quantised-model}} \\
  %   \subfloat[]{\includegraphics[scale = 0.4]{blur-light-quantised-parameter-error}} 
  %   &
  %     \subfloat[]{\includegraphics[scale = 0.4]{blur-light-quantised-ellipses}}
  %     \end{tabular}
  %
  \caption{Experiment 4. Estimation results on a sequence of real  quantised images ($64 \times 64$ pixels and $32$ grey levels) where the lens was defocused to induce a blurred image. We permuted the sequence of images in the bottom left panel so that the error of the point-based direct ellipse fit is sorted in descending order. The bottom right panel shows the estimates for the entire sequence overlayed on a single figure. The dotted black line demarcates the true ellipse. }
  \label{fig:experiment4}
\end{figure}

In the fourth experiment, we repeated the first experiment but this time configured the camera to quantise the luminance to $5$ bits ($32$ grey levels).  We initialised the Gaussian PSF with a standard deviation of $.5\%$ and set $b$ equal to $4$.  The results are displayed in Figure~\cref{fig:experiment4}.

\paragraph*{Experiment 5}
\label{sec:experiment-5}

\begin{figure}[!t]
  \centering
  \captionsetup[subfigure]{labelformat=empty}
  \subfloat[Actual image]{\includegraphics[scale = 0.2]{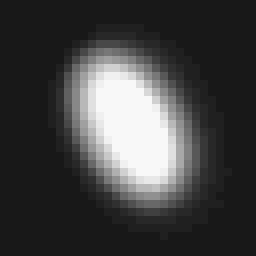}}
  \hfil
  \subfloat[Synthesised image]{\includegraphics[scale = 0.2]{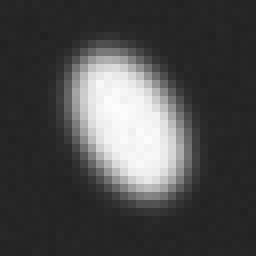}}
   
  \subfloat[]{\includegraphics[scale = 0.4]{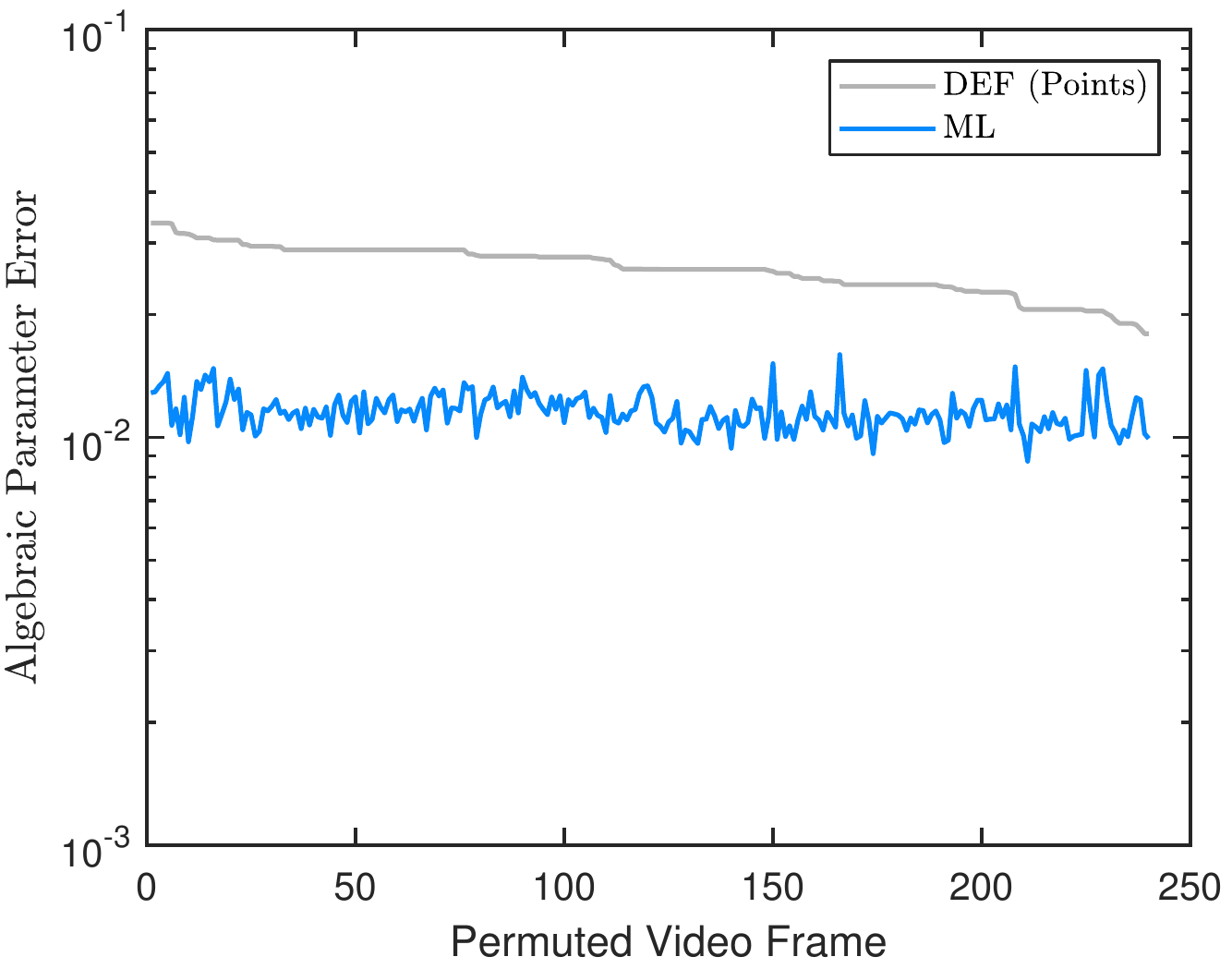}} 
  \hfil
  \subfloat[]{\includegraphics[scale = 0.4,trim=0 -2.8cm 0 0]{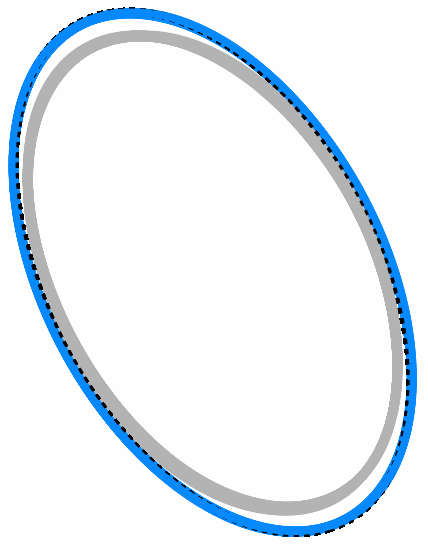}}
  %
  % \begin{tabular}{cc}
  %   \subfloat[Actual image]{\includegraphics[scale = 0.4]{blur-light-quantised-downsampled-2-camera}}
  %   &
  %     \subfloat[Synthesised image]{\includegraphics[scale = 0.4]{blur-light-quantised-downsampled-2-model}}
  %   \\
  %   \subfloat[]{\includegraphics[scale = 0.4]{blur-light-quantised-downsampled-2-parameter-error}} 
  %   &
  %     \subfloat[]{\includegraphics[scale = 0.4]{blur-light-quantised-downsampled-2-ellipses}}
  %   \end{tabular}
  %
  \caption{Experiment 5. Estimation results on a sequence of real  quantised images ($32 \times 32$ pixels and $32$ grey levels) where the lens was defocused to induce a blurred image. We permuted the sequence of images in the bottom left panel so that the error of the point-based direct ellipse fit is sorted in descending order. The bottom right panel shows the estimates for the entire sequence overlayed on a single figure. The dotted black line demarcates the true ellipse. }
  \label{fig:experiment5}
\end{figure}

The fifth experiment mirrored the second experiments, except that we also configured the camera to quantise the luminance to $5$ bits ($32$ grey levels).  We initialised the Gaussian with a standard deviation of $1\%$ and set $b$ equal to $4$.  The results are displayed  in \cref{fig:experiment5}.

\paragraph*{Experiment 6}
\label{sec:experiment-6}

\begin{figure}[!t]
  \centering
  \captionsetup[subfigure]{labelformat=empty}
  \subfloat[Actual image]{\includegraphics[scale = 0.2]{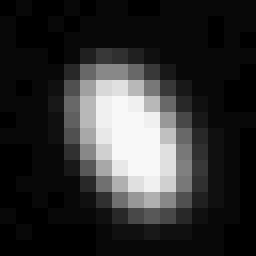}} 
  \hfil
  \subfloat[Synthesised image]{\includegraphics[scale = 0.2]{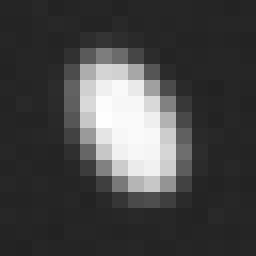}}
  
  \subfloat[]{\includegraphics[scale = 0.4]{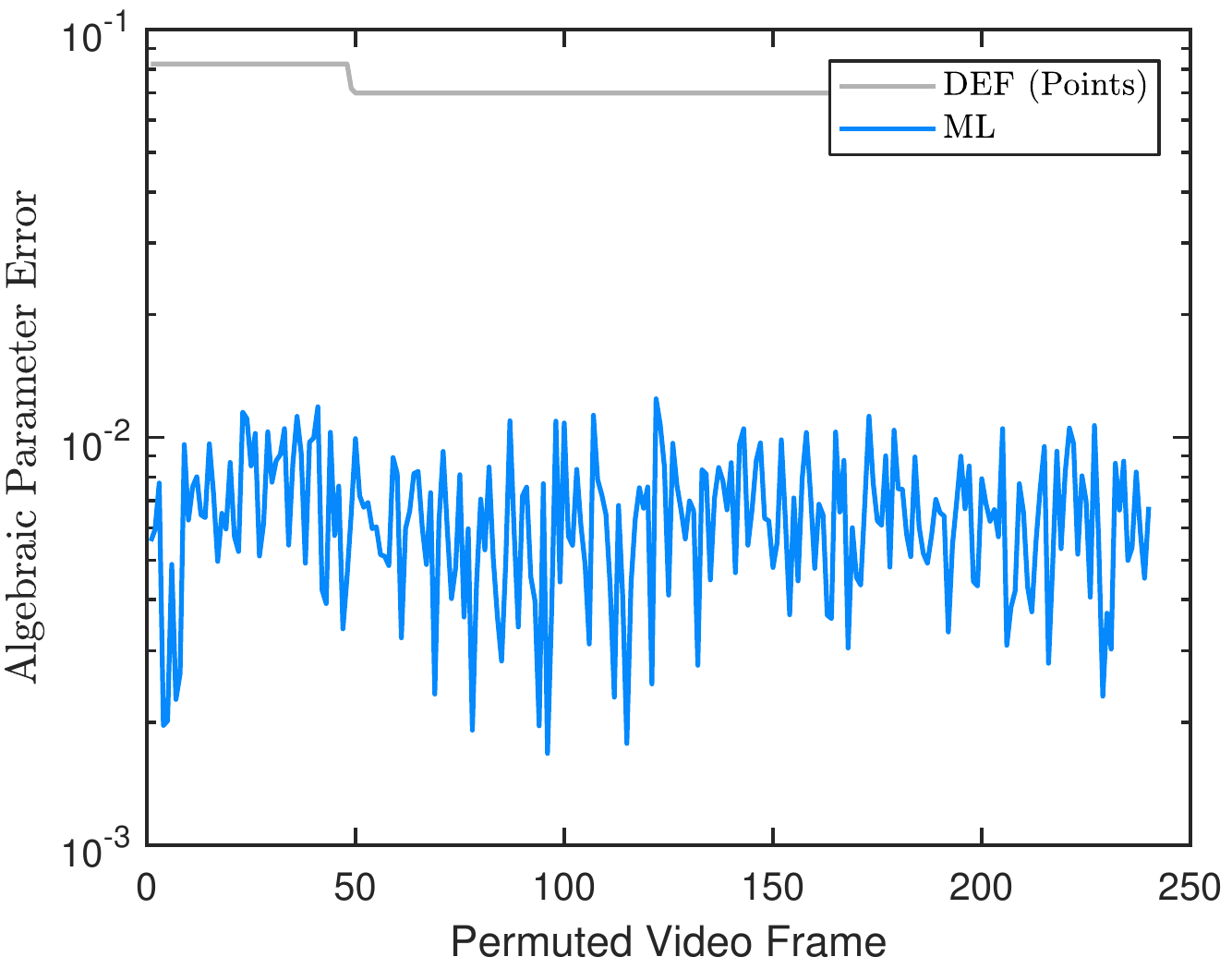}} 
  \hfil
  \subfloat[]{\includegraphics[scale = 0.4,trim=0 -2.7cm 0 0]{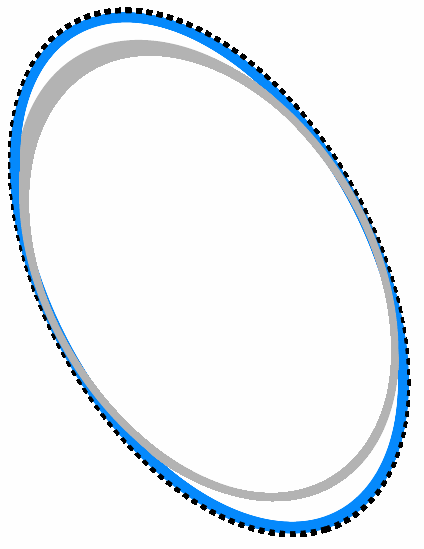}}
  %
  % \begin{tabular}{cc}
  %   \subfloat[Actual image]{\includegraphics[scale = 0.4]{blur-light-quantised-downsampled-4-camera}} 
  %   &
  %     \subfloat[Synthesised image]{\includegraphics[scale = 0.4]{blur-light-quantised-downsampled-4-model}} \\
  %   \subfloat[]{\includegraphics[scale = 0.4]{blur-light-quantised-downsampled-4-parameter-error}} 
  %   &
  %     \subfloat[]{\includegraphics[scale = 0.4]{blur-light-quantised-downsampled-4-ellipses}}
  %   \end{tabular}
  %
  \caption{Experiment 6. Estimation results on a sequence of real  quantised images ($16 \times 16$ pixels and 32 grey levels) where the lens was defocused to induce a blurred image. We permuted the sequence of images in the bottom left panel so that the error of the point-based direct ellipse fit is sorted in descending order. The bottom right panel shows the estimates for the entire sequence overlayed on a single figure. The dotted black line demarcates the true ellipse. }
  \label{fig:experiment6}
\end{figure}

The sixth experiment mirrored the third experiments, except that we also configured the camera to quantise the luminance to $5$ bits ($32$ grey levels).  We initialised the Gaussian PSF with a standard deviation of $2\%$ and set $b$ equal to $4$.  The results are displayed in \cref{fig:experiment6}.

\section{Discussion}
\label{sec:discussion}

The experiments that we conducted on real imagery have further demonstrated the correctness and versatility of our statistical model.  It is remarkable that for each experiment, the synthetic image associated with the maximum likelihood solution is visually almost indistinguishable from the real picture.  Evidently, our mathematical development strikes the correct balance between tractability and authenticity.

For each experiment, our maximum likelihood method outperformed the point-based direct ellipse fit by several orders of magnitude. The variance of the ML estimator is also substantially less than the point-based estimates. The stability of the ML estimate is apparent in the overlayed ellipse plots. Substantially only a single blue ellipse (ML) is evident for each experiment in contrast to numerous grey curves~(DEF).

\section{Conclusion}
\label{sec:conclusion}

We have developed and tested a coherent mathematical framework for estimating the parameters of a planar shape from a single low-resolution, photon-limited digital image. Our work unifies the uncertainty due to discretisation, photon noise, and quantisation into a unique manageable statistical model. We have presented a careful and meticulous exposition of each component of the model. Comprehensive experiments on real and synthetic data have also demonstrated the groundbreaking accuracy of our approach. The ideas presented in this report provide new foundations for working on image processing problems at the limits of resolution. Our future work will focus on generalising the method to other more complicated shapes, with one possible approach being the use of level-set methods and dynamic implicit surfaces. The main problem to resolve is how to compute the area of intersection of a pixel with a particular shape.

% \paragraph*{\large Funding}
% Australian Defence Science and Technology Group (DSTGroup) (CERA grant no.~52).

% \paragraph*{\large Acknowledgements}

% The authors would like to thank Garry Newsam for many fruitful discussions and Nick Redding for framing the research question and advocating the work.

\section*{Appendix A: Area of intersection of an ellipse and a rectangle}
%\label{area-intersection-ellipse-rectangle}
\setcounter{equation}{0}
\renewcommand{\theequation}{A{\arabic{equation}}}

The problem of determining the area of intersection between a rectangle and an ellipse in the case that the sides of the rectangle are parallel to the semi-axes of the ellipse was addressed in 1963 by Groves \cite{groves1963area} in the military context of devising mathematical methods for the evaluation of small arms.  Our exposition of the solution, including several diagrams, is based on Groves' systematic account.

\begin{figure}[!t]
  \centering
  \includegraphics[scale                                             =
  1]{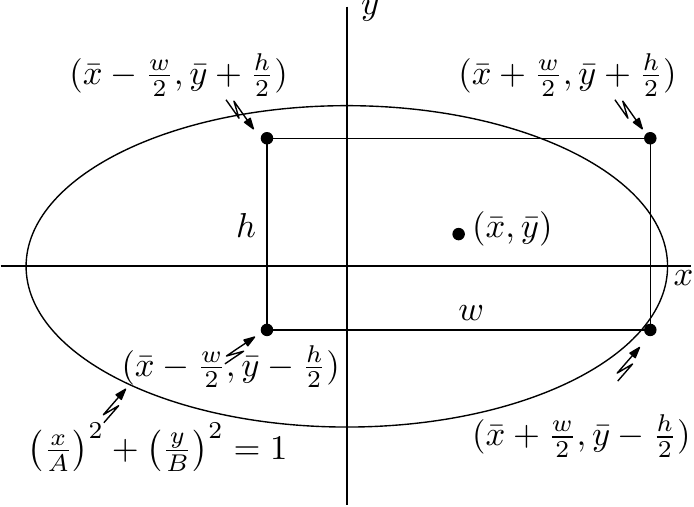}
  \caption{General situation.}
  \label{ellipse-rectangle-intersection-general-situation}
\end{figure}

\begin{figure}[!t]
  \centering
  \includegraphics[scale                                             =
  1]{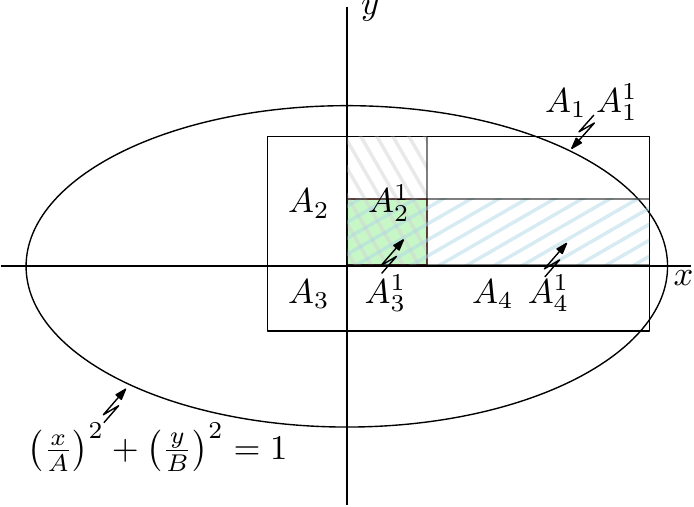}
  \caption{A partitioning  of the  rectangle into  four sub-rectangles
    $A_1,A_2,A_3$, and  $A_4$   together  with  equivalent  rectangles
    $A_{1}^{1},A_{2}^{1},A_{3}^{1}$, and $A_{4}^{1}$ all located in the
    first quadrant of the Cartesian coordinate system.}
  \label{ellipse-rectangle-intersection-split-2}
\end{figure}

\begin{figure}
  \centering
  \includegraphics[scale                                             =
  1]{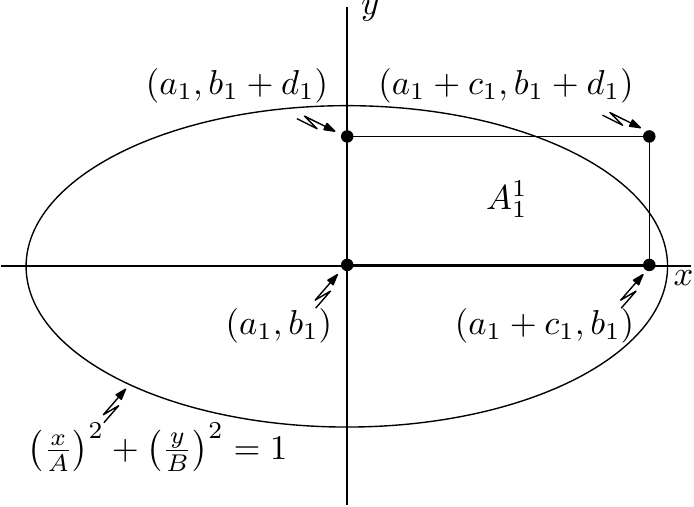}
  \caption{Representation of $A_1^{1}$.}
  \label{ellipse-rectangle-intersection-coordinates}
\end{figure}

We shall assume that the ellipse is described by the ellipse equation in standard form
\begin{equation}
  \left(\frac{x}{A}\right)^2 + \left(\frac{y}{B}\right)^2 = 1,
\end{equation}
where $A$ and $B$ represent the ellipse's semi-major and semi-minor axis lengths, respectively.  Furthermore, we shall assume that the rectangle is centred at a point $(\bar{x}, \bar{y})$ and has width $w$ and height $h$ (see \cref{ellipse-rectangle-intersection-general-situation}).
Groves' solution for computing the intersection area between an arbitrary rectangle and a standard ellipse involves partitioning the rectangle into sub-rectangles $A_i$ ($i = 1\ldots 4$) such that each sub-rectangle is entirely contained in one of the four quadrants of the Cartesian coordinate system. Then, for each part, one constructs an equivalent first-quadrant rectangle $A_i^{1}$ and calculates its area of intersection, $S_i$, with the ellipse (see \cref{ellipse-rectangle-intersection-split-2}).
If $S$ denotes the total area of intersection for the original rectangle, then $S = \sum_{i=1}^4 S_i$.  Each first quadrant rectangle $A_i^{1}$ will be specified by four non-negative numbers ($a_i, b_i, c_i, d_i$), where ($a_i,b_i$) are the coordinates of the vertex of $A_i^{1}$ closest to the origin, and $c_i$ is the width and $d_i$ is the height of $A_i^{1}$ in $x$ and $y$ directions, respectively (see \cref{ellipse-rectangle-intersection-coordinates} for an example).
Groves derived formulae for ($a_i, b_i, c_i, d_i$) in terms of the original rectangle by enumerating the different ways in which the rectangle can span the four quadrants. There are nine possible cases:
\begin{enumerate*}
\item  [(1--4)]the  rectangle  is  completely   in  one  of  the  four
  quadrants;
\item [(5)] partly in quadrant \rom{1} and \rom{2};
\item [(6)] partly in quadrant \rom{2} and \rom{3};
\item [(7)] partly in quadrant \rom{3} and \rom{4};
\item [(8)] partly in quadrant \rom{4} and \rom{1};
\item [and  (9)] one vertex  of the rectangle is  in each of  the four
  quadrants.
\end{enumerate*}
These nine cases are all simultaneously handled by the following formulae:
\begin{equation}
  \begin{aligned}[b]
    A_1^1: \qquad
    a_1 & = \max\left\{0, \bar{x} - \frac{w}{2} \right\},
    \\
    b_1 & = \max\left\{0, \bar{y} - \frac{h}{2} \right\},
    \\
    c_1 & = \max\left\{0, \bar{x} + \frac{w}{2} - a_1 \right\},
    \\
    d_1 & = \max\left\{0, \bar{y} + \frac{h}{2} - b_1 \right\};
  \end{aligned}
\end{equation}
\begin{equation}
  \begin{aligned}[b]
    A_2^1: \qquad
    a_2 &= \max\left\{0, -\bar{x} - \frac{w}{2} \right\},
    \\
    b_2 &= \max\left\{0, \bar{y} - \frac{h}{2} \right\},
    \\
    c_2 &= \max\left\{0, -\bar{x} + \frac{w}{2} - a_2 \right\},
    \\
    d_2 &= \max\left\{0, \bar{y} + \frac{h}{2} - b_2 \right\};
  \end{aligned}
\end{equation}
\begin{equation}
  \begin{aligned}[b]
   A_3^1: \qquad
  a_3 &= \max\left\{0, -\bar{x} - \frac{w}{2} \right\},
  \\
  b_3 &= \max\left\{0, -\bar{y} -\frac{h}{2} \right\},
  \\
  c_3 &= \max\left\{0, -\bar{x} + \frac{w}{2} - a_3 \right\},
  \\
  d_3 &= \max\left\{0, -\bar{y} + \frac{h}{2} - b_3 \right\};
\end{aligned}
\end{equation}
\begin{equation}
  \begin{aligned}[b]
  A_4^1: \qquad
  a_4 &= \max\left\{0, \bar{x} - \frac{w}{2} \right\},
  %\displaybreak[3]
  \\
  b_4 &= \max\left\{0, -\bar{y} -\frac{h}{2} \right\},
  %\displaybreak[3]
  \\
  c_4 &= \max\left\{0, \bar{x} + \frac{w}{2} - a_4 \right\},
  %\displaybreak[3]
  \\
  d_4 &= \max\left\{0, -\bar{y} + \frac{h}{2} - b_4 \right\}.
\end{aligned}
\end{equation}
which can also be expressed as a function of $i$,
\begin{equation}
\begin{aligned}[b]
  A_i^1: \qquad
  a_i &= \max\left\{0, (-1)^{\frac{1}{2}(i^2-1)}\bar{x} - \frac{w}{2} \right\},
  %\displaybreak[3]
  \\
  b_i &= \max\left\{0, (-1)^{\frac{1}{2}(i^2+i-2)}\bar{y}- \frac{h}{2} \right\},
  %\displaybreak[3]
  \\
  c_i &= \max\left\{0, (-1)^{\frac{1}{2}(i^2-i)}\bar{x} + \frac{w}{2} - a_i\right\},
  %\displaybreak[3]
  \\
  d_i &= \max\left\{0, (-1)^{\frac{1}{2}(i^2+i-2)} \bar{y} + \frac{h}{2} - b_i \right\}.
\end{aligned}
\end{equation}
If the original rectangle does not overlap with the $i$th quadrant, then $A_i^1$ will reduce to a line segment with no area (either $c_i$ or $d_i$ will be zero) and the area of intersection $S_i$ will be zero.

Dropping the subscripts, we now focus exclusively on deriving formulae for the intersection area of a rectangle in the first quadrant. Let the four vertices of the rectangle be indexed in the following manner according to their coordinates:
\begin{equation}
  \begin{aligned}[b]
    v_2&: (a, b+d), &
    v_3&: (a+c, b+d), \\
    v_1&: (a,b), & v_4&: (a+c, b).
  \end{aligned}
\end{equation}
There are six distinct intersection cases that need to be considered, depending on which vertices are inside the ellipse.  These are:
%\begin{enumerate}
\begin{enumerate}[label={\sffamily\bfseries \Roman*.}]
% \begin{enumerate}[label={\normalfont\bfseries \Roman*}]
\item no vertices inside the ellipse,
\item $v_1$ inside; $v_2, v_3$ and $v_4$ outside,
\item $v_1$ and $v_4$ inside; $v_2$ and $v_3$ outside,
\item $v_1$ and $v_2$ inside; $v_3$ and $v_4$ outside,
\item $v_1, v_2$ and $v_4$ inside; $v_3$ outside
\item all vertices inside the ellipse.
\end{enumerate}

\paragraph{Case \textbf{\rom{1}}}

This      case      is      identified      by      the      condition
\begin{equation}
  \left(\frac{a}{A}\right)^2 + \left(\frac{b}{B}\right)^2 \geq 1
\end{equation}
which  indicates that  the vertex  closest  to the  origin, $v_1$,  is
outside the ellipse.  Consequently,  the area of intersection, denoted
by $S^{\rom{1}}$, is zero.

\paragraph{Case \textbf{\rom{2}}}
\begin{figure}[!t]
  \centering
  \includegraphics[scale                                             =
  1]{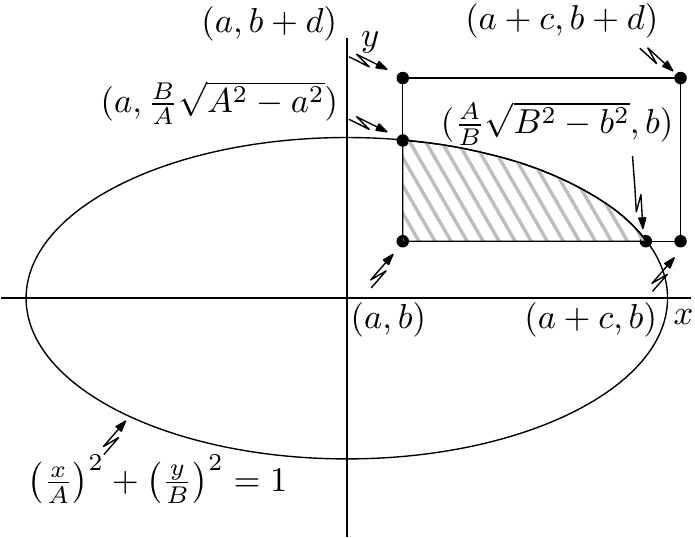}
  \caption{Case \rom{2}: $v_1$ inside, and $v_2, v_3$, and $v_4$ outside.}
  \label{ellipse-rectangle-intersection-case-2}
\end{figure}
\begin{figure}[!t]
  \centering
  \includegraphics[scale                                             =
  1]{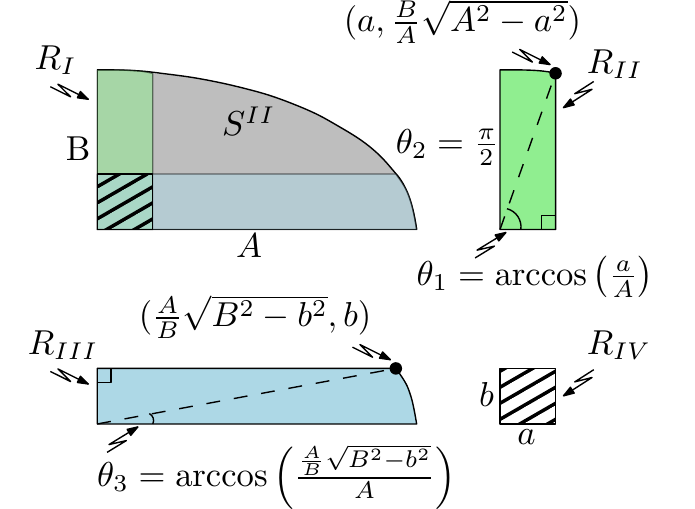}
  \caption{Case     \rom{2}.     The      area     of     intersection
    $S^{\rom{2}}  =   R_{\rom{1}}  -   R_{\rom{2}}  -   R_{\rom{3}}  +
    R_{\rom{4}}$.}
  \label{ellipse-rectangle-intersection-case-2-partition}
\end{figure}
The conditions required to identify this case are

% \begin{align}
%   \left(\frac{a}{A}\right)^2  +  \left(\frac{b}{B}\right)^2
%   & <  1  && \text{($v_1$ inside)},
%   \\
%   \left(\frac{a}{A}\right)^2  +  \left(\frac{b+d}{B}\right)^2
%   & \geq  1 && \text{($v_2$ outside)}
% \end{align}
% and
% \begin{align}
%                \left(\frac{a+c}{A}\right)^2 +
%                \left(\frac{b}{B}\right)^2 
%   & \geq  1 &&  \text{($v_4$ outside)}.
% \end{align}

\begin{align}
  \left(\frac{a}{A}\right)^2  +  \left(\frac{b}{B}\right)^2
  & <  1  && \text{($v_1$ inside)},
  \\
  \left(\frac{a}{A}\right)^2  +  \left(\frac{b+d}{B}\right)^2
  & \geq  1 && \text{($v_2$ outside)}
               \shortintertext{and}
               \left(\frac{a+c}{A}\right)^2 +
               \left(\frac{b}{B}\right)^2 
  & \geq  1 &&  \text{($v_4$ outside);}
\end{align}
see \cref{ellipse-rectangle-intersection-case-2}.
If we partition the first quadrant into four regions as illustrated in \cref{ellipse-rectangle-intersection-case-2-partition}, then the area of intersection is given by
\begin{align}
  S^{\rom{2}}
  & = R_{\rom{1}} - R_{\rom{2}} - R_{\rom{3}} + R_{\rom{4}} 
    \nonumber
  \\
  & = \underbrace{\left( \vphantom{\frac{(\theta_2 -
    \theta_1)AB}{2} } \frac{\pi}{4}AB
    \right)}_{R_{\rom{1}}}  -  \underbrace{\left(
    \frac{(\theta_2 - \theta_1)AB}{2} +
    \frac{aB}{2A}\sqrt{A^2-a^2} \right)}_{R_{\rom{2}}}
    \nonumber  
  \displaybreak[3] \\
  & \quad - \underbrace{\left( \frac{\theta_3 AB}{2} +
    \frac{bA}{2B}\sqrt{B^2-b^2} \right) }_{R_{\rom{3}}}
    +\underbrace{\left( \vphantom{\frac{\theta_3 AB}{2}}ab
    \right)}_{R_{\rom{4}}}   \label{sectorTriangle} 
    \displaybreak[3] \\
  & = \frac{\pi}{4}AB + ab
    \nonumber
  \\
  & \quad - \underbrace{\left(
    \frac{AB}{2}\arcsin{\left(\frac{a}{A} \right)} +
    \frac{aB}{2}\sqrt{1- \left(\frac{a}{A} \right)^2}
    \right)}_{R_{\rom{2}}}  
    \nonumber
  \\
  & \quad -  \multiunderbracee{\left( \frac{\pi}{4}AB  -
    \frac{AB}{2} \arcsin{\left( \sqrt{1-\left(\frac{b}{B}
    \right)^2} \right)} \right.}_{\hspace{4\mytextsize}
    R_{\rom{3}}} 
    \nonumber
  \\
  & \quad +  \multiunderbraced{
    \left. \frac{bA}{2}\sqrt{1-\left(\frac{b}{B}
    \right)^2} \right)}  \label{arcsinArccos}   
    \displaybreak[3] \\
  & = \frac{AB}{2}\left(  \arcsin{\left(
    \sqrt{1-\left(\frac{b}{B} \right)^2} \right)} -
    \arcsin{\left(\frac{a}{A} \right)}  \right. 
    \nonumber
  \\
  &  \quad -
    \left. \frac{a}{A}\sqrt{1-\left(\frac{a}{A}\right)^2}
    -  \frac{b}{B}\sqrt{1-\left(\frac{b}{B}\right)^2} + 2
    \left(\frac{a}{A}\right) \left(\frac{b}{B} \right)
    \right).  
    %\nonumber
\end{align}
In \eqref{sectorTriangle} regions $R_{\rom{2}}$ and $R_{\rom{3}}$ are each partitioned into the sum of two terms: the area of an ellipse sector (first term) and the area of a right-angled triangle (second term). The angles $\theta_k$ ($k = 1,3$) that are formed between the $x$ axis and corresponding points $(x_k,y_k)$ on the ellipse are found from the first of the following parametric ellipse equations:
\begin{equation}
\begin{aligned}[b]
  x = A\cos \theta
  & \implies
    \theta = \arccos \frac{x}{A}, \\
  y = B\sin \theta
  & \implies
    \theta  = \arcsin \frac{y}{B}.
  \end{aligned}
\end{equation}
In \eqref{arcsinArccos} we use the complementary angle relation, $ \arccos x = \frac{\pi}{2} - \arcsin x $, to write the ellipse sector area in $R_{\rom{2}}$ and $R_{\rom{3}}$ in terms of the arcsine function.  For further simplification, let
\begin{equation}
  \theta =  \arcsin{\left( \sqrt{1-\left(\frac{b}{B} \right)^2}
  \right)} 
           - \arcsin{\left(\frac{a}{A} \right)}. 
\end{equation}
Then
\begin{equation}
  \sin \theta =  \sqrt{1-\left(\frac{b}{B} \right)^2}
  \sqrt{1-\left(\frac{a}{A} \right)^2}
  -
  \left(\frac{a}{A}\right) \left(\frac{b}{B}\right)
  \label{doubleAnglePythag}
\end{equation}
or
\begin{equation}
  \theta =  \arcsin{\left(\sqrt{1-\left(\frac{b}{B} \right)^2}  \sqrt{1-\left(\frac{a}{A} \right)^2} - \left(\frac{a}{A}\right) \left(\frac{b}{B}\right) \right)}.
\end{equation}
In \eqref{doubleAnglePythag} we used the angle-difference identity for sine followed by the Pythagorean trigonometric identity.  In conclusion,
\begin{equation}
  S^{\rom{2}}
  =
  \frac{AB}{2} F\left(\frac{a}{A}, \frac{b}{B}\right),
\end{equation}
where
\begin{equation}
  \begin{aligned}[b]
    F\left(U,V \right)
    & =
    \arcsin{\left(\sqrt{1-U^2}\sqrt{1-V^2}  - UV\right)}
    \\
    & \quad - U\sqrt{1-U^2} - V\sqrt{1-V^2} + 2UV.
  \end{aligned}
\end{equation}

\paragraph{Case \textbf{\rom{3}}}

\begin{figure}
  \centering
  \includegraphics[scale                                             =
  1]{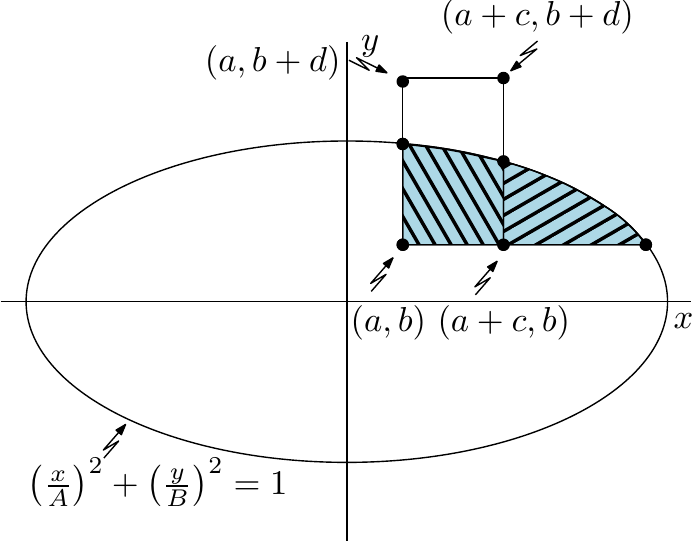}
  \caption{Case \rom{3}: $v_1$  and $v_2$ inside, and  $v_3$ and $v_4$
    outside. The  area of  intersection equals the  joint area  of the
    falling  and  rising  striped  region  minus  the  rising  striped
    region. Both these regions are captured by case~\rom{2}. }
  \label{ellipse-rectangle-intersection-case-3}
\end{figure}

The situation is identified by the conditions
\begin{align}
  \left(\frac{a+c}{A}\right)^2  +  \left(\frac{b}{B}\right)^2  & <  1  && \text{($v_4$ inside)} \\
  \shortintertext{and}
  \left(\frac{a}{A}\right)^2 + \left(\frac{b+d}{B}\right)^2 & \geq  1 &&  \text{($v_2$ outside)}.
\end{align}
The illustration \cref{ellipse-rectangle-intersection-case-3} suggests that the area $S^{\rom{3}}$ is simply the difference between two areas of the type considered in the second case.  Thus
\begin{equation}
  S^{\rom{3}}
  = \frac{AB}{2} \left(F\left(\frac{a}{A}, \frac{b}{B} \right)
  -   F\left(\frac{a+c}{A}, \frac{b}{B} \right)  \right).
\end{equation}

\paragraph{Case \textbf{\rom{4}}}

The conditions required to identify this case are
\begin{align}
  \left(\frac{a}{A}\right)^2  +  \left(\frac{b+d}{B}\right)^2  & <  1  && \text{($v_2$ inside)} \\
  \shortintertext{and}
  \left(\frac{a+c}{A}\right)^2 + \left(\frac{b}{B}\right)^2 & \geq  1 &&  \text{($v_4$ outside)}.
\end{align}
This area is also difference between  two areas of the type considered
in Case \rom{2}:
\begin{equation}
  S^{\rom{4}} = \frac{AB}{2} \left(F\left(\frac{a}{A}, \frac{b}{B} \right) -   F\left(\frac{a}{A}, \frac{b+d}{B} \right)  \right).
\end{equation}

\paragraph{Case \textbf{\rom{5}}}

The three conditions required to identify this case are
\begin{align}
  \left(\frac{a}{A}\right)^2  +  \left(\frac{b+d}{B}\right)^2  & <  1  && \text{($v_2$ inside)} \\
  \left(\frac{a+c}{A}\right)^2 + \left(\frac{b}{B}\right)^2 & <  1 &&  \text{($v_4$ inside)}.
                                                                      \shortintertext{and}
                                                                      \left(\frac{a+c}{A}\right)^2 + \left(\frac{b+d}{B}\right)^2 & \geq  1 &&  \text{($v_3$ outside)}.
\end{align}
Applying the result given in Case \rom{2}, the area is
\begin{equation}
  S^{\rom{5}} =  \frac{AB}{2} \left(F\left(\frac{a}{A}, \frac{b}{B} \right) -   F\left(\frac{a+c}{A}, \frac{b}{B} \right)  - 
                F\left(\frac{a}{A}, \frac{b+d}{B} \right)  \right).
\end{equation}

\paragraph{Case \textbf{\rom{6}}}

The sole condition required to identify this case is
\begin{align}
  \left(\frac{a+c}{A}\right)^2  +  \left(\frac{b+d}{B}\right)^2  & \le  1  && \text{($v_3$ inside)}.
\end{align}
Since all  of the vertices  are inside the ellipse,  the intersection
area is simply $S^{\rom{6}} = cd$.

\section*{Acknowledgements}

The authors would like to thank Garry Newsam for many fruitful discussions and Nick Redding for framing the research question and advocating the work. Figures were rendered with colours suitable for red-green colour blind readers using Peter Kovesi's colour maps \cite{kovesi15:_colour}.  This study was funded by the Australian Defence Science and Technology Group under CERA grant no.~52.

%\FloatBarrier

% Bibliography

%\bibliographystyle{osajnl}
% \bibliographystyle{IEEEtran}

% \bibliography{target-bib}


\end{document}